%% file: main.tex
\documentclass[11pt]{article}

\usepackage[final]{acl}

\usepackage{times}
\usepackage{latexsym}
\usepackage{pifont}
\usepackage{graphicx}
\usepackage{subcaption}

\usepackage[T1]{fontenc}

\usepackage[utf8]{inputenc}

\usepackage{microtype}
\usepackage[table]{xcolor}
\usepackage{multirow}

\usepackage{booktabs}
\usepackage{longtable}
\usepackage{makecell,tabularx}
\usepackage{inconsolata}
\usepackage{amsmath, amssymb}
\usepackage{graphicx}
\usepackage[ruled,vlined]{algorithm2e}

\usepackage[most]{tcolorbox}
\newcommand{\gconst}[1]{\tcbox[colback=gray!10,colframe=gray!60,boxrule=0.4pt,arc=2pt,
  on line, boxsep=0.6pt, left=2pt, right=2pt, top=0.6pt, bottom=0.6pt]{\texttt{#1}}}

\usepackage[table]{xcolor}
\definecolor{OpenBest}{RGB}{226,245,230}
\definecolor{ClosedBest}{RGB}{226,235,252}

\newcolumntype{C}[1]{>{\centering\arraybackslash}m{#1}}

\definecolor{BlueLight}{RGB}{226,235,252}
\definecolor{BlueMain}{RGB}{47,99,171}

\usepackage{fvextra} 

\DefineVerbatimEnvironment{COTBlock}{Verbatim}{
  breaklines=true,
  breakanywhere=true,
  breaksymbolleft={},
  fontsize=\footnotesize
}
\DeclareCaptionType{prompt}[Prompt][List of Prompts]

\newtcolorbox{instance}[1]{
  enhanced,
  breakable,
  sharp corners,
  colback=white,
  colframe=black!50,
  boxrule=0.6pt,
  left=6pt,right=6pt,top=4pt,bottom=4pt,
  before skip=0pt, after skip=0pt,
  title=\textbf{#1},
  fonttitle=\normalsize,
}

\tcbuselibrary{listings,breakable}
\usepackage{listings}
\newtcblisting{cotrow}[1]{
  enhanced,
  breakable,
  sharp corners,
  colback=white,
  colframe=black!50,
  boxrule=0.6pt,
  left=6pt,right=6pt,top=4pt,bottom=4pt,
  title=\textbf{#1},
  listing engine=listings,
  listing only,
  listing options={
    basicstyle=\ttfamily\footnotesize,
    breaklines=true,
    breakatwhitespace=false,
    columns=fullflexible,
    keepspaces=true,
    mathescape=true,                 
    escapeinside={(*@}{@*)}          
  }
}
\usepackage{setspace}
\usepackage{enumitem}
\usepackage{capt-of} 
\tcbuselibrary{skins}

\newtcolorbox{promptboxstar}[1]{%
  enhanced,breakable,%
  colback=white,colframe=black,boxrule=0.6pt,arc=0pt,%
  left=10pt,right=10pt,top=8pt,bottom=8pt,%
  before skip=8pt,after skip=8pt,%
  title={#1},fonttitle=\bfseries,%
  colbacktitle=white,coltitle=black,%
  attach title to upper,%
  boxed title style={%
    boxrule=0pt,%
    colframe=white,%
    interior style={fill=white},%
    sharp corners%
  },%
}

\newcommand{\cotrowsep}{\vspace{-0.6pt}}

\usepackage{ragged2e}
\usepackage{xurl}
\usepackage{hyperref}

\title{Reasoning in a Combinatorial and Constrained World: Benchmarking LLMs on Natural-Language Combinatorial Optimization}

\author{
  \textbf{Xia Jiang\textsuperscript{1,*}},
  \textbf{Jing Chen\textsuperscript{2,*}},
  \textbf{Cong Zhang\textsuperscript{3}},
  \textbf{Jie Gao\textsuperscript{4}},
\\
  \textbf{Chengpeng Hu\textsuperscript{1}},
  \textbf{Chenhao Zhang\textsuperscript{5}},
  \textbf{Yaoxin Wu\textsuperscript{1}},
  \textbf{Yingqian Zhang\textsuperscript{1}}
\\
\\
  \textsuperscript{1}Eindhoven University of Technology,
  \textsuperscript{2}Brandenburg University of Technology,
\\
  \textsuperscript{3}Nanyang Technological University,
  \textsuperscript{4}Delft University of Technology,
\\
  \textsuperscript{5}Southeast University
\\
  \small{
    \textsuperscript{*}Equal contribution. 
    \textbf{Correspondence:} \href{mailto:y.wu2@tue.nl}{y.wu2@tue.nl}
  }
}

\begin{document}
\maketitle
\begin{abstract}

While large language models (LLMs) have shown strong performance in math and logic reasoning, their ability to handle combinatorial optimization (CO)---searching high-dimensional solution spaces under hard constraints---remains underexplored. To bridge the gap, we introduce NLCO, a \textbf{N}atural \textbf{L}anguage \textbf{C}ombinatorial \textbf{O}ptimization benchmark that evaluates LLMs on end-to-end CO reasoning: given a language-described decision-making scenario, the model must output a discrete solution without writing code or calling external solvers. NLCO covers 43 CO problems and is organized using a four-layer taxonomy of variable types, constraint families, global patterns, and objective classes, enabling fine-grained evaluation. We provide solver-annotated solutions and comprehensively evaluate LLMs by feasibility, solution optimality, and reasoning efficiency. Experiments across a wide range of modern LLMs show that high-performing models achieve strong feasibility and solution quality on small instances, but both degrade as instance size grows, even if more tokens are used for reasoning. We also observe systematic effects across the taxonomy: set-based tasks are relatively easy, whereas graph-structured problems and bottleneck objectives lead to more frequent failures.

\end{abstract}

\input{1_intro}

\input{3_preliminary}
\input{4_method}

\input{5_experiments}

\section{Conclusion}

We introduce NLCO benchmark for assessing LLM reasoning on a broad range of CO tasks. NLCO organizes 43 problems under a four-layer taxonomy and tests LLMs as end-to-end solvers that produce decisions directly from natural-language scenarios. Our empirical study suggests that current LLMs often succeed on small instances, but achieving reliable optimality on larger ones remains challenging, especially for graph-structured and bottleneck objectives, and extra tokens do not fully close the gap. Thus, LLM-based decision support for real CO workflows will require methods that better integrate reasoning over global constraints and deliver more efficient inference. We hope NLCO serves as a reusable testbed to witness more reliable and efficient LLM reasoning for language-based CO.

\clearpage
\section*{Limitations}
Our work has several limitations that point to important directions for future research:

\noindent
\textbf{Scope of tasks.}
Although NLCO covers over 40 CO problems and spans multiple families, it remains a finite sample of the broader landscape of constrained and combinatorial decision-making. Moreover, to keep end-to-end inference tractable for current LLMs, instances are scaled to moderate sizes. As a result, observed performance may not extrapolate to substantially larger instances where classical solvers typically become essential.

\noindent
\textbf{Robustness of natural-language specification.}
NLCO instances are embedded into contextualized textual scenarios, but they are still generated from underlying formal problems with controlled templates and parameterizations. This design improves coverage and reproducibility, yet it may not fully capture the diversity and ambiguity of real-world specifications (e.g., missing constraints, underspecified objectives, or ambiguous requirements). Evaluating robustness to noisy and user-authored problem statements remains an open direction.

\noindent
\textbf{Generalization across prompting/agentic strategies.}
We intentionally evaluate models under a specific prompting and inference protocol. The design choice aims to isolate and assess the inherent reasoning ability of LLMs on CO problems, rather than performance gains introduced by task-specific prompting strategies, tool use, self-verification, or agentic workflows. While alternative prompting strategies or tool-augmented setups may further improve performance, our results reflect controlled end-to-end reasoning ability under a standardized evaluation setting. Building on NLCO, future work can investigate how prompt engineering, tools, or agentic solvers interact with LLM reasoning to address CO problems.


\bibliography{custom}
\clearpage

\input{appendix}

\end{document}

%% file: 1_intro.tex
\section{Introduction}

Recent advances in large language models (LLMs) have performed remarkably across domains such as language understanding~\citep{NEURIPS2024_ad236edc}, programming~\cite{lei2025planning}, and planning~\citep{shao2025personal}.
From arithmetic calculation to strategic decision making, LLMs are increasingly being explored as general-purpose problem solvers \cite{mirzadeh2025gsmsymbolic, singhi2025when, Alkayyal2026}, sparking interest in the evaluation of their limits in reasoning~\citep{Plaat2025}. However, most existing evaluations emphasize relatively simple reasoning competencies, such as arithmetic~\cite{zheng2022miniff}, multi-hop question answering (QA)~\cite{wu2025cofca}, or rule satisfaction~\cite{zhou-etal-2025-rulearena, chen2025lr2bench}, which, while informative, offer limited insight into how well LLMs perform decision making under high-dimensional and constraint settings. Thus, there is a need for benchmarks that move beyond surface-level pattern recognition and probe LLMs by tasks with explicit constraints and objectives~\cite{mondorfbeyond}, so as to systematically assess their potential in more decision-support scenarios.

Combinatorial optimization (CO) offers a rigorous evaluation challenge as it requires reasoning beyond simple arithmetic. Deductive CO reasoning often involves: 1) objective/constraint modeling, 2) constructing discrete decisions (e.g., routes, schedules) that scale with instance size, 3) global consistency checking where a single violation invalidates the solution, and 4) reasoning about solution improvement and near-optimality under a combinatorial landscape. Unlike open-ended dialogue or F1-scored QA tasks, which often reward locally plausible or partially correct outputs,
CO exposes the non-local nature of reasoning. Greedy or myopic solutions may trigger infeasibility or drastic objective degradation~\citep{solomon1987algorithms}. Critically, CO enables two-level evaluation: solutions are first judged by feasibility (as a binary indicator), and feasible ones are then ranked by objective values, yielding a fine-grained criterion.
Since many CO problems are NP-hard, evaluation also naturally involves a quality-efficiency trade-off. Understanding how LLMs perform also matters beyond benchmarking: CO underlie real-world decision making, such as routing, planning, allocation~ \citep{bengio2021machine}, so it also helps gauge LLMs' potential as decision-support agents for realistic scenarios.

In this paper, we introduce NLCO, the \textbf{N}atural \textbf{L}anguage \textbf{C}ombinatorial \textbf{O}ptimization benchmark, which frames CO tasks as contextualized decision making scenarios, and LLMs are evaluated by taking as input textual CO specifications to reason out explicit discrete solutions. The solution feasibility is assessed by problem-specific constraints and optimization quality is measured by an objective. NLCO significantly differs from optimization benchmarks that primarily evaluate code writing, heuristic calling, or external solver invoking~\cite{yang2025optibench, imajuku2025alebench, Sun2025COBench}. 
These indirect prompts resort to solvers, confounding model performance with external search and constraint handling, making it hard to attribute success/failure to the model’s internal reasoning. We deliberately remove the usage of external solvers so that LLMs must reason over the original constrained, combinatorial solution space without tool assistance. This setup exposes failure modes (e.g., constraint violation, missing variable), facilitating diagnostic analysis rather than mere task accuracy.

To support systematic coverage and diagnostic analysis, NLCO tasks are organized using a four-layer taxonomy, which is specified by the following dimensions: variable type, constraint-family group, global constraint pattern(s), and objective class. This taxonomy enables controlled comparisons along structural dimensions (e.g., graph vs.\ set) and supports decomposed (and structural) evaluations by reporting metrics within each dimension.
Within this framework, NLCO comprises 43 problems that span routing, scheduling, packing, etc. Problem instances are drawn from standard CO libraries and real-world scenarios. All instances are scaled to be tractable for LLMs, and rendered as textual scenarios while preserving their original problem structure. The comprehensive and well-processed instances provide a reusable resource for future LLM benchmarking.

\noindent
\textbf{Research Questions.} The end-to-end requirement and four-layer taxonomy in NLCO 
enable systematic investigation on when and why LLMs succeed or fail for CO by internal reasoning capability. Accordingly, we focus on three research questions (RQs): \textbf{RQ1 (End-to-end solving capability)}: To what extent can current LLMs reason over a constrained, combinatorial solution space, solving CO problems end-to-end from textual descriptions? \textbf{RQ2 (Problem structure-aware reliability)}: How does problem structure (e.g., variable types, constraint families) influence reasoning reliability, and which, among them, primarily drive infeasible or suboptimal solutions? \textbf{RQ3 (Quality-efficiency trade-off)}:  How well can LLMs balance reasoning efficacy (e.g., optimality) against efficiency (e.g., token usage) when solving NP-hard CO problems? We answer these RQs using the NLCO benchmark to probe the limits of LLMs reasoning with real-world constraints and combinatorics, and to assess their reliability as decision-support agents under verifiable constraints and resource budgets.

\paragraph{Related Work and Positioning.} 
\begin{table}[h]
\centering
\small
\setlength{\tabcolsep}{4.5pt}
\renewcommand{\arraystretch}{1}
\resizebox{\columnwidth}{!}{
\begin{tabular}{p{4.5cm}p{0.5cm}p{0.9cm}p{0.5cm}p{0.5cm}p{2.8cm}}
\toprule
\textbf{Benchmark} 
& \textbf{CO} 
& \textbf{\#Prob.} 
& \textbf{E2E} 
& \textbf{Ctx.} 
& \textbf{Taxonomy} \\
\midrule
FrontierCO~\cite{feng2025comprehensive} 
& \textcolor{green!60!black}{\ding{51}} 
& 8 
& \textcolor{red!70!black}{\ding{55}} 
& \textcolor{red!70!black}{\ding{55}} 
& Application domain \\
\midrule
CO-Bench~\cite{Sun2025COBench} 
& \textcolor{green!60!black}{\ding{51}} 
& 36 
& \textcolor{red!70!black}{\ding{55}} 
& \textcolor{red!70!black}{\ding{55}} 
& Application domain \\
\midrule
CP-Bench~\cite{michailidis2025cp} 
& \textcolor{red!70!black}{\ding{55}} 
& N/A 
& \textcolor{red!70!black}{\ding{55}} 
& \textcolor{red!70!black}{\ding{55}} 
& N/A  \\
\midrule
NPHardEval~\cite{fan2024nphardeval} 
& \textcolor{red!70!black}{\ding{55}} 
& 5 
& \textcolor{green!60!black}{\ding{51}} 
& \textcolor{red!70!black}{\ding{55}} 
& Complexity class \\
\midrule
GraphArena~\cite{tang2025grapharena} 
& \textcolor{red!70!black}{\ding{55}} 
& 6 
& \textcolor{green!60!black}{\ding{51}} 
& \textcolor{red!70!black}{\ding{55}} 
& Complexity class \\
\midrule
\rowcolor{black!8}
\textbf{NLCO (this work)} 
& \textbf{\textcolor{green!60!black}{\ding{51}}} 
& \textbf{43} 
& \textbf{\textcolor{green!60!black}{\ding{51}}} 
& \textbf{\textcolor{green!60!black}{\ding{51}}} 
& \textbf{Four-layer taxonomy} \\
\bottomrule
\end{tabular}
}
\caption{
Comparison of NLCO with existing benchmarks.
\textbf{CO} denotes whether a benchmark is primarily CO-oriented; 
\textbf{\#Prob.} is the number of distinct CO tasks covered; 
\textbf{E2E} (End-to-end) specifies whether LLMs must produce solutions directly from textual input; 
\textbf{Ctx.} (Contextualization) marks whether instances are contextualized in natural-language scenarios (vs. fixed formats); 
\textbf{Taxonomy} reports how the benchmark organizes tasks for problem selection or structural evaluation.
}
\label{tab:benchmark-comparison}
\end{table}
LLM reasoning benchmarks have rapidly expanded, from math word problems~\cite{cobbe2021training, uesato2022solving} to logic reasoning~\cite{liu2020logiqa, zhu2025autologi} and commonsense reasoning~\cite{roewer2025accord}. Our work targets a more distinct challenge with complex CO tackled by end-to-end reasoning. As summarized in Table~\ref{tab:benchmark-comparison}, related work largely fall into two complementary lines: 1) Optimization-centric suites, such as FrontierCO~\cite{feng2025comprehensive}, CO-Bench~\citep{Sun2025COBench}, and CP-Bench~\citep{michailidis2025cp}, often assess LLMs as interfaces that produce code for heuristics or solver calls, making it hard to isolate internal end-to-end solving capability, diagnose reasoning failures in a structural manner, and measure reasoning efficiency; 2) Direct-output benchmarks, such as NPHardEval~\citep{fan2024nphardeval} and GraphArena~\citep{tang2025grapharena}, require solution generation but cover few problem types. By organizing tasks mainly by complexity class, they are limited in structural analysis and broad comparisons. NLCO bridges these gaps: it unifies 40+ CO tasks in contextualized scenarios under a four-layer taxonomy, and forces end-to-end solving, enabling controlled analyses of problem structure-aware reliability and quality-efficiency trade-offs. An extended related work on 1) LLMs for optimization, 2) LLM reasoning evaluations, and 3) optimization benchmark datasets is elaborated in Appendix~\ref{app:literature}.

%% file: 3_preliminary.tex
\section{Preliminaries}

\subsection{Combinatorial Optimization}
CO involves finding an optimal configuration of discrete variables that satisfies a set of constraints while minimizing (or maximizing) an objective. Formally, a CO problem can be defined as:
\begin{equation}
    \min_{x \in \mathcal{X}} f(x) \quad \text{s.t.} \quad x \in \mathcal{F},
\end{equation}
where $\mathcal{X}$ is the discrete search space, $f(x)$ denotes the objective function, and $\mathcal{F} \subseteq \mathcal{X}$ encodes the feasible set satisfying all problem-specific constraints.


\subsection{Problem Statement}

In the context of LLM-based reasoning, we consider CO problems expressed entirely in natural language. Each instance consists of: 

\noindent
\textbf{Input Description ($D$)}: A natural-language prompt specifying a CO problem description (i.e., an instance), explicitly or implicitly implying the variables, constraints, and optimization objectives, with an example illustrated in Figure~\ref{fig:main}(a). 

\noindent
\textbf{Underlying Formal Model ($P = \langle \mathcal{X}, \mathcal{F}, f \rangle$)}: A hidden mathematical formulation associated with $D$. This model is used only for solution checking and scoring, and is never given to the LLM.

\noindent
\textbf{Solution Representation ($S$)}: 
A structured solution \(x^*\) is defined as the one that is feasible (\(x^* \in \mathcal{F}\)) and optimal for \(f(x)\). It serves as the reference label for the problem instance.

Given an instance, the LLM receives only the text and output a solution $\hat{x}$, which is supposed to align with the formal constraints of $P$. Let $\text{Eval}(\cdot)$ be a function that measures both feasibility violations and solution optimality. At a high level, the reasoning objective can be written as $\hat{x} = \arg\min_{x \in \mathcal{X}} \mathbb{E}_{\text{prompt}(D)} [\text{Eval}(x, \mathcal{F}, f)]$, where we assume infeasibility violation and objective function are minimized.

%% file: 4_method.tex
\section{NLCO Benchmark}
\label{sec:method}

\begin{figure*}[t]
  \centering

  \begin{subfigure}[t]{\linewidth}
    \centering
    \includegraphics[width=\linewidth]{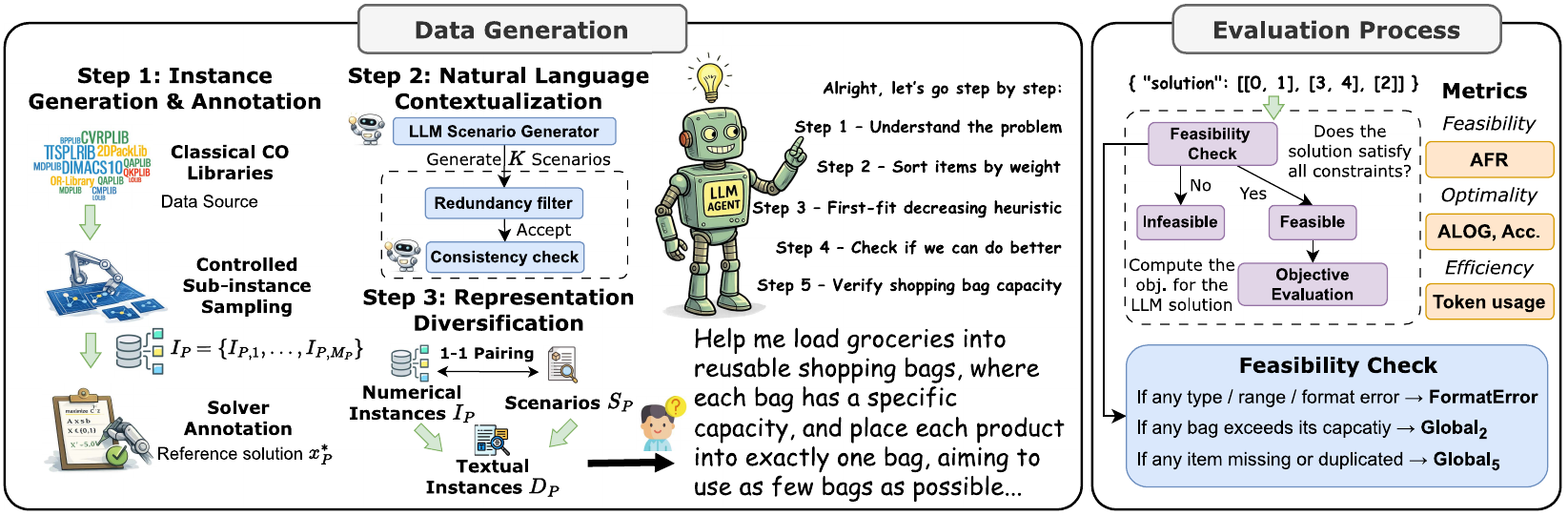}
    \caption{NLCO data generation and evaluation process (bin packing shown as an example).}
  \end{subfigure}
  \hfill
  \begin{subfigure}[t]{\linewidth}
    \centering
    \includegraphics[width=\linewidth]{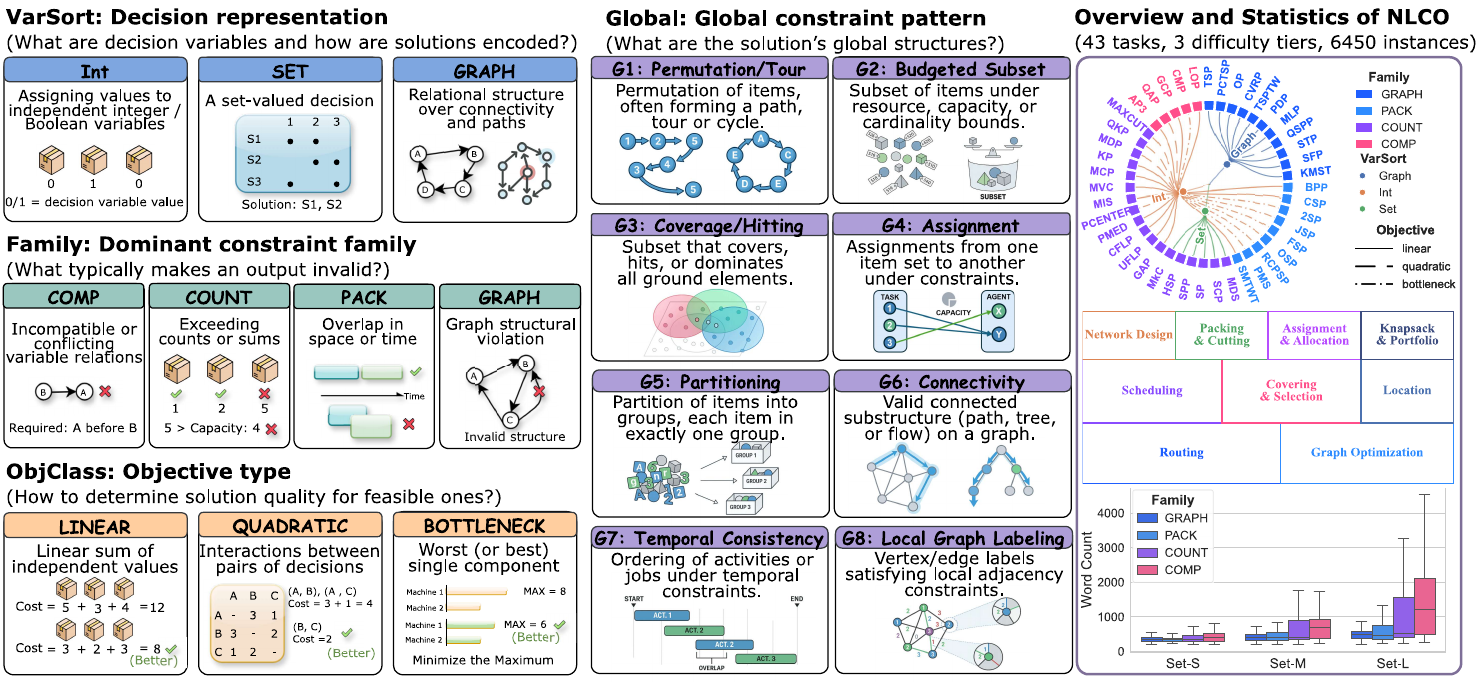}
    \caption{NLCO four-layer taxonomy and dataset overview.}
  \end{subfigure}

  \caption{Overview of the NLCO benchmark.}
  \label{fig:main}
\end{figure*}

NLCO benchmark features the first large-scale dataset for end-to-end natural language CO reasoning. More than 40 CO problems are organized under a novel taxonomy that captures diverse decision domains, constraint structures, and objective classes, as shown in Figure~\ref{fig:main}(b). For each problem, NLCO provides instances at three difficulty levels. Detailed criteria for difficulty levels of each problem are provided in Table~\ref{tab:detailed_summary}.

\subsection{Benchmark Construction}

NLCO is constructed in four steps, including 1) Taxonomy Design (Section~\ref{sec:taxo}): we propose a four-layer taxonomy as a unified design space for diverse CO tasks; 2) Task Selection (Section~\ref{sec:ts}): we specify a set of diverse CO problems according to taxonomy, covering heterogeneous variable domains, constraint families, and objective classes; 3) Data Generation (Section~\ref{sec:dg}): we synthesize CO instances and contextualize them in natural language; 4) Evaluation Process (Section~\ref{sec:ep}): we specify solution checking protocols and metrics to gauge solution quality and LLM reasoning ability.

\subsubsection{Taxonomy Design}
\label{sec:taxo}
We propose a four-layer taxonomy to guide task selection and diagnose reasoning challenges across different CO problem structures.
Our taxonomy, shown in Figure~\ref{fig:main}(b), builds on concepts from the constraint programming (CP), which has long studied how diverse CO problems can be expressed using a set of reusable constraint templates. We ground our taxonomy in the XCSP$^3$ specification \cite{boussemart2016xcsp3}, a widely used standard for combinatorial constraint modeling. We adopts it as a core principle to group CO tasks into structurally coherent families used to specify comprehensive and diverse aspects for LLM reasoning.
For each $P=\langle \mathcal{X}, \mathcal{F}, f\rangle$, we assign a tuple:
\begin{multline}
    \tau(P) = (\text{VarSort}(P), \text{Family}(P), \\ \text{Global}(P), \text{ObjClass}(P)),
\end{multline}
where each axis captures one aspect of the underlying CP model of a CO task, as described below.

\noindent
\textbf{Variable Sort} ($\text{VarSort}$): the primary decision domain used in a canonical CP formulation, drawn from \textsc{Int} (integer / Boolean variables), \textsc{Set} (set-valued variables), and \textsc{Graph} (graph-structured variables). These sorts correspond to algebraic and set-theoretic reasoning, as well as reasoning over relational structures (e.g., paths and connectivity).

\noindent
\textbf{Constraint-family Group} ($\text{Family}$): the dominant structural family governing feasibility, such as \textsc{Comp} (comparison constraints such as orderings and inequalities);
\textsc{Count} (counting and summation constraints);
\textsc{Pack} (packing and scheduling constraints); and
\textsc{Graph} (graph-defined constraints encoded over discrete variables).

\noindent
\textbf{Global Constraint Pattern(s)} ($\text{Global}$):
the global constraints that a valid solution must satisfy.
Instead of recording all fine-grained CP global constraints (e.g., \gconst{allDifferent}), which reflect solver-level modeling choices but are overly detailed for analyzing LLM behavior, we group them into a small set of patterns used to summarize feasible solutions.
Concretely, we use eight patterns, such as Permutation/Tour, Budgeted Subset, etc, which are presented in the middle part of Figure~\ref{fig:main}(b); each CO task is associated with one or more patterns, indicating which global structural requirements are central to its feasibility. We detail formal definitions and the mapping from classical CP global constraints to patterns in Appendix~\ref{app:taxonomy}.

\(\text{Global}(P)\) plays a dual role in NLCO.
On the modeling side, it groups standard CP global constraints into shared structural types (e.g., mapping \gconst{circuit} and \gconst{path} to permutation/tour), keeping a direct link to canonical formulations.
On the evaluation side, it determines pattern-specific feasibility checks for LLM outputs. For the bin packing example in Figure~\ref{fig:main}(a), we parse the output as a nested list and flag infeasibility if the solution is malformed, exceeds any bin’s capacity, or misses/duplicates items.
This pattern-aware validation enables fine-grained diagnosis. In particular, by recording which patterns are violated, we can pinpoint global structures (e.g., partitioning or connectivity) that LLMs most often fail to satisfy.

\noindent
\textbf{Objective Class} ($\text{ObjClass}$): the cost or utility functions, grouped into linear, quadratic, and bottleneck (min–max / max–min) objectives.

We stress that the taxonomy is a coarse but interpretable abstraction rather than a complete classification of CO problems. Details of the taxonomy and tasks under $\tau(\cdot)$ are provided in Appendix~\ref{app:taxonomy}.


\subsubsection{Task Selection}
\label{sec:ts}
In the absence of a universally accepted “gold-standard’’ task set, we treat $\tau(\cdot)$ as a design space and instantiate the task suite to ensure broad coverage of decision domains and constraint structures.

\noindent
\textbf{Decision Domain Coverage.} We ensure that all three variable types in $\text{VarSort}$ are integrated into different CO tasks, 
so that evaluation can be conducted across heterogeneous decision domains, comparing their effects in LLM reasoning.

\noindent
\textbf{Constraint Coverage.} Meanwhile, we target a wide range of CO tasks with various constraints. Within each constraint family (\textsc{Comp}, \textsc{Count}, \textsc{Pack}, \textsc{Graph}), we include multiple CO tasks whose CP formulations pertain to different global constraint patterns. On top of that, we can still specify tasks by their applications (routing, scheduling, etc.). Such taxonomy-based task selections favorably organize the task suite, so as to readily diagnose LLM reasoning from distinct dimensions (variable type, constraint family, etc.) 

\noindent
\textbf{Difficulty and LLM Solvability.}
Since complexity of NP-hard CO often exponentially grows against problem sizes, we offer a natural scalability axis: increasing the problem size expands the combinatorial space and thus yields harder reasoning challenges. We scale each problem into three tiers: Set-S, Set-M, and Set-L, based on its size-related parameters and structural factors (e.g., number of nodes, graph density). Intuitively, Set-S instances are small enough so that the solution space could be sufficiently explored; Set-M typically requires heuristic reasoning or stochastic exploration for performance enhancement; Set-L further enlarges the combinatorial solution space, making feasibility and efficiency by reasoning more intractable. We generate 50 instances per tier for each CO task.

With the above criteria, NLCO covers 43 tasks, 3 difficulty tiers, and 6450 instances, spanning domains of network design, packing, cutting, assignment, allocation, knapsack, portfolio, scheduling, covering, selection, location, routing, and graph optimization problems, as shown in the right panel of Figure~\ref{fig:main}(b). The word count for textual instances ranges from 190 to 10022, indicating difficulties of language understanding and reasoning. More taxonomy details are offered in Table~\ref{tab:cop-taxonomy} of Appendix~\ref{app:taxonomy}.

\subsubsection{Data Generation}
\label{sec:dg}

As shown in Figure~\ref{fig:main}(a), the data generation process encompasses three steps, including 1) instance generation and annotation, 2) natural-language contextualization, and 3) representation diversification.

\noindent
\textbf{Instance Generation and Solver Annotation.}
Instead of synthesizing CO instances using specific data distributions (e.g., uniform distributions)~\cite{kool2018attention, fan2024nphardeval, hua2026uspr}, we construct our dataset by extracting and adapting instances from well-established optimization libraries (e.g., TSPLIB~\cite{reinelt1991tsplib}) that are sourced from realistic decision-making scenarios. As such, we aim to obtain a set of numerical instances $\mathcal{I}_P = \{ I_{P,1}, I_{P,2}, \dots, I_{P,M_P} \}$ for each CO task $P$.
CO instances in libraries are often defined at scales (e.g., with hundreds of nodes) beyond the feasibility of LLM reasoning (e.g.,exceeding the model’s context budget or inducing an intractably large search space). Hence, we generate sub-instances $I_{P,j}$ by controlled sampling (e.g., randomly sampling 20 nodes from a 200-node graph). This empowers different levels of tractability for end-to-end reasoning but also preserves the characteristics of the original CO tasks.

For each generated instance $I_{P,j} \in \mathcal{I}_P$, we utilize a domain-specific solver to obtain its (near-)optimal solution $x^*_{P,j}$. This solution is stored as the reference label associated with a task $P = \langle \mathcal{X}, \mathcal{F}, f \rangle$, and is used to assess reasoning results. More details of instance generation and solver annotation for each task are specified in Appendix~\ref{sec:problem details}.

\noindent
\textbf{Natural-language Contextualization.}
After generating and annotating numerical CO instances, we further contextualize each instance into a natural language description that embeds the problem within real-world decision-making scenarios (e.g., logistics planning or task scheduling). Similar to the work on dialogue generation~\cite{finch2024diverse} and optimization problem generation~\cite{huang2025orlm}, we begin by generating a set of scenarios through calling an LLM, and employ an iterative generate-filter-verify procedure for better diversity and accuracy, as detailed below.

Given a problem $P$, our goal is to construct a set of linguistically diverse yet semantically equivalent scenarios
$\mathcal{S}_P = \{ s_{P,1}, \dots, s_{P,N_P} \}$,
where each scenario gives an everyday narrative of $P$ while preserving its objective and constraints. To this end, in each iteration, an LLM proposes $K$ candidate scenarios for $P$, which are subsequently verified and filtered to enforce correctness and diversity.

\noindent
\emph{Semantic Redundancy Filtering.}
To promote scenario diversity and avoid near-duplicate scenarios, we apply an embedding-based novelty criterion.
We encode all textual scenarios using a SentenceTransformer (\texttt{all-MiniLM-L6-v2}), which produces an embedding $\mathbf{e}(\cdot)$ for each scenario.
For each new candidate $c$, we compute its cosine similarity with each previously accepted scenario $s \in \mathcal{S}_P$.
If $\max_{s \in \mathcal{S}_P} \cos\big(\mathbf{e}(c), \mathbf{e}(s)\big) > 0.7$, the candidate is discarded as a semantic paraphrase.
\noindent
\emph{Problem-Consistency Verification.}
The semantically diverse candidate is further verified by an LLM-based checker conditioned on the formal specification of $P$. By doing so, we ensure that the scenario still preserves the original problem structure without altering the CO task.

The generate-filter-verify procedure continues till $|\mathcal{S}_P| \ge M_P$.
Finally, for each numerical instance $I_{P,j} \in \mathcal{I}_P$, we associate it with exactly one scenario
$s_{P,j} \in \mathcal{S}_P$,
resulting in a textual instance
$d_{P,j} = (I_{P,j}, s_{P,j})$.
This procedure ensures that all instances are contextualized by coherent and semantically valid natural-language descriptions.

\noindent
\textbf{Representation Diversification.}
To reflect variation in user input, we randomly sample $d_{P,j}$ to create instance variants that change only data presentation while preserving the same optimization instance. Concretely, we apply simple transformations to $d_{P,j}$: 
1) \textbf{Format changes}, presenting the same data inputs in natural language, CSV, JSON, or Markdown table; and 
2) \textbf{Indexing changes}, using different ways to label variables, such as number-based indices or alphabetical labels. These diverse inputs further involve influence of different formats in LLM reasoning, which is seldom considered in current benchmarks.
Further implementation details, used prompts, and examples are provided in Appendix \ref{sec:appendix-context}.







\subsection{Evaluation Process}
\label{sec:ep}

As shown in Figure~\ref{fig:main}(a), we evaluate LLM solutions along 3 dimensions: feasibility, optimality, and inference efficiency. Given a textual instance $d_{P,i}$ of a CO problem $P=\langle \mathcal{X},\mathcal{F},f\rangle$, the model produces a textual solution, which we parse into a numerical decision $\hat{x}_{P,i}\in\mathcal{X}$. We assess feasibility by checking $\hat{x}_{P,i}$ against $\mathcal{F}$ with global-pattern validation. Unparseable outputs or constraint violations are deemed infeasible. For feasible ones, we compute their objectives $f(\hat{x}_{P,i})$ for comparison.

\noindent
\textbf{Core Metrics.}
We report four primary metrics.
1) \textbf{Average feasibility rate (AFR)}: fraction of instances, for which the LLM output is feasible (i.e., it meets all constraints).
2) \textbf{Average accuracy (Acc.)}: fraction of instances, for which the output is feasible and achieves the reference objective, i.e., $f(\hat{x}_{P,i})=f^*_{P,i}$ (within a small tolerance).
3) \textbf{Average log optimality gap (ALOG)}: among feasible instances, we measure the gap of objective values to the optimum. The gaps are averaged in log-space to reduce the influence of heavy-tailed errors. We compute the relative gap $\mathrm{Gap}_{P,i}=\frac{f(\hat{x}_{P,i})-f^*_{P,i}}{\max\{|f^*_{P,i}|,\epsilon\}}$ and report $\log(1+\mathrm{Gap})$ averaged over feasible instances.
4) \textbf{Token usage}: we track average output tokens as a simple proxy for LLM reasoning cost.

%% file: 5_experiments.tex
\section{Experiments}
\label{sec:exp}

\newcommand{\legendbox}[1]{%
  \raisebox{0.2ex}{\colorbox{#1}{\rule{0pt}{0.8ex}\hspace{0.6em}}}%
}

\begin{table*}[t]
\centering
\small
\setlength{\tabcolsep}{3.5pt}
\renewcommand{\arraystretch}{1}
\resizebox{\textwidth}{!}{
\begin{tabular}{lcccc cccc cccc}
\toprule
\multirow{2}{*}{\textbf{Model}} 
& \multicolumn{4}{c}{\textbf{Set-S}} 
& \multicolumn{4}{c}{\textbf{Set-M}} 
& \multicolumn{4}{c}{\textbf{Set-L}} \\
\cmidrule(lr){2-5}\cmidrule(lr){6-9}\cmidrule(lr){10-13}
& \textbf{AFR} $\uparrow$ & \textbf{Acc.} $\uparrow$ & \textbf{ALOG} $\downarrow$ & \textbf{tok.} $\downarrow$
& \textbf{AFR} $\uparrow$ & \textbf{Acc.} $\uparrow$ & \textbf{ALOG} $\downarrow$ & \textbf{tok.} $\downarrow$
& \textbf{AFR} $\uparrow$ & \textbf{Acc.} $\uparrow$ & \textbf{ALOG} $\downarrow$ & \textbf{tok.} $\downarrow$ \\
\midrule

\multicolumn{10}{l}{\textbf{Open-weight Models}} \\
\midrule
Qwen3-14B & 70.4 & 38.7 & 0.1734 & 2492 & 58.1 & 17.5 & 0.5359 & 3053 & 48.9 & 7.8 & 0.7255 & 3328 \\
Ministral-3-14B & 72.0 & 47.7 & 0.0979 & 10326 & 58.7 & 25.4 & 0.2656 & 14325 & 46.6 & 12.0 & 0.4292 & 15643 \\
Nemotron3-Nano-30B & 64.2 & 26.8 & 0.2627 & 2208 & 50.4 & 9.6 & 0.6302 & 3663 & 37.4 & 4.0 & 0.8820 & 4283 \\
Llama-4-Maverick-Instruct & 78.8 & 42.8 & 0.1467 & 1471 & 65.1 & 19.0 & 0.3111 & 1730 & 52.7 & 7.9 & 0.4314 & 2009 \\
Qwen3-235B-Instruct & 95.3 & 80.1 & 0.0248 & 6701 & 87.6 & 54.9 & 0.0933 & 8848 & 79.7 & 36.3 & 0.2207 & 9988 \\
DeepSeek-V3.2 & 93.4 & 72.1 & 0.0427 & 2613 & 86.1 & 47.7 & 0.1466 & 3510 & 78.5 & 31.1 & 0.3361 & 4019 \\
MiMo-V2-Flash & 75.0 & 58.4 & 0.0711 & 13733 & 62.8 & 37.4 & 0.1732 & 21751 & 54.3 & 24.7 & 0.2763 & 25501 \\
Qwen3-14B (reasoning) & 90.7 & 69.9 & 0.0427 & 8929 & 79.1 & 42.4 & 0.1834 & 11576 & 69.4 & 25.5 & 0.3298 & 12351 \\
Nemotron3-Nano-30B (reasoning) & 94.2 & 80.1 & 0.0360 & 11708 & 88.2 & 54.2 & 0.1328 & 21797 & 79.7 & 38.2 & 0.3355 & 26449 \\
QwQ-32B (reasoning) & 89.9 & 68.5 & 0.0533 & 11484 & 73.8 & 39.1 & 0.1998 & 14541 & 61.1 & 19.6 & 0.3517 & 15472 \\
DeepSeek-V3.2 (reasoning) 
& \cellcolor{OpenBest}98.7 & \cellcolor{OpenBest}88.2 & \cellcolor{OpenBest}0.0071 & 9262
& \cellcolor{OpenBest}96.8 & \cellcolor{OpenBest}69.1 & \cellcolor{OpenBest}0.0403 & 15484
& \cellcolor{OpenBest}92.9 & \cellcolor{OpenBest}54.1 & \cellcolor{OpenBest}0.1238 & 19403 \\

\midrule
\multicolumn{10}{l}{\textbf{Proprietary Models}} \\
\midrule
Grok-4.1-Fast (reasoning)
& 98.3 & 88.7 & 0.0067 & 6582
& 93.6 & 70.9 & 0.0312 & 12389
& 83.7 & 53.4 & 0.1441 & 16641 \\
Claude-Sonnet-4.5 (reasoning) 
& 98.7 & 89.0 & 0.0061 & 10755
& 96.7 & 70.4 & 0.0393 & 16802
& 93.2 & 53.0 & 0.1282 & 20590 \\
OpenAI o4-mini (reasoning) 
& \cellcolor{ClosedBest}98.9 & 89.3 & 0.0058 & 9544
& 96.5 & 70.7 & 0.0471 & 19118
& 92.5 & 53.1 & 0.2102 & 23426 \\
OpenAI GPT-5.1 (reasoning) 
& \cellcolor{ClosedBest}98.9 & 90.9 & \cellcolor{ClosedBest}0.0056 & 6573
& \cellcolor{ClosedBest}98.0 & 72.8 & 0.0462 & 12823
& \cellcolor{ClosedBest}95.5 & 57.2 & 0.1716 & 16646 \\
Gemini-3-Flash (reasoning) 
& 98.8 & \cellcolor{ClosedBest}91.6 & 0.0068 & 16993
& 97.5 & \cellcolor{ClosedBest}76.8 & \cellcolor{ClosedBest}0.0203 & 24325
& 95.4 & \cellcolor{ClosedBest}60.9 & \cellcolor{ClosedBest}0.0899 & 27700 \\

\bottomrule
\end{tabular}
}
\caption{Performance on NLCO across three difficulty tiers. AFR and Acc. are reported in percentages (\%).
\quad \textbf{Legend:} \legendbox{OpenBest}: Best results achieved by open-weight LLMs;\;
\legendbox{ClosedBest}: Best results achieved by proprietary LLMs.}
\label{tab:nlco_main_results}
\end{table*}
\subsection{Experimental Setup}

\begin{figure*}[t]
    \centering
    \begin{subfigure}[t]{0.592\textwidth}
        \centering
        \includegraphics[width=\linewidth]{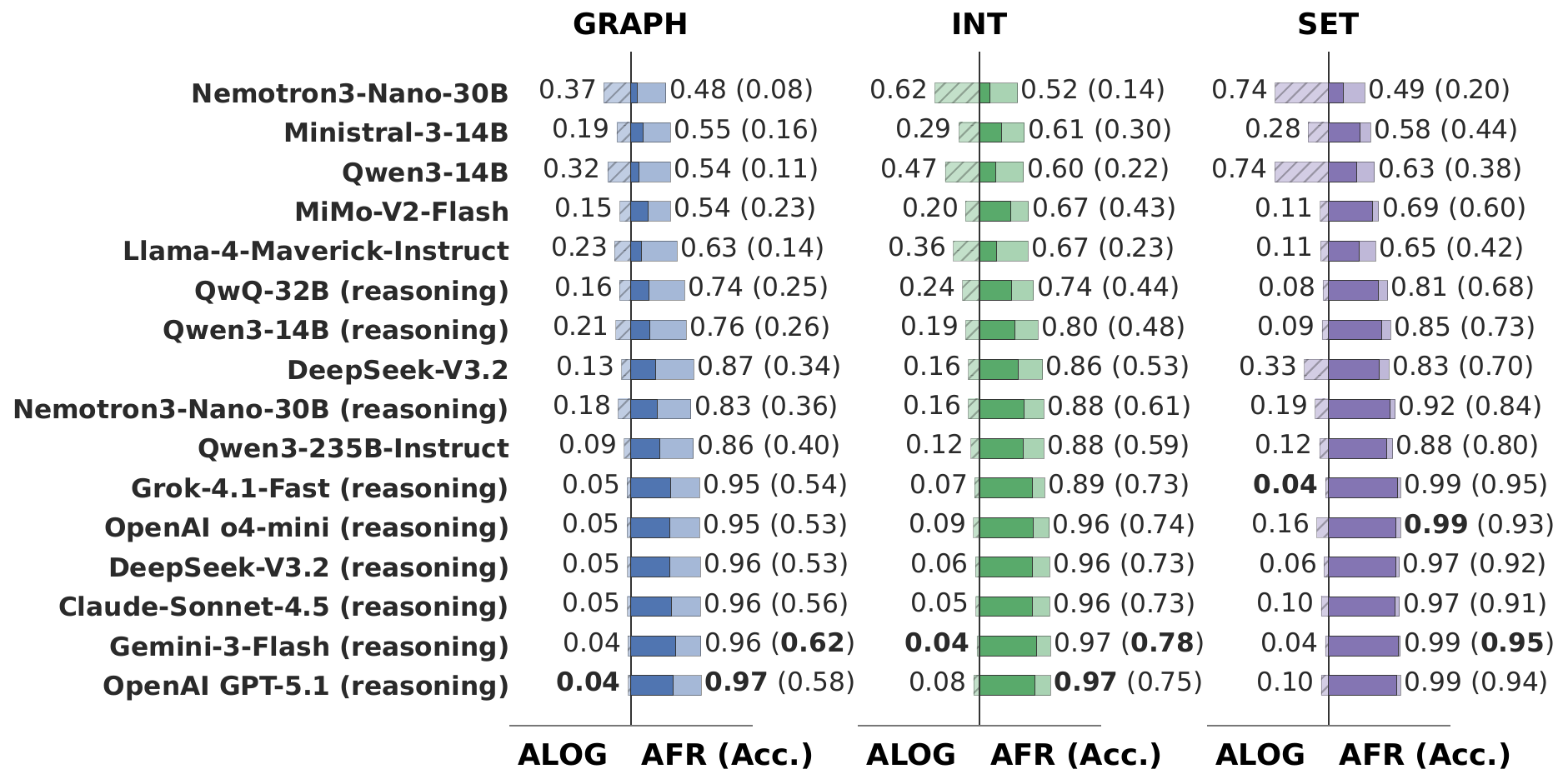}
        \caption{}
    \end{subfigure}
    \hfill
    \begin{subfigure}[t]{0.401\textwidth}
        \centering
        \includegraphics[width=\linewidth]{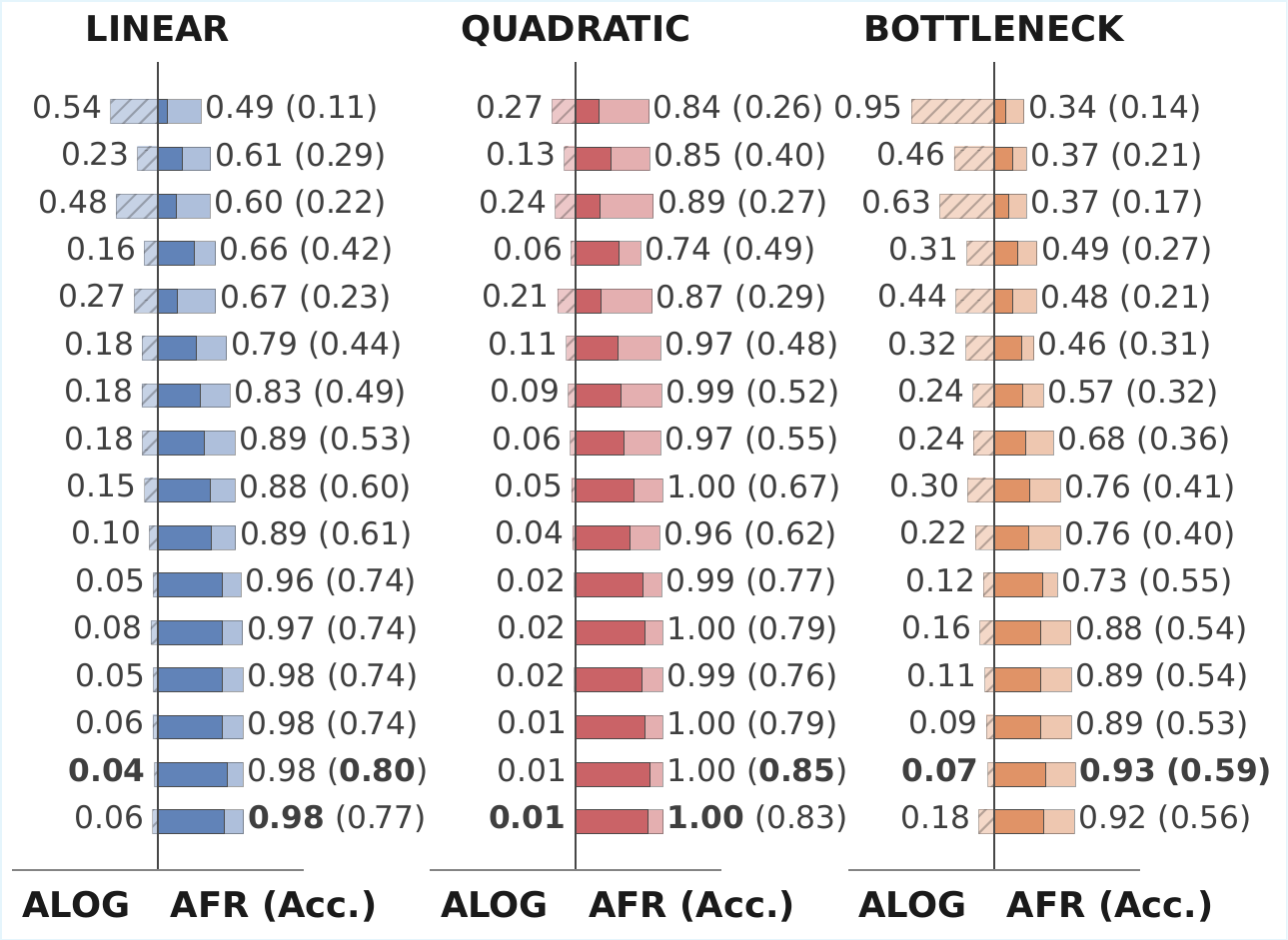}
        \caption{}
    \end{subfigure}
\caption{Aggregated performance across (a) $\text{VarSort}$ and (b) $\text{ObjClass}$ dimensions in NLCO taxonomy.}
    \label{fig:nlco_group}
\end{figure*}

We use NLCO to evaluate eight \textbf{open-weighted LLMs: }Qwen3-14B~\citep{yang2025qwen3}, Ministral-3-14B~\citep{mistral3_14b}, NVIDIA Nemotron3-Nano-30B-A3B~\citep{nvidia_report_2025}, QwQ-32B~\citep{qwen2025qwq32b}, Llama-4-Maverick (17Bx128E)-Instruct ~\citep{meta_llama4_maverick}, Qwen3-235B-A22B-Instruct-2507~\citep{yang2025qwen3}, Xiaomi MiMo-V2-Flash~\citep{mimo2025flash}, DeepSeek-V3.2~\citep{liu2025deepseek}, and five \textbf{proprietary LLMs:} OpenAI o4-mini~\citep{openai_o4mini_system}, GPT-5.1~\citep{openai_gpt51_2025}, Anthropic Claude-Sonnet-4.5~\citep{anthropic_claude_sonnet45}, xAI Grok-4.1-Fast~\citep{xai_grok41_fast_2025}, and Gemini-3-Flash~\cite{gemini-3-flash}. We cover both standard chat LLMs and explicit reasoning LLMs. For models that offer separate the two variants, we use notation '(reasoning)' to denote its reasoning version. Detailed experiment configurations are provided in Appendix~\ref{sec:config}.

\subsection{Overall Performance}

To answer \textbf{RQ1} on \textbf{end-to-end capability}, Table~\ref{tab:nlco_main_results} reports NLCO results, evaluating feasibility (AFR), optimality (Acc.\ and ALOG), and inference cost (tok.). Unlike prior reports that LLMs rarely solve CO~\cite{fan2024nphardeval, tang2025grapharena}, we find that frontier LLMs can often translate textual specifications into globally feasible decisions: on Set-S, the best models reach near-saturated feasibility (up to \textbf{98.9\%} AFR) and achieve high exact-optimality rates (up to \textbf{91.6\%} Acc.), demonstrating strong end-to-end reasoning on small CO instances. 
However, this capability does not scale smoothly with problem size. As instances grow from Set-S to Set-L, feasibility and optimality deteriorate sharply across models (AFR/Acc.\ decrease and ALOG increases), while token usage increases consistently. 
This coupled trend indicates that larger combinatorial search spaces still pose a substantial challenge for LLMs: models tend to expend more inference-time computation yet are increasingly unable to maintain feasibility and reach optimal solutions.


Smaller LLMs generally struggle to solve CO problems reliably even on easier tiers: Nemotron3-Nano-30B attains only 64.2\%/50.4\%/37.4\% AFR on Set-S/M/L with low Acc. and relatively large ALOG. Similarly, Qwen3-14B drops from 70.4\% AFR on Set-S to 48.9\% on Set-L, while Acc. falls from 38.7\% to 7.8\%. The results suggest that limited model capacity substantially constrains CO reasoning, especially as the combinatorial space grows. We further evaluate LLMs by performance profiles \cite{dolan2002benchmarking}, and report average per-task results, as presented in Appendix~\ref{sec:app_res}.

\subsection{Aggregated Analysis}

To answer \textbf{RQ2} on \textbf{structure-aware reliability}, we aggregate results across taxonomy dimensions. As shown in Figure~\ref{fig:nlco_group}, model performance exhibits heterogeneity across variable types ($\text{VarSort}$)  and objective classes ($\text{ObjClass}$), indicating that failures are not uniform but tied to problem structure.

\noindent
\textbf{Impact of VarSort and ObjClass.}
Across frontier LLMs, \textsc{Set} tasks are the easiest, with high feasibility and markedly higher exact-optimality than other sorts. In comparison, \textsc{Graph} problems are more error-prone, likely because they require maintaining global relational consistency rather than making relatively local, enumerable choices.
Objective form also induces performance swings. Linear and quadratic objectives are reliable: frontier LLMs reach high Acc. (e.g., GPT-5.1: 77\% and 83\%). Notably, bottleneck tasks remain challenging, e.g., GPT-5.1 drops to 56\% Acc. The consistent degradation suggests a characteristic failure mode: when the objective is dominated by a worst-case component, local errors or incomplete global verification can disproportionately harm optimality, and models struggle to identify the optimal solution reliably.

\begin{figure*}[thb]
    \centering
        \includegraphics[width=0.35\linewidth]{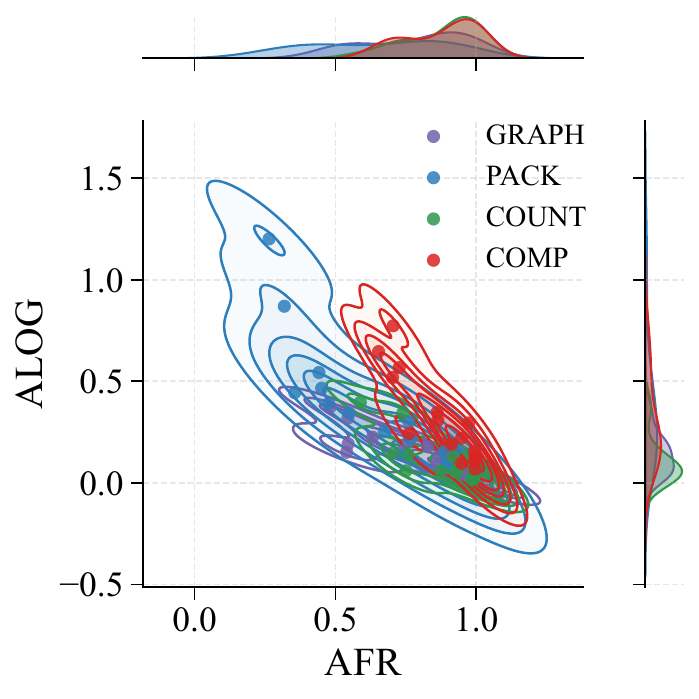}
        \includegraphics[width=0.35\linewidth]{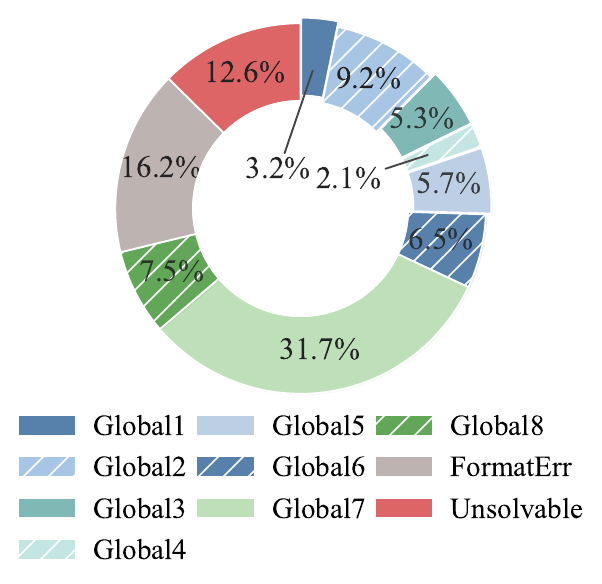}
        \captionof{figure}{AFR–ALOG joint distribution by family (left) and infeasibility mode distribution (right). \emph{FormatError} is raised when the solution cannot be parsed with missing fields or non-finite numbers. \emph{Unsolvable} denotes instances that remain unsolved under the token limit.}
        \label{fig:constraint_gropued}

\end{figure*}

\begin{figure*}[thb]
        \centering
        \includegraphics[width=0.79\linewidth]{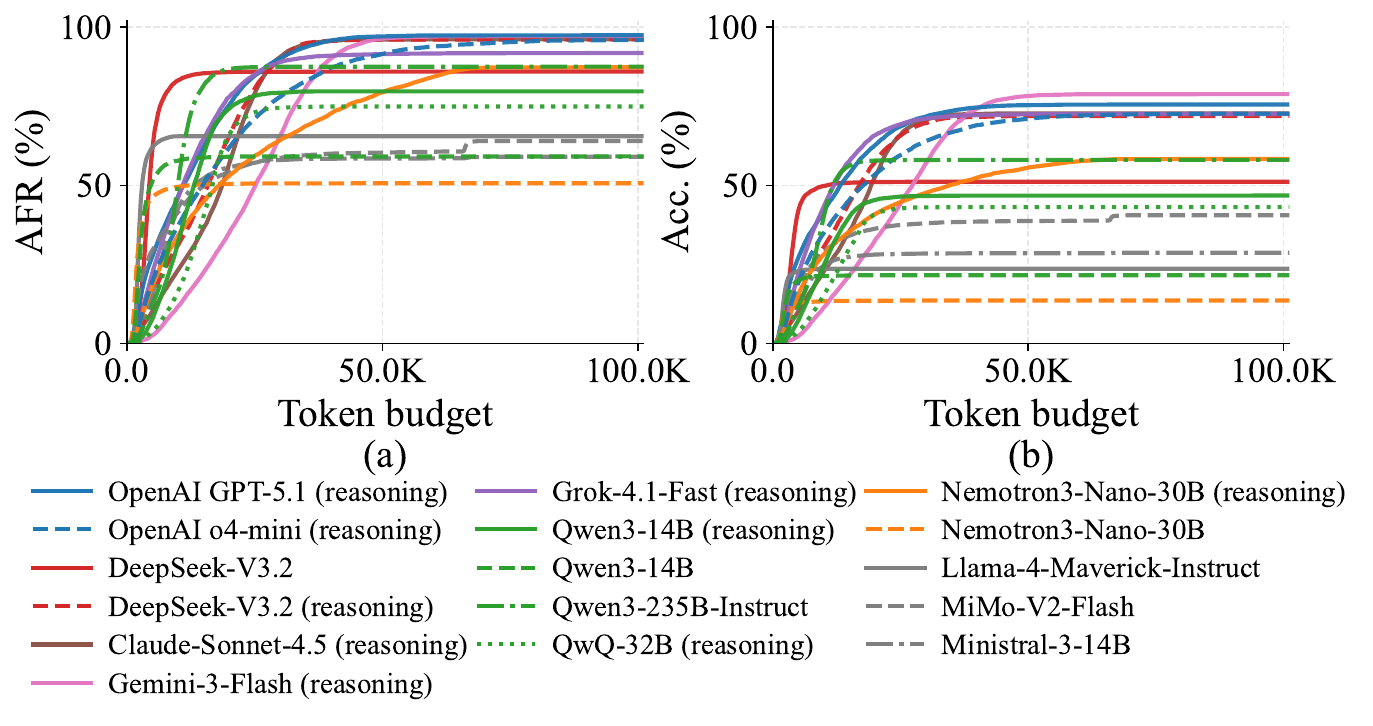}
        \captionof{figure}{LLM data profiles across all difficulty tiers. (a) AFR vs. Token budget; (b) Acc. vs. Token budget.}
        \label{fig:data_profile}
\end{figure*}

\noindent
\textbf{Constraint Families and Global Patterns.}
To better diagnose the drivers of infeasibility and suboptimality, we pool outputs from all LLMs and break down performance by \text{Family} and global patterns. Figure~\ref{fig:constraint_gropued} (left) plots the joint distribution of AFR and ALOG across families, revealing structure dependence: \textsc{Pack} exhibits the largest variance and the heaviest failure tail (low AFR with high ALOG), whereas \textsc{Count} and \textsc{Graph} concentrate near high AFR and low ALOG. 
Figure~\ref{fig:constraint_gropued} (right) further attributes infeasible cases to specific patterns, which are highly concentrated in a few groups, most notably \(\text{Global}_7\) (31.7\%) and  \(\text{Global}_2\) (9.2\%), followed by mid-sized patterns (e.g., \(\text{Global}_{8/6/5}\)). Finally, Unsolvable accounts for 12.6\% of failures, indicating that a non-trivial portion of instances exceed the inference budget and remain unsolved under token limits. Per-model infeasibility mode is analyzed in Appendix~\ref{sec:app_res}.


\begin{table*}[t]
\centering
\setlength{\tabcolsep}{4pt}
\begin{tabular}{llrr}
\toprule
\textbf{Type} & \textbf{Model} & \textbf{Acc.} & \textbf{tok.} \\
\midrule
\multirow{7}{*}{Non-reasoning}
& qwen3-14b                              & 39.58 & 2276  \\
& ministral-14b-2512                    & 40.60 & 2511  \\
& nemotron3-Nano-30b                    & 35.26 & 3515  \\
& llama-4-maverick-17b-128e-instruct-maas & 53.07 & 1430  \\
& qwen3-235b-instruct                   & 69.26 & 2209  \\
& DeepSeek-V3.2                         & 66.98 & 1828  \\
& mimo-v2-flash                         & 41.72 & 7482  \\
\midrule
\multirow{9}{*}{Reasoning}
& qwen3-14b (reasoning)                 & 51.21 & 11816 \\
& nemotron3-Nano-30b (reasoning)        & 41.44 & 6655  \\
& qwq-32b (reasoning)                   & 54.70 & 11921 \\
& DeepSeek-V3.2 (reasoning)             & 76.19 & 4689  \\
& grok-4.1-fast (reasoning)             & 67.26 & 5462  \\
& claude-sonnet-4.5 (reasoning)         & 80.90 & 5798  \\
& o4-mini (reasoning)                   & 56.47 & 7555  \\
& gpt-5.1 (reasoning)                   & 76.88 & 4343  \\
& gemini-3-flash (reasoning)            & \textbf{83.77} & 20651 \\
\bottomrule
\end{tabular}
\caption{Performance comparison of different LLMs on solver-calling code generation.}
\label{tab:model_perf}
\end{table*}

\subsection{Reasoning Cost Analysis}

From Table~\ref{tab:nlco_main_results}, we can find that the high-performing LLMs (e.g., Gemini-3-Flash) achieve a high AFR and relatively low ALOG, but at the cost of extremely high token usage. To better answer \textbf{RQ3} for the \textbf{quality-efficiency trade-off} of reasoning, we introduce data profile~\citep{more2009benchmarking}, a standard benchmarking tool in optimization.

More precisely, for each instance $i$ and model $m$, let $t_{i,m}$ be the token cost of the model output.
We fix a success criterion (i.e., the solution is feasible or optimal).
We define the fraction of instances solved within a budget $B$ tokens as $
d_m(B)=\frac{1}{N}\sum_{i=1}^{N}\mathbb{I}\big[t_{i,m}\le B \ \wedge\ \text{success}(i,m)\big]$ ($N$ is the number of instances). In our settings, $d_m(B)$ for feasibility and optimality naturally translates to two metrics used in the paper, i.e., AFR and Acc. Plotting $d_m(B)$ against $B$ yields a curve where higher is better: a higher curve means the model solves more instances under the same token budget, and a curve that rises earlier indicates better efficiency.

As shown in Figure~\ref{fig:data_profile}, feasibility increases rapidly with budget and then saturates, suggesting that extra tokens mainly help with constraint satisfaction and self-checking up to a “knee” region, after which returns diminish. Optimality increases more gradually and plateaus well below feasibility, indicating that finding a valid solution is easier than reliably improving it to high quality. Across models, reasoning-enabled variants typically gain faster at small budgets (higher token efficiency) and reach higher plateaus. We also see that some models (e.g., DeepSeek-V3.2) attain high feasibility with modest budgets but convert additional tokens into little optimality improvement, implying that longer generations often boost validity more than objective value. Finally, although Gemini-3-Flash achieves the best overall results, it consistently uses more tokens than other top-tier models (e.g., DeepSeek), highlighting the need for more efficient reasoning.

We further analyze inference-time compute scaling via Best-of-$N$ sampling in Appendix~\ref{sec:app_res}.

\subsection{Solver-calling Capability Benchmarking}

Since solver code generation for optimization is important in real-world scenarios, we conduct experiments by prompting LLMs to write Gurobi code to solve the NLCO problems, and the results on Set-L are summarized in Table~\ref{tab:model_perf}. 

While the ability to write solver-calling code significantly improves results compared to the end-to-end setting, models continue to exhibit notable heterogeneity in both accuracy and reasoning efficiency. More specifically, models with exceptionally strong coding foundations like Claude-Sonnet-4.5 and Gemini-3-Flash perform significantly better. Meanwhile, even with the assistance of world-class solvers, model performance on NLCO is still not saturated (i.e., the best-performing LLM achieves 83.77\% accuracy). This confirms that NLCO is not only a test of internal logic but also a suitable benchmarking framework for the coding and mathematical modeling capabilities of LLMs. 

%% file: appendix.tex
\appendix

\section{Extended Related Work}
\label{app:literature}

\subsection{LLM for Optimization}

LLMs have attracted growing interest in optimization and operations research (OR) due to their reasoning and code-generation abilities~\cite{da2025large}. Unlike symbolic or algebraic reasoning, CO requires understanding not only numerical relationships but also the logical structure and interplay of constraints, as well as trade-offs induced by objective functions. This makes CO more challenging than typical math problem solving~\cite{101145math, wei2025coinmath}, where reasoning often reduces to deriving closed-form expressions or performing arithmetic transformations.

Existing work applying LLMs to optimization largely follows two paradigms: \emph{code-synthesis} and \emph{end-to-end} approaches. In the code-synthesis paradigm, LLMs generate or refine heuristic algorithms~\cite{ye2024reevo, eoh, guo2025nested, zhengmonte} or construct model formulations and solver calls for classical optimization engines~\cite{zhang2024solving, ahmaditeshnizi2024optimus, xiao2024chainofexperts, huang2025orlm, jiang2025droc}. While often effective, this approach fails to isolate the models' intrinsic reasoning ability because it is tightly coupled to pretrained knowledge and dependent on external solvers.

The second paradigm is end-to-end solving, where LLMs directly interpret natural-language descriptions, devise feasible solutions, or iteratively refine them without solver assistance~\cite{yang2024large, huang2025graphthought, abgaryan2025starjob, jiang2025large}. Such an approach has the potential to minimize user expertise and lower the barrier for applying CO in practice, while more faithfully reflecting the models' own reasoning ability. However, existing work typically focuses on a small number of problem families or bespoke datasets, and there is still no systematic assessment of end-to-end reasoning across heterogeneous CO problems presented purely in natural language.

\subsection{LLM Reasoning Benchmarking}

Reasoning is widely regarded as a cornerstone of higher intelligence and is crucial for understanding the practical potential of LLMs~\cite{ni2025survey}. LLM reasoning benchmarking has evolved significantly over the past several years, transitioning from simple grade-school math evaluation~\cite{cobbe2021training, uesato2022solving, shi2023language} to multifaceted assessment frameworks, such as Olympic competition problems~\cite{huang2024olympicarena}, commonsense reasoning~\cite{Bhagavatula2020Abductive, roewer2025accord}, logical reasoning~\cite{liu2020logiqa, zhu2025autologi}, and domain-specific reasoning~\cite{reddy2024docfinqa, zuomedxpertqa, yang2025probench, mostajabdaveh2025evaluating}. These benchmarks have played an important role in charting the rapid progress of LLMs across different reasoning regimes and domains.

With the rapid development of reasoning techniques, emergent LLMs now attain strong performance on many widely used benchmarks. For example, the Qwen-3 model~\cite{yang2025qwen3} achieves a score of 94.39 on GSM8K~\cite{cobbe2021training}, 98.0 on MATH-500~\cite{lightman2023let}, and 89.0 on AutoLogi~\cite{zhu2025autologi}, suggesting that current benchmarks are increasingly saturated at the high end. Meanwhile, most of them focus on numerical puzzles, formal logic, or multiple-choice QA, and therefore weakly capture the challenges posed by CO: reasoning under constraints, combinatorial decision spaces, and explicit objective functions, which are instantiated in our benchmark.

\subsection{Optimization Benchmarks}

Traditional optimization benchmark datasets, such as TSPLIB~\cite{reinelt1991tsplib}, have long been used as standards for evaluating algorithmic performance. However, these datasets are not well-suited for benchmarking LLMs, as they lack natural-language context and often operate at scales that exceed the tractable reasoning capacity of current models.

More relevant to NLCO are language-based benchmarks that focus on optimization or graph reasoning tasks. As summarized in Table~\ref{tab:benchmark-comparison}, early efforts largely evaluated the code-generation capabilities of LLMs, reflecting a period when model reasoning was comparatively weak and reliable problem solving required translating problems into executable programs~\cite{ramamonjison2022augmenting, xiao2024chainofexperts, huang2025llms, yang2025optibench, michailidis2025cp}. When end-to-end problem solving was considered at all, benchmarks typically centered on relatively simple, polynomial-time tasks~\cite{wang2023can, fan2024nphardeval, li2024visiongraph, tang2025grapharena}, as the limited capacity of early LLMs made NP-hard CO prohibitively difficult to evaluate meaningfully.

By contrast, NLCO is built around canonical CO problems that are all framed in natural language, enabling a faithful assessment of LLM reasoning on constraint-driven decision-making problems.

\onecolumn
\renewcommand\arraystretch{0.83}
\setlength{\tabcolsep}{1pt}

\begin{longtable}{@{}>{\small}r >{\small}l >{\small}c >{\small}c >{\small}c >{\raggedright\arraybackslash\small}c >{\small}c@{}}
\toprule
\# & CO Problem & Var. Sort & Family & Global pattern(s) & Canonical global constraint(s) & Objective \\
\midrule
\endfirsthead

\toprule
\# & CO Problem & Var. Sort & Family & Global pattern(s) & Canonical global constraint(s) & Objective \\
\midrule
\endhead

\midrule
\multicolumn{7}{r}{\emph{continued on next page}}\\
\bottomrule
\endfoot

\bottomrule
\addlinespace[5pt] 
\caption{Taxonomy of NLCO problems.
The column ``Global pattern(s)'' lists the set of global constraint patterns \(\text{Global}_k\) associated with the canonical CP model, where
\(\text{Global}_1\)–\(\text{Global}_8\) denote permutation / tour, budgeted subset, coverage / hitting, assignment, partitioning, connectivity, temporal consistency, and local graph labeling, respectively. The column ``Canonical global constraint(s)'' records the canonical CP global constraints.}
\label{tab:cop-taxonomy}\\
\endlastfoot

\multicolumn{7}{l}{\cellcolor{gray!10}\textit{\textbf{Constraint-family group: \textsc{Graph}}}}\\
\midrule
1 & Traveling Salesman Problem (TSP)                & \textsc{Graph} & \textsc{Graph} & $\text{Global}_1$          & \gconst{circuit} & linear \\
2 & Prize-Collecting TSP (PCTSP)             & \textsc{Graph} & \textsc{Graph} & $\text{Global}_{1,2}$      & \gconst{circuit}, \gconst{sum} & linear \\
3 & Orienteering Problem (OP)                & \textsc{Graph} & \textsc{Graph} & $\text{Global}_{1,2}$      & \gconst{circuit}, \gconst{sum} & linear \\
4 & Capacitated VRP (CVRP)                   & \textsc{Graph} & \textsc{Graph} & $\text{Global}_{1,2}$      & \gconst{circuit}, \gconst{knapsack} & linear \\
5 & TSP with Time Windows (TSPTW)            & \textsc{Graph} & \textsc{Graph} & $\text{Global}_{1,7}$    & \gconst{circuit}, \gconst{cumulative} & linear \\
6 & Pickup-and-Delivery Problem (PDP)        & \textsc{Graph} & \textsc{Graph} & $\text{Global}_{1,7}$
& \begin{tabular}[t]{@{}l@{}}
    \gconst{circuit}, \gconst{precedence},\\
    \gconst{cumulative}
  \end{tabular}
& linear \\
7 & Minimum Latency Problem (MLP)            & \textsc{Graph} & \textsc{Graph} & $\text{Global}_1$          & \gconst{circuit} & linear \\
8 & Quadratic Shortest Path Problem (QSPP)          & \textsc{Graph} & \textsc{Graph} & $\text{Global}_1$          & \gconst{path} & quadratic \\
9 & Steiner Tree Problem (STP)               & \textsc{Graph} & \textsc{Graph} & $\text{Global}_6$          & \gconst{tree} & linear \\
10 & Steiner Forest Problem (SFP)                   & \textsc{Graph} & \textsc{Graph} & $\text{Global}_6$          & \gconst{tree}, \gconst{nTrees} & linear \\
11 & $k$-Minimum Spanning Tree (KMST)       & \textsc{Graph} & \textsc{Graph} & $\text{Global}_{6,2}$      & \gconst{tree}, \gconst{cardinality} & linear \\ [0.5em]

\multicolumn{7}{l}{\cellcolor{gray!10}\textit{\textbf{Constraint-family group: \textsc{Pack}}}}\\
\midrule
12 & Bin Packing Problem (BPP)                & \textsc{Int}   & \textsc{Pack}  & $\text{Global}_{2,5}$      & \gconst{binPacking} & linear \\
13 & Cutting Stock Problem (CSP)              & \textsc{Int}   & \textsc{Pack}  & $\text{Global}_2$          & \gconst{knapsack}, \gconst{sum} & linear \\
14 & 2D Strip Packing (2SP)                        & \textsc{Int}   & \textsc{Pack}  & $\text{Global}_7$          & \gconst{noOverlap}, \gconst{maximum} & bottleneck \\
15 & Job-Shop Scheduling Problem (JSP)                & \textsc{Int}   & \textsc{Pack}  & $\text{Global}_7$          & \gconst{noOverlap}, \gconst{precedence} & bottleneck \\
16 & Flow-Shop Scheduling Problem (FSP)               & \textsc{Int}   & \textsc{Pack}  & $\text{Global}_7$          & \gconst{noOverlap}, \gconst{precedence} & bottleneck \\
17 & Open-Shop Scheduling Problem (OSP)               & \textsc{Int}   & \textsc{Pack}  & $\text{Global}_7$          & \gconst{noOverlap} & bottleneck \\
18 & RCPSP (makespan)                         & \textsc{Int}   & \textsc{Pack}  & $\text{Global}_{7,2}$      & \gconst{cumulative}, \gconst{precedence} & bottleneck \\
19 & Parallel Machines $P\parallel T_{\max}$ (PMS) & \textsc{Int}   & \textsc{Pack}  & $\text{Global}_7$          & \gconst{noOverlap} & bottleneck \\
20 & Single-Machine Total Weighted Tardiness & \textsc{Int}   & \textsc{Pack}  & $\text{Global}_7$          & \gconst{noOverlap} & linear \\
[0.5em]

\multicolumn{7}{l}{\cellcolor{gray!10}\textit{\textbf{Constraint-family group: \textsc{Count}}}}\\
\midrule
21 & Minimum Dominating Set (MDS)                  & \textsc{Set}   & \textsc{Count} & $\text{Global}_3$          & \gconst{coverage} & linear \\
22 & Set Cover Problem (SCP)                        & \textsc{Set}   & \textsc{Count} & $\text{Global}_{3,2}$      & \gconst{coverage}, \gconst{cardinality} & linear \\
23 & Set Packing  (SP)                          & \textsc{Set}   & \textsc{Count} & $\text{Global}_{2,5}$      & \gconst{disjoint}, \gconst{cardinality} & linear \\
24 & Set Partitioning Problem (SPP)                & \textsc{Set}   & \textsc{Count} & $\text{Global}_5$          & \gconst{partition} & linear \\
25 & Hitting Set Problem (HSP)                       & \textsc{Set}   & \textsc{Count} & $\text{Global}_3$          & \gconst{coverage} & linear \\
26 & Max $k$-Coverage (MkC)               & \textsc{Set}   & \textsc{Count} & $\text{Global}_{2}$       & \gconst{coverage}, \gconst{cardinality} & linear \\
27 & Generalized Assignment Problem (GAP)     & \textsc{Int}   & \textsc{Count} & $\text{Global}_{4,2}$      & \gconst{sum}, \gconst{channel} & linear \\
28 & Uncapacitated Facility Location (UFLP)   & \textsc{Int}   & \textsc{Count} & $\text{Global}_4$          & \gconst{sum}, \gconst{channel} & linear \\
29 & Capacitated Facility Location (CFLP)     & \textsc{Int}   & \textsc{Count} & $\text{Global}_{4,2}$      & \gconst{sum}, \gconst{channel} & linear \\
30 & $p$-Median (PMED)                             & \textsc{Int}   & \textsc{Count} & $\text{Global}_{4,2}$      & \gconst{cardinality}, \gconst{sum} & linear \\
31 & $p$-Center (PCENTER)                             & \textsc{Int}   & \textsc{Count} & $\text{Global}_{4,2}$      & \gconst{cardinality}, \gconst{maximum} & bottleneck \\
32 & Maximum Independent Set (MIS)            & \textsc{Int}   & \textsc{Count} & $\text{Global}_8$          & \gconst{sum}$^{\dagger}$ & linear \\
33 & Minimum Vertex Cover (MVC)               & \textsc{Int}   & \textsc{Count} & $\text{Global}_8$          & \gconst{sum}$^{\dagger}$ & linear \\
34 & Maximum Clique Problem (MCP)                       & \textsc{Int}   & \textsc{Count} & $\text{Global}_8$          & \gconst{sum}$^{\ddagger}$ & linear \\
35 & Knapsack Problem (KP)                    & \textsc{Int}   & \textsc{Count} & $\text{Global}_2$          & \gconst{knapsack}, \gconst{sum} & linear \\
36 & Maximum Diversity Problem (MDP)          & \textsc{Int}   & \textsc{Count} & $\text{Global}_2$          & \gconst{cardinality} & quadratic \\
37 & Quadratic Knapsack Problem (QKP)         & \textsc{Int}   & \textsc{Count} & $\text{Global}_2$          & \gconst{knapsack} & quadratic \\
38 & Maximum Cut Problem (MAXCUT)               & \textsc{Int}   & \textsc{Count} & $\text{Global}_8$          & \gconst{weightedSum}, \gconst{xor} & quadratic \\[0.5em]

\multicolumn{7}{l}{\cellcolor{gray!10}\textit{\textbf{Constraint-family group: \textsc{Comp}}}}\\
\midrule
39 & Three-Index Assignment (AP3)             & \textsc{Int}   & \textsc{Comp}  & $\text{Global}_4$          & \gconst{allDifferent}, \gconst{channel} & linear\\
40 & Quadratic Assignment Problem (QAP)       & \textsc{Int}   & \textsc{Comp}  & $\text{Global}_4$          & \gconst{allDifferent} & quadratic \\
41 & Graph Coloring Problem (GCP) (min colors)              & \textsc{Int}   & \textsc{Comp}  & $\text{Global}_{5,8}$      & \gconst{neq}$^{\dagger}$, \gconst{nValues} & linear \\
42 & Cutwidth Minimization Problem (CMP)      & \textsc{Int}   & \textsc{Comp}  & $\text{Global}_1$          & \gconst{allDifferent}, \gconst{maximum} & bottleneck \\
43 & Linear Ordering Problem (LOP)                   & \textsc{Int}   & \textsc{Comp}  & $\text{Global}_1$          & \gconst{allDifferent} & linear \\

\multicolumn{7}{@{}p{\linewidth}@{}}{\footnotesize
$^{\dagger}$Applied over edges of the input graph (e.g., one of the incident endpoints must satisfy the corresponding constraint).
\qquad
$^{\ddagger}$Applied over non-edges of the input graph (e.g., at most one endpoint of each non-edge may be selected).}\\

\end{longtable}

\twocolumn

\section{NLCO Benchmark Taxonomy}
\label{app:taxonomy}

Section~\ref{sec:method} introduces the NLCO taxonomy
\(\tau(P) = (\text{VarSort}(P), \text{Family}(P), \text{Global}(P), \text{ObjClass}(P))\)
that we use throughout the task selection and evaluation processes.
This appendix provides additional details on how we connect this taxonomy to established standard and how we assign labels to each dimension in a deterministic way.

Our starting point is the XCSP$^3$ specification~\cite{boussemart2016xcsp3}, which offers a compact language and catalogue of global constraints for representing constrained optimization and satisfaction problems, and thus applicable to CO problems.
For each CO task in NLCO, we consider a canonical Constraint Programming (CP) model
(either directly from the XCSP$^3$ catalogue or from standard CP/OR formulations) and
attach the four labels in \(\tau(P)\) to this underlying model rather than to any particular
natural-language description.

\paragraph{Variable sort.}
As discussed in Section~\ref{sec:method}, we group the primary decision
domains into \textsc{Int} (including \textsc{Bool}), \textsc{Set}, and \textsc{Graph}.
In practice, many CO problems admit multiple equivalent encodings. For example,
Set Cover can be written either with set variables or with $0/1$ indicator variables, while TSP can be written either with graph-structured variables or
with integer indices and auxiliary constraints. To make labels reproducible, we adopt a \emph{native-sort-first} convention:
whenever there exists a widely used formulation that naturally uses \textsc{Set}
or \textsc{Graph} variables, we assign that sort as primary, even if there also
exists a purely \textsc{Int}-based encoding. Concretely:
\begin{itemize}
    \item Knapsack and related packing problems are labeled as \textsc{Int}; the canonical models rely on integer or Boolean decision variables only. Likewise, classic vertex-selection problems defined on aa graph (e.g., MIS/MVC/MC) are labeled as \textsc{Int} rather than \textsc{Graph}: the primary decisions are $0/1$ labels $x_v\in\{0,1\}$ for vertices.
    In this taxonomy, \textsc{Graph} is reserved for problems whose decision objects are graph-structured (e.g., tours, paths) and are most naturally expressed with graph-scoped global constraints (e.g., \gconst{circuit} or \gconst{tree}).
    \item Set Cover and Set Partitioning are labeled as \textsc{Set}; their natural CP models use set variables with cardinality and inclusion constraints.
    \item Routing and network-design problems such as TSP and Steiner Tree are labeled as \textsc{Graph}, reflecting that their structure is most naturally expressed over graphs.
\end{itemize}
From the LLM's perspective, this convention explicitly distinguishes the primary mode of reasoning required for success: algebraic/logical (\textsc{Int}), set-based (\textsc{Set}), or relational/graph-based (\textsc{Graph}).

\paragraph{Constraint-family group.}
The \(\text{Family}(P)\) label identifies the dominant structural family
controlling feasibility, using the top-level groupings of XCSP$^3$:
\textsc{Comp}, \textsc{Count}, \textsc{Pack}, and \textsc{Graph}
(see Section~\ref{sec:method}).
Intuitively, each family corresponds to a prototypical reasoning pattern:
\begin{itemize}
    \item \textsc{Comp}: satisfying precedence or inequality relations, e.g.,
    assignment with order constraints.
    \item \textsc{Count}: respecting capacities, budgets, or coverage via sums
    of decision variables, e.g., knapsack or simple covering problems.
    \item \textsc{Pack}: arranging items in time or space without overlap, e.g.,
    job-shop scheduling or bin packing in space or time.
    \item \textsc{Graph}: satisfying constraints scoped to graph structure, e.g., tours/cycles, matchings, or flows/cuts.
\end{itemize}
In ambiguous cases, the family label is tied to which constraints would typically
drive propagation in a constraint satisfaction judgement (see ``primary constraint rule'' below).

\paragraph{Global constraint pattern(s).}
Within the chosen family, \(\text{Global}(P)\) abstracts away from classical global constraints and instead records the set of global patterns that a valid solution must satisfy.
It provides a more fine-grained dimension for analyzing feasibility than the constraint-family group.

Concretely, we start from a canonical CP model and its associated global constraints
(e.g., \gconst{circuit}, \gconst{binPacking}, \gconst{noOverlap}, \gconst{allDifferent}, \gconst{coverage}, \gconst{partition}, \gconst{networkFlow})
and map them into a small vocabulary of solution-structure templates, such as permutation, subset, and coverage.
Intuitively, each pattern describes what a correct answer `looks like' and what global relation over the decision variables must hold, as it would be stated in the text and checked by facility-checking scripts.

This design is motivated by two considerations.
First, it keeps the taxonomy anchored in standard CP practice, as the patterns are derived from well-studied global constraints~\cite{rossi2006handbook,regin2010global}, while providing a coarser, human-interpretable layer that better matches the structure of LLM-generated solutions.
Second, it reduces the granularity of the global-constraint space to a handful of recurring reasoning motifs, since there are more than 400 global constraints recorded in existing catalogues~\cite{beldiceanu2010global}.
Our approach makes it possible to aggregate and compare performance across superficially different CO tasks that share the same underlying patterns.

Formally, we describe the eight global constraint patterns below:
\begin{description}
    \item[\(\text{Global}_1\) (permutation / tour).]
    Solutions are permutations of a given set of items, which typically represent a path or a cycle (e.g., visiting all cities exactly once).
    Canonical CP constraints include \gconst{circuit}, \gconst{path}, \gconst{nPaths}, and \gconst{allDifferent} when used to enforce a permutation structure.

    \item[\(\text{Global}_2\) (budgeted subset).]
    Solutions select a subset of items subject to resource, capacity, or cardinality limits (e.g., weight or number of chosen items cannot exceed a bound).
    This pattern arises from \gconst{sum}, \gconst{weightedSum}, \gconst{knapsack}, \gconst{binPacking}, \gconst{cumulative}, and \gconst{cardinality}-type constraints when used as budget or capacity constraints.

    \item[\(\text{Global}_3\) (coverage / hitting).]
    Solutions ensure that a ground set is covered or dominated by the selected items or sets (e.g., every element is covered at least once, or every vertex is dominated by a chosen vertex).
    Canonical constraints include \gconst{coverage} and related hitting/dominating-set encodings.

    \item[\(\text{Global}_4\) (assignment).]
    Solutions assign items from one set to items in another (e.g., tasks to machines, facilities to customers), often under one-to-one or many-to-one constraints.
    This pattern is induced by \gconst{allDifferent}, \gconst{channel}, \gconst{element}, and similar linking constraints in assignment models.

    \item[\(\text{Global}_5\) (partitioning).]
    Solutions partition a ground set into disjoint groups or classes so that each element belongs to exactly one group (e.g., partitioning into sets or color classes).
    Canonical constraints include \gconst{partition}, \gconst{disjoint}, and \gconst{nValues} when used to encode partitions.

    \item[\(\text{Global}_6\) (connectivity).]
    Solutions form connected substructures such as paths, trees, or flows on a graph (e.g., connecting a set of terminals).
    Typical constraints are \gconst{path}, \gconst{tree}, \gconst{nTrees}, \gconst{networkFlow}, and \gconst{circuit} when used primarily to enforce connectivity.

    \item[\(\text{Global}_7\) (temporal consistency).]
    Solutions assign times or orders to activities under temporal, precedence, or non-overlap constraints (e.g., scheduling jobs on machines without conflict).
    Canonical constraints include \gconst{noOverlap}, \gconst{cumulative}, \gconst{precedence}, \gconst{sequence}, and \gconst{regular} in scheduling and rostering models.

    \item[\(\text{Global}_8\) (local graph labeling).]
    Solutions label vertices (or edges) of a graph so that local adjacency constraints are satisfied (e.g., adjacent vertices receive different colors, edges in a cut cross the partition).
    This pattern is induced by constraints such as \gconst{neq} on edges, \gconst{xor}, and \gconst{weightedSum} over incident variables in cut or independent-set formulations.
\end{description}

For each problem \(P\), \(\text{Global}(P)\) is a subset of \(\{\text{Global}_1,\dots,\text{Global}_8\}\) obtained by mapping the propagation-critical global constraints in a canonical CP model to these patterns.
We report the resulting patterns as \(\text{Global}_{k_1,k_2,\dots}\) in Table~\ref{tab:cop-taxonomy}. 

For example:
\begin{itemize}
    \item TSP and related routing problems are mapped to the \emph{permutation / tour} pattern (i.e., \(\text{Global}_1\)): a solution is a permutation of cities forming a Hamiltonian tour (captured in CP by \gconst{circuit}); prize-collecting and capacitated variants additionally carry \emph{budgeted subset} patterns (i.e., \(\text{Global}_2\)) via sum/knapsack-style constraints.
    \item Set Cover, Hitting Set, and Minimum Dominating Set are mapped to the \emph{coverage / hitting} pattern (i.e., \(\text{Global}_3\)): a solution is a subset (or family of subsets) that covers a ground set (captured by \gconst{coverage}-style constraints); some variants also include a \emph{budgeted subset} pattern when the size or cost of the chosen sets is bounded.
    \item Set Partitioning and closely related models fall under \emph{partitioning} (i.e., \(\text{Global}_5\)): the ground set is split into disjoint groups, typically represented by \gconst{partition}, \gconst{disjoint}, or \gconst{nValues}.
    \item Job-Shop Scheduling, Flow-Shop, RCPSP, and Parallel-Machine scheduling are mapped to \emph{temporal consistency} (i.e., \(\text{Global}_7\)): a solution assigns start times to tasks so that resources do not overlap and precedence relations are respected (captured by \gconst{noOverlap}, \gconst{cumulative}, and \gconst{precedence}); resource-constrained variants further add a \emph{subset under budget or capacity} pattern on renewable resources.
    \item Graph Coloring, Max-Cut, and independent-set-type problems are mapped to \emph{local graph labeling} (i.e., \(\text{Global}_8\)): each vertex receives a label or selection bit, and local adjacency constraints such as inequality or cut edges (\gconst{neq}, \gconst{xor}) must be satisfied.
\end{itemize}

We report \(\text{Global}(P)\) as an ordered list and refer to the first element as the primary pattern.
Anchoring evaluation to these (possibly multiple) patterns lets us interpret violations at the level of reasoning modes:
a candidate solution can be reported as ``violating the permutation/tour pattern'' (not a valid tour) or ``violating the coverage pattern'' (uncovered elements remain), rather than simply ``infeasible,''
while Table~\ref{tab:cop-taxonomy} still records the underlying global constraints for CP-oriented analysis.

\paragraph{Objective class.}
The \(\text{ObjClass}(P)\) label groups objectives into three classes:
linear, quadratic, and bottleneck (min--max / max--min). This is a coarse
but useful abstraction:
\begin{itemize}
    \item Linear objectives cover a large portion of classical optimization tasks
    (e.g., linear assignment, shortest paths, knapsack).
    \item Quadratic objectives arise in interaction-heavy problems such as
    Quadratic Assignment or Max-Cut, where costs depend on pairs of decisions.
    \item Bottleneck objectives appear in scheduling,
    where the goal is to minimize a worst-case quantity such as makespan.
\end{itemize}
Grouping tasks along this axis allows us to distinguish, for example, whether
a model that performs well on linear assignment-type problems generalizes to
quadratic interaction objectives.


\paragraph{Scope of the taxonomy.}
The taxonomy is not meant as an exhaustive classification of all CO problems.
Instead, it is a principled coarsening of the XCSP$^3$ landscape tailored to
the tasks in NLCO, chosen to balance granularity and interpretability
when aggregating results across related tasks.
Based on the framework above, Table~\ref{tab:cop-taxonomy} summarizes the
taxonomic labels for all tasks in NLCO and can be used as a reference when
interpreting the taxonomy-aware results.

\renewcommand\arraystretch{1.0}

\onecolumn
\small
\begin{longtable}{>{\raggedright\arraybackslash}p{5.1cm} p{3.9cm} @{\hspace{0.3cm}}p{2.4cm} >{\raggedright\arraybackslash}p{3.8cm}}

\toprule
\textbf{Problem} & \textbf{Data Source} & \textbf{Utilized Solver} & \textbf{Instance scale (S/M/L)} \\
\midrule
\endfirsthead

\multicolumn{4}{c}{\tablename\ \thetable\ -- \textit{Continued from previous page}} \\
\toprule
\textbf{Problem} & \textbf{Data Source} & \textbf{Utilized Solver} & \textbf{Instance scale (S/M/L)} \\
\midrule
\endhead

\midrule
\multicolumn{4}{r}{\textit{Continued on next page}} \\
\endfoot

\bottomrule
\addlinespace[5pt] 
\caption{ Summary of CO problem data sources, utilized solvers, and instance scales. The last column reports the small (S), medium (M), and large (L) instance size ranges used in NLCO, expressed in terms of the main size parameter for the corresponding problem (e.g., number of nodes, customers, items, vertices, or elements).}
\label{tab:detailed_summary}
\endlastfoot

\midrule
Travelling Salesman Problem (TSP) 
    & TSPLIB \cite{reinelt1991tsplib} 
    & LKH3 
    & 5--10 / 11--15 / 16--25 nodes \\

Prize-Collecting TSP (PCTSP) 
    & TSPLIB \cite{reinelt1991tsplib} 
    & Gurobi 
    & 5--10 / 11--15 / 16--25 nodes \\

Orienteering Problem (OP) 
    & Data in \cite{tsiligirides1984heuristic} 
    & GA \cite{KOBEAGA201842}, Gurobi 
    & 5--10 / 11--15 / 16--25 nodes \\

Capacitated Vehicle Routing (CVRP) 
    & CVRPLIB Set-X \cite{UCHOA2017845} 
    & HGS \cite{PyVRP_2024} 
    & 5--10 / 11--15 / 16--25 nodes (incl.\ depot) \\

TSP with Time Windows (TSPTW)  
    & TSPLIB \cite{reinelt1991tsplib} 
    & LKH3 
    & 5--10 / 11--15 / 16--25 nodes \\

Pickup and Delivery Problem (PDP)  
    & Li \& Lim benchmark \cite{li2001metaheuristic} 
    & Gurobi 
    & 3--5 / 6--8 / 9--14 requests \\

Minimum Latency Problem (MLP)  
    & TSPLIB \cite{reinelt1991tsplib} 
    & Gurobi 
    &  5--10 / 11--15 / 16--25 nodes \\

Quadratic Shortest Path Problem (QSPP)  
    & Synthetic data following \cite{rostami2018quadratic} 
    & Gurobi 
    &  5--10 / 11--15 / 16--25 nodes \\

Steiner Tree Problem (STP) 
    & Data in \cite{Leitner2014_STP_RealWorld} 
    & Gurobi 
    &  10--15 / 16--20 / 21--30 vertices\\

Steiner Forest Problem (SFP) 
    & Data in \cite{Leitner2014_STP_RealWorld} 
    & Gurobi 
    &  10--15 / 16--20 / 21--30 vertices\\

$k$-Minimum Spanning Tree (KMST) 
    & Data in \cite{Leitner2014_STP_RealWorld} 
    & Gurobi 
    &  10--15 / 16--20 / 21--30 vertices\\ [0.5em]


\midrule
Bin Packing Problem (BPP) 
    & Data in \cite{falkenauer1996hybrid} 
    & Gurobi 
    & 5--10 / 11--15 / 16--25 items \\

Cutting Stock Problem (CSP) 
    & BPPLIB \cite{delorme2018bpplib} 
    & Gurobi 
    & 5--10 / 11--15 / 16--25 items \\

2D Strip Packing (2SP)  
    & 2DPackLib \cite{iori20222dpacklib} 
    & Gurobi 
    & 5--10 / 11--15 / 16--25 items  \\

Job-Shop Scheduling Problem (JSP) 
    & Taillard Generator~\cite{taillard1993benchmarks}  
    & CP-SAT 
    & \makecell[l]{2--5 / 6--8 / 9--12 jobs\\2--3 / 3--5 / 5--6 machines }\\

Flow-Shop Scheduling Problem (FSP) 
    & Taillard Generator~\cite{taillard1993benchmarks}  
    & CP-SAT 
    & \makecell[l]{2--5 / 6--8 / 9--12 jobs\\2--3 / 3--5 / 5--6 machines }\\

Open-Shop Scheduling Problem (OSP) 
    & Taillard Generator~\cite{taillard1993benchmarks}  
    & CP-SAT 
    &  \makecell[l]{2--5 / 6--8 / 9--12 jobs\\2--3 / 3--5 / 5--6 machines }\\

Resource-Constrained Project Scheduling Problem (RCPSP) 
    & Constraint benchmarking tools suite~\cite{PelleauMTB13} 
    & CP-SAT 
    & 4--10 / 11--15 / 16--21 tasks\\

Parallel Machines Scheduling (PMS) 
    & Tanaka and Araki benchmark~\cite{tanaka2008branch} 
    & CP-SAT 
    & \makecell[l]{5--10 / 11--15 / 16--25 jobs\\2--2 / 2--3 / 2--5 machines }\\

Single-Machine Total Weighted Tardiness (SMTWT) 
    & OR-library~\cite{beasley1990or} 
    & CP-SAT 
    & 5--10 / 11--20 / 21--30 jobs\\[0.5em]


\midrule
Minimum Dominating Set (MDS)      
    & DIMACS10 Street Network~\cite{Bader2014} 
    & Gurobi 
    & 8--12 / 13--18 / 19--25 vertices \\

Set Covering Problem (SCP)      
    & DIMACS10 Road Network~\cite{Bader2014} 
    & Gurobi 
    & 8--12 / 13--18 / 19--25 elements \\

Set Packing Problem (SP)   
    & DIMACS10 Road Networks~\cite{Bader2014} 
    & Gurobi 
    & 8--12 / 13--18 / 19--25 elements \\

Set Partitioning Problem (SPP) 
    & Synthetic Benchmark Instances (following~\cite{kadioglu2012non}) 
    & Gurobi 
    & 8--12 / 13--18 / 19--25 elements \\

Hitting Set Problem (HSP)   
    & DIMACS10 Road Networks~\cite{Bader2014} 
    & Gurobi 
    & 8--12 / 13--18 / 19--25 elements \\

Max \texorpdfstring{$k$}{k}-Coverage (MkC)   
    & DIMACS10 Road Networks~\cite{Bader2014} 
    & Gurobi 
    & 8--12 / 13--18 / 19--25 elements \\

Generalized  Assignment Problem (GAP) 
    & OR-Library \cite{chu1997genetic} 
    & Gurobi 
    & 5--10 / 11--15 / 16--25 items \\

Uncapacitated Facility Location (UFLP) 
    & UflLIB \cite{hoefer2006ufllib} 
    & Gurobi 
    & 5--8 / 9--14 / 15--20 customers \\

Capacitated Facility Location (CFLP) 
    & Data in \cite{holmberg1999exact} 
    & Gurobi 
    & 5--8 / 9--14 / 15--20 customers \\

$p$-Median Facility Location (PMED)
    & OR-Library \cite{beasley1985note} 
    & Gurobi 
    &  5--8 / 9--14 / 15--20 customers \\

$p$-Center Facility Location (PCENTER)
    & OR-Library \cite{beasley1985note} 
    & Gurobi 
    &  5--8 / 9--14 / 15--20 customers \\

Maximum Independent Set (MIS)      
    & DIMACS10 Redistricting Graph \cite{Bader2014} 
    & Gurobi 
    & 8--12 / 13--18 / 19--25 vertices \\

Minimum Vertex Cover (MVC)      
    & DIMACS10 Redistricting Graph \cite{Bader2014} 
    & Gurobi 
    & 8--12 / 13--18 / 19--25 vertices \\

Maximum Clique Problem (MCP)       
    & DIMACS10 Citation / Co-author Networks~\cite{Bader2014} 
    & Gurobi 
    & 8--12 / 13--18 / 19--25 vertices \\

Knapsack Problem (KP) 
    & Synthetic (uncorrelated) data following \cite{pisinger2005hard} 
    & Gurobi 
    & 5--10 / 11--20 / 21--30 items \\

Maximum Diversity Problem (MDP) 
    & MDPLIB~\cite{marti2021mdplib} 
    & Gurobi 
    &  8--12 / 13--18 / 19--25 items\\

Quadratic Knapsack Problem (QKP) 
    & QKPLIB~\cite{jovanovic2023qkplib} 
    & Gurobi 
    & 5--10 / 11--15 / 16--25 items\\

Maximum Cut Problem (MAXCUT)   
    & DIMACS10 Redistricting Graph \cite{Bader2014} 
    & Gurobi 
    & 8--12 / 13--18 / 19--25 vertices \\ [0.5em]

\midrule
Three-Index  Assignment Problem (AP3) 
    & Data in \cite{balas1991algorithm} 
    & Gurobi 
    & 3--5 / 6--8 / 9--12 items \\

Quadratic Assignment Problem (QAP) 
    & QAPLIB \cite{burkard1997qaplib} 
    & Gurobi 
    & 3--5 / 6--8 / 9--12 items \\

Graph Coloring Problem (GCP)       
    & DIMACS10 Citation / Co-author Networks~\cite{Bader2014} 
    & Gurobi 
    & 8--12 / 13--18 / 19--25 vertices \\

Cutwidth Minimization Problem (CMP)
    & CMPLIB \cite{marti2013branch}
    & Gurobi 
    & 8--12 / 13--18 / 19--25 vertices \\

Linear Ordering Problem (LOP)
    & LOLIB \cite{marti2012benchmark} 
    & Gurobi 
    &  5--10 / 11--15 / 16--25 items \\

\end{longtable}

\twocolumn
\normalsize

\section{Problem Details}
\label{sec:problem details}
This section introduces all the NLCO tasks, along with their data generation processes. Table~\ref{tab:detailed_summary} provides an overview on the data source, utilized solver, and instance scale of each task, while a more detailed statistic for set problems (i.e., SCP, MkC, SP, HSP, and SPP) and graph problems (i.e., MIS, MVC, MCP, GCP, MDS, and MAXCUT) are summarized in Table~\ref{tab:set_summary} and  Table~\ref{tab:graph_summary}, respectively.

For all problems, the utilized solver (e.g., Gurobi) automatically determines whether a feasible solution exists for the sampled instance, thereby ensuring the feasibility of the generated data.

\subsection{Constraint-family Group: GRAPH}
\subsubsection{Traveling Salesman Problem (TSP)}
The Traveling Salesman Problem (TSP) seeks a minimum-length tour that visits each of the \(n\) nodes exactly once and returns to the starting point.

\paragraph{Instance Creation.}
For each TSP instance, node coordinates are extracted by sampling \(n\) points from a randomly selected TSP instance from TSPLIB~\cite{reinelt1991tsplib}. The sampled coordinates are then normalized to an integer grid within $[0,100]^2$. Each resulting instance is solved by LKH-3 for the reference solution.

\subsubsection{Prize-Collecting TSP (PCTSP)}

\paragraph{Instance Creation.}
PCTSP instances are generated using the same TSPLIB-based coordinate sampling procedure as standard TSP instances. The coordinates are then normalized to the integer grid \([0,100]^2\), with node 1 designated as the depot.

Node prizes are sampled i.i.d.\ from the discrete uniform distribution \(\{0,\dots,100\}\), with the depot assigned a prize of zero. 
The minimum required collected prize is set to \(25n\), corresponding to half of the expected total prize under our discrete \(\{0,\dots,100\}\) prize distribution. 

Penalties are scaled using the size–dependent tour-length estimates \(L_n\) (we sample 1,000 instances for each setting and calculate tour length for the corrsponding TSP), i.e.,
\(
L_n \in \{334,\,388,\,451\}
\)
for instance ranges \(n\in[5,10]\), \(n\in[11,20]\), and \(n>21\), respectively. Each penalty is sampled independently as $\beta_i \sim \mathcal{U}\!\left(0,\; \frac{3L_n}{2n}\right),$
following the settings in \citet{kool2018attention}.

\subsubsection{Orienteering Problem (OP)}
The Orienteering Problem (OP) is a classical NP-hard routing problem, which chooses a subset of \(n\) nodes to visit within a limited travel-length budget $T_n$ in order to maximize the total collected prize.

\paragraph{Instance Creation.}
For each OP instance, node coordinates are sampled from the instances used by \citet{tsiligirides1984heuristic} and normalized to an integer grid within $[0,100]^2$. The instance is modeled as a closed tour that starts and ends at the origin (the first point), which is with zero prize. A travel-length budget $T_n$ is then sampled uniformly at random from $\{5,...,30\}$. When generating an instance of size $n$, the depot is always retained and the remaining $n-1$ nodes are uniformly sub-sampled without replacement from the non-depot nodes.

For instances with $n \leq 20$, an exact Gurobi Mixed Integer Linear Programming (MILP) formulation is used; otherwise, a Genetic Algorithm (GA)-based solver \cite{KOBEAGA201842} is applied. The recorded objective value is the collected prize of the returned tour.




\subsubsection{Capacitated Vehicle Routing Problem (CVRP)}
The Capacitated Vehicle Routing Problem (CVRP) seeks a set of routes that serve all customers exactly once using a fleet of identical vehicles with limited capacity. Each route starts and ends at a single depot, and the sum of customer demands on any route must not exceed the vehicle capacity $Q$. Under Euclidean distances, the objective is to minimize the total travel distance across all routes.

\paragraph{Instance Creation.}
For each CVRP instance, a base problem is sampled uniformly at random from a CVRPLIB instance \cite{UCHOA2017845}, where each entry provides depot and customer coordinates, node demands, and a vehicle capacity. Given a target size \(n\) (counting the depot), the extractor always retains the depot (node \(0\)) and sub-samples \(n-1\) customers uniformly without replacement from the remaining nodes. The resulting coordinates are then min--max normalized to \([0,100]^2\) (independently per coordinate dimension), while demands and capacity are kept from the original instance. The depot index is fixed to \(0\).

Each instance is solved using a Hybrid Genetic Search (HGS) solver under the original vehicle capacity \(Q\). The solver returns a set of vehicle routes and the corresponding total travel distance. The recorded objective value is this total distance.

\subsubsection{Traveling Salesman Problem with Time Windows (TSPTW)}
The Traveling Salesman Problem with Time Windows (TSPTW) extends the classical TSP by adding temporal constraints. Each customer node $i$ is associated with a time window $[l_i,\ u_i] \in \mathbb{Z}^+$ and $l_i < u_i$) and the vehicle must arrive within this interval; early arrivals must wait until $l_i$, while late arrivals are infeasible. The objective is to construct a TSP-like minimum-length tour while respecting all time-window constraints.

\paragraph{Instance Creation.}


For instance creation, each instance is created by sampling $n$ nodes from instances in TSPLIB and normalizing all coordinates to an integer grid in \([0,100]^2\). To set a meaningful time scale for sampling time windows, we follow the generation scheme in recent works \cite{bi2024learning,chen2024looking}, which requires a size–dependent estimate of the expected TSP tour length \(L_n\). Because the spatial distribution is not uniform, we calibrate these values offline by generating 1000 additional TSP instances per size range, solving them using LKH, and computing the average tour length $L_n$\footnotemark
\footnotetext{
Using 1000 sampled instances per size range, the empirical tour lengths are
\(L_{5\text{--}10}=334.20\),
\(L_{11\text{--}20}=388.07\),
\(L_{21\text{--}30}=451.38\), so we set
\(
L_n \in \{334, 388, 451\}
\)}
for instance ranges $n \in [5,10]$, $n \in [11,20]$, and $n>20$, respectively.

For each customer node $i$, the lower bound of its time window is sampled uniformly from $l_i \sim \mathrm{Unif}[0,\, L_n]$,
and the window width is drawn from $W_i = L_n \cdot \mathrm{Unif}[\alpha,\beta]$,
with $\alpha = 0.5$ and $\beta = 0.75$, giving the upper time window bound $u_i = l_i + W_i$.  
The depot is assigned a wide and always–feasible window $[0,\,(2+\beta)L_n]$ to allow free departure and return. Each instance is then solved with LKH-3 to obtain the (near-)optimal total travel distance.

\subsubsection{Pickup and Delivery Problem (PDP)}
The Pickup and Delivery Problem (PDP) requires constructing a minimum-distance route that starts and ends at a depot while serving a set of pickup–delivery request pairs. Each pickup must be visited before its corresponding delivery. In this implementation, travel time between any pair of nodes is propotinal to their Euclidean distance, and time windows are included for each node.

\paragraph{Instance Creation.}

We sample instances from the Li \& Lim benchmark, which provides node coordinates, demands, time windows, and pickup–delivery pairs. To create a sub-instance with a target number of requests $r$, we first sample $r$ pickup–delivery pairs. The depot (node 0) is always included, forming a complete instance of $n=1+2r$ nodes. Finally, we normalize the node coordinates and consistently re-index the retained time windows.

Each generated instance is solved using a Gurobi MILP implementation that minimizes total travel distance. The model enforces a single route from a start depot to an end depot (implemented by duplicating the depot as the end node), eliminates sub-tours via MTZ-style ordering variables, enforces time-window feasibility via continuous arrival-time variables with big-M propagation, and enforces precedence constraints for every pickup–delivery pair. The targeted objective is to minimize the total travel distance.

\subsubsection{Minimum Latency Problem (MLP)}
The Minimum Latency Problem (MLP) seeks a tour that visits each of the $n$ nodes exactly once while minimizing the sum of arrival times (total latency) rather than the total tour length. Under Euclidean distances, the latency of a node is the cumulative travel distance from the depot along the tour until the node is first reached, and the MLP objective is the sum of these latencies over all visited nodes.

\paragraph{Instance Creation.}
For each MLP instance, node coordinates are obtained from the same data sources as TSP, and the data generation also follows the process of TSP. 

The exact optimal solution to each resulting instance is solved using a Gurobi-based formulation for MLP (as LKH is tailored to tour-length objectives). The reference tour is computed by considering node with index $0$ as the depot, and the recorded objective value is the total latency, i.e., the sum of arrival times along the returned tour.

\subsubsection{Quadratic Shortest Path Problem (QSPP)}
The Quadratic Shortest Path Problem (QSPP) is defined on a directed graph \(G=(V,A)\) with a designated source node \(s\) and target node \(t\). Each arc \(a\in A\) is associated with a binary decision variable \(x_a\in\{0,1\}\) indicating whether the arc is selected. The objective combines linear and quadratic interaction costs $\min \; x^\top Q x + g^\top x + c$,
where \(g\in\mathbb{R}^{|A|}\) are linear arc costs, \(Q\in\mathbb{R}^{|A|\times |A|}\) is a (possibly dense) quadratic cost matrix, and \(c\) is a constant. Feasibility is enforced by one-unit \(s\rightarrow t\) flow constraints:
\[
\sum_{a\in \delta^+(v)} x_a - \sum_{a\in \delta^-(v)} x_a =
\begin{cases}
1 & v=s,\\
-1 & v=t,\\
0 & \text{otherwise},
\end{cases}
\]
which selects a directed \(s\)--\(t\) path (potentially with extra cycles in general, though the path is extracted from the selected subgraph).

\paragraph{Instance Creation.}
To reliably control the instance size, we generate instances directly using the grid families proposed by~\citet{rostami2018quadratic}. Given a target number of nodes \(n=|V|\), we construct one of the following families:

\begin{itemize}
  \item \textbf{Grid1:} a \(k\times k\) directed grid with \(n=k^2\) nodes. Arcs connect each node to its \emph{right} and \emph{up} neighbors when they exist. The source is the lower-left corner and the target is the upper-right corner.
  \item \textbf{Grid2:} an \(n_r\times n_c\) transshipment grid plus explicit source and target, so \(n=n_r n_c + 2\). The source connects to all nodes in the first column, all nodes in the last column connect to the target, and internal arcs connect each transshipment node to its \emph{right} and \emph{down} neighbors when they exist. To match a requested \(n\), we set \(N=n-2\) and choose \((n_r,n_c)\) either as a factor pair closest to \(\sqrt{N}\) (square) or using a \(16\times (N/16)\) heuristic when divisible (long/wide); otherwise we fall back to \(1\times N\) or \(N\times 1\).
\end{itemize}

For each generated graph, arc indices are assigned in the construction order (\texttt{var\_index = 0,\dots,|A|-1}). Linear costs are sampled i.i.d. as integers \(g_a \sim \mathrm{Unif}\{1,\dots,10\}\). Quadratic costs are sampled as a dense symmetric matrix: for \(i\le j\), \(Q_{ij}\sim \mathrm{Unif}\{1,\dots,10\}\), and we set \(Q_{ji}=Q_{ij}\). The constant term is set to \(c=0\).

Each instance is solved with Gurobi as a (generally nonconvex) binary quadratic program. We enforce the unit-flow constraints above and minimize \(x^\top Q x + g^\top x + c\). The recorded solution is the extracted \(s\)--\(t\) path (found by running BFS in the subgraph induced by arcs with \(x_a=1\)), and the recorded objective value is the solver’s optimal objective \(x^\top Q x + g^\top x + c\).

\subsubsection{Steiner Tree Problem (STP)}
The Steiner Tree Problem (STP) seeks a minimum-cost tree in an undirected weighted graph that connects a given set of terminal nodes. To reduce the overall cost, the solution may include additional non-terminal vertices, referred to as Steiner nodes.

\paragraph{Instance Creation.}
We construct STP instances based on real-world benchmark graphs from ~\citet{Leitner2014_STP_RealWorld}. For each instance, we randomly select a source graph and parse its undirected weighted edges. 

Given a target size \(n\), we sample a structurally coherent node subset \(V'\) using a random walk with restart (RWR), following the procedure in GraphArena~\cite{tang2025grapharena}. Starting from a randomly chosen seed node (sampled from available terminals when provided, otherwise from all vertices), the walk is executed for \(L=5000\) steps with restart probability \(\alpha = 0.15\). Vertices are ranked by visitation frequency, and the top-\(n\) vertices are retained to form an induced subgraph \(G' = G[V']\). Disconnected samples are rejected.
We then generate the terminal set \(T' \subseteq V'\) by sampling a terminal ratio \(r \in \{0.1, 0.2, 0.3, 0.4, 0.5\}\) and selecting \(|T'| = \max(2, \lfloor r|V'| \rfloor)\) terminals uniformly at random from \(V'\). Finally, vertices in \(G'\) are relabeled to contiguous indices \(\{1, \dots, |V'|\}\), while original edge costs are preserved.

We categorize instances into three difficulty tiers based on subgraph size:
\[
\begin{array}{ll}
\text{Set-S:}  & |V'| \in [10, 15], \\
\text{Set-M:} & |V'| \in [16, 20], \\
\text{Set-L:}  & |V'| \in [21, 30].
\end{array}
\]

Table~\ref{tab:stp_stats} summarizes the statistics of the generated STP instances across the three size levels. 
Ground truth solutions are computed using Gurobi.

\begin{table}[thb]
\centering

\renewcommand{\arraystretch}{1.20}
\resizebox{\linewidth}{!}{
\begin{tabular}{lccccccc}
\toprule
\textbf{Tier} 
& $\overline{D}$ & $D_{\min}$ & $D_{\max}$ 
& $\overline{|T|}$ & $|T|_{\min}$ & $|T|_{\max}$ 
& $\overline{|T|/|V|}$ \\
\midrule
Set-S  & 0.2064 & 0.1429 & 0.2727 & 3.68 & 2 & 7 & 0.3051 \\
Set-M & 0.1447 & 0.1158 & 0.1833 & 5.78 & 2 & 10 & 0.3233 \\
Set-L  & 0.1037 & 0.0805 & 0.1381 & 6.92 & 2 & 14 & 0.2768 \\
\bottomrule
\end{tabular}
}
\caption{Statistics of generated Steiner Tree (STP) instances across three levels. $D$ denotes graph density, $|T|$ denotes the number of terminals, and $|V|$ denotes the number of vertices in the sampled subgraph.}
\label{tab:stp_stats}
\end{table}

\subsubsection{Steiner Forest Problem (SFP)}

The Steiner Forest Problem (SFP) generalizes the Steiner Tree Problem by requiring connectivity only within multiple disjoint terminal groups. The objective is to find a minimum-cost forest such that terminals belonging to the same group are connected, while terminals from different groups need not be connected.

\paragraph{Instance Creation.}
We generate SFP instances using the same subgraph extraction pipeline as in STP.
Given the sampled subgraph, we generate multiple terminal groups.
We first sample a terminal ratio $r \in \{0.1, 0.2, 0.3, 0.4, 0.5\}$ and select $|T|$ terminals accordingly.
The selected terminals are then randomly partitioned into $g$ groups, where $g \in [2,4]$ is sampled subject to feasibility constraints, and each group contains at least two terminals.
All vertices are relabeled to contiguous indices, and original edge costs are preserved.

We categorize SFP instances into the same three difficulty tiers based on subgraph size as in STP.

We solve Steiner forest instances using Gurobi.
Table~\ref{tab:sfp_stats} reports summary statistics of the generated SFP instances across the three size levels.

\begin{table}[thb]
\centering
\renewcommand{\arraystretch}{1.20}
\begin{tabular}{lccccccc}
\toprule
\textbf{Tier} 
& $\overline{D}$ & $D_{\min}$ & $D_{\max}$
& $\overline{|T|}$ 
& $\overline{|T|/|V|}$
& $\overline{\#\text{Groups}}$ \\
\midrule
Set-S  & 0.2063 & 0.1333 & 0.2667 & 4.48 & 0.3743 & 2.14 \\
Set-M & 0.1447 & 0.1158 & 0.1833 & 5.64 & 0.3175 & 2.24 \\
Set-L  & 0.1036 & 0.0813 & 0.1333 & 7.28 & 0.2900 & 2.42 \\
\bottomrule
\end{tabular}
\caption{Statistics of generated Steiner Forest (SFP) instances across three levels. $D$ denotes graph density, $|T|$ the total number of terminals, $|V|$ the number of vertices in the sampled subgraph, and $\#\text{Groups}$ the number of terminal groups}
\label{tab:sfp_stats}
\end{table}

\subsubsection{$k$-Minimum Spanning Tree (KMST)}

The $k$-Minimum Spanning Tree (KMST) problem seeks a minimum-cost tree that spans exactly $k$ vertices of an undirected weighted graph.

\paragraph{Instance Creation.}
We generate KMST instances using the same subgraph extraction pipeline as in STP. 
Given the sampled subgraph, we determine the target size $k$ by sampling a ratio 
$r \in \{0.3, 0.4, 0.5\}$ relative to the subgraph size and setting $k=\max(2,\lfloor r|V'|\rfloor)$. 
All vertices are relabeled to contiguous indices, and original edge costs are preserved. We categorize $k$-MST instances into the same three difficulty tiers based on subgraph size as in STP. We solve all KMST instances to optimality using Gurobi.
Table~\ref{tab:kmst_density} reports summary statistics of the generated KMST instances.

\begin{table}[thb]
\centering
\renewcommand{\arraystretch}{1.20}
\begin{tabular}{lcccccc}
\toprule
\textbf{Tier} 
& $\overline{k}$ 
& $\overline{k/|V|}$ 
& $\overline{D}$ 
& $D_{\min}$ 
& $D_{\max}$ \\
\midrule
Set-S  &  4.74  & 0.3906 & 0.2127 & 0.1429 & 0.2667 \\
Set-M &  7.32  & 0.4120 & 0.1451 & 0.1170 & 0.1750 \\
Set-L  &  10.76 & 0.4287 & 0.1045 & 0.0805 & 0.1333 \\
\bottomrule
\end{tabular}
\caption{Statistics of generated KMST instances across three levels. $k$ denotes the number of selected vertices, $|V|$ the number of vertices in the sampled subgraph, and $D$ the graph density.}
\label{tab:kmst_density}
\end{table}

\subsection{Constraint-family Group: PACK}
\subsubsection{Bin Packing Problem (BPP)}

The Bin Packing Problem (BPP) asks to pack \(n\) items of given sizes into identical bins of fixed capacity while using as few bins as possible. Each item must be assigned to exactly one bin, and the total size of items cannot exceed the bin capacity.

\paragraph{Instance Creation.}
For all BPP instances, item sizes and bin capacities are drawn from the OR-Library datasets~\citep{falkenauer1996hybrid}. For each instance, a dataset file is selected at random and \(n\) items are uniformly sampled, while the original bin capacity is retained. Each sampled instance is solved using Gurobi.

\subsubsection{Cutting Stock Problem (CSP)}
The Cutting Stock Problem (CSP) seeks to determine how to cut large stock materials (e.g., rolls or bars) into smaller requested piece sizes while meeting all demands and minimizing waste (or, equivalently, minimizing the number of stock pieces used). Each stock piece has a fixed length/capacity, each item type has a required size and demand, and a solution specifies cutting patterns that exactly cover the demands without exceeding the stock capacity.

\paragraph{Instance Creation.}
CSP instances are generated by sub-sampling from BPPLIB \cite{delorme2018bpplib}. For each generated instance, we specify the number of distinct item types, $m$. We first select a source problem from the dataset and inherit its bin capacity $C$. We then randomly sample $m$ item types from the source data, extracting their associated weights $w_i$ and demands $d_i$. Ground truth optimal solutions are computed using a pattern-based integer linear programming formulation solved by Gurobi, where the decision variables represent the usage frequency of feasible cutting patterns.

\subsubsection{2D Strip Packing (2SP)}
The goal of 2D Strip Packing (2SP) is to place a set of axis-aligned rectangles, without overlap, into a strip of fixed width and unbounded height so as to minimize the used height. Each rectangle must be packed entirely within the strip, and the objective is to produce a compact layout with the smallest possible makespan height.

\paragraph{Instance Creation.}
2SP instances are generated by sub-sampling item types from the 2DPackLib benchmark~\cite{iori20222dpacklib}. 
Each item type $i$ is characterized by width $w_i$, height $h_i$, and demand $d_i$, yielding a total of $\sum_i d_i$ rectangles. 

To avoid trivial instances where the optimal solution height equals the tallest item, we apply several techniques.
The strip width is set to $W = 0.8 \cdot W_{\mathrm{o}}$ to increase packing density ($W_{\mathrm{o}}$ is the original bin width of the benchmark data).
We scale the height limit to preserve the parent instance tightness while ensuring non-triviality: letting $A=\sum_i w_i h_i d_i$ and $LB=\left\lceil A/W\right\rceil$, we compute the parent tightness ratio $\rho = H_{\mathrm{parent}}/LB_{\mathrm{parent}}$ and set the sub-instance height to $H_{\mathrm{sub}}=\max\!\left(\left\lceil 1.3 \cdot \max_i h_i\right\rceil,\ \left\lceil \rho \cdot LB_{\mathrm{sub}}\right\rceil\right)$.
During sampling, we reject item subsets that lack height diversity (requiring at least two items with height $\geq 0.7 \cdot \max_i h_i$), have insufficient material (total stacked height $\sum_i h_i d_i < 1.5 \cdot H_{\mathrm{sub}}$), can fit in a single row (total width demand $\sum_i w_i d_i < 2W$), or have area utilization outside $[0.60, 0.95]$.
After solving with Gurobi, instances with optimal height $H^* < 1.1 \cdot \max_i h_i$ are resampled (up to 30 retries) to ensure non-trivial lower bounds.
Optimal solutions are obtained via Gurobi with a position-based MIP formulation, where non-overlap constraints are enforced via binary indicator variables.

\subsubsection{Job-Shop Scheduling Problem (JSP)}
Job-shop Scheduling Problem (JSP) aims at minimizing the makespan by finding an optimal schedule for a set of jobs, each consisting of a sequence of operations that must be processed on specific machines in a fixed order. 

\paragraph{Instance Creation.}
Instances are generated following the well-established standard of ~\citet{taillard1993benchmarks}. Processing times are independently drawn from a uniform integer distribution in $[1,99]$ using Taillard’s linear congruential random number generator to ensure reproducibility. Each job consists of exactly $m$ operations, where $m$ is the number of machines, and visits every machine exactly once. For each job, the machine order is given by a uniformly random permutation. The reference solutions are obtained using OR-Tools CP-SAT solver.

\subsubsection{Flow-Shop Scheduling Problem (FSP)}
The Flow-shop Scheduling Problem (FSP) considers a set of jobs that must all be processed on the same set of machines in an identical order. Each job consists of exactly one operation per machine. The objective is typically to minimize the makespan. 

\paragraph{Instance Creation.}
Instances are generated following~\citet{taillard1993benchmarks}. Processing times of generated instances are drawn uniformly from $[1,99]$ following Taillard’s benchmark methodology, and all jobs follow the fixed machine sequence $(1,2,\dots,m)$. The reference solutions are also obtained using OR-Tools CP-SAT solver.

\subsubsection{Open-Shop Scheduling Problem (OSP)}
The Open-shop Scheduling Problem (OSP) generalizes both JSP and FSP settings by removing predefined operation orders within jobs. Each job requires processing on every machine exactly once, but the order in which its operations are executed is not fixed and must be decided by the scheduler. 

\paragraph{Instance Creation.}
As in the flow-shop case, processing times are generated uniformly in $[1,99]$ following~\citet{taillard1993benchmarks}. The reference solutions are calculated using OR-Tools CP-SAT solver.

\subsubsection{Resource-Constrained Project Scheduling Problem (RCPSP)}

The Resource-Constrained Project Scheduling Problem (RCPSP) consists of a set of tasks subject to precedence constraints and limited renewable resources. Each task is characterized by a fixed processing duration and a vector of resource demands, and each resource has a fixed capacity. A task can only start once all its predecessors have finished and sufficient resource capacity is available. The objective considered in this work is the minimization of the makespan, i.e., the completion time of the last task.

\paragraph{Instance Creation.}
Generated instances are resampled from the RCPSP benchmark~\cite{PelleauMTB13} by applying random modifications. We specify the number of tasks and resources, resource capacities, task durations, resource demands, and precedence relations. Task durations, resource demands, and resource capacities are scaled by independent random factors drawn from predefined ranges, while preserving feasibility by ensuring that each resource capacity is at least as large as the maximum demand of any task. Additionally, instances can be extended or reduced by adding or removing tasks and precedence relations, with care taken to avoid cycles in the precedence graph. We use CP-SAT solver for reference solutions.

\subsubsection{Parallel Machines Scheduling (PMS)}


Parallel Machines Scheduling (PMS) considers a set of independent jobs that must be processed on a set of identical parallel machines. Each job is characterized by a processing time, a release date, and a deadline, and can be assigned to any of the available machines. At most one job can be processed on a machine at any time, and a job may only start after its release date. It minimizes tardiness.

\paragraph{Instance Creation.}
Each generated instance specifies the number of jobs and machines, followed by job-specific processing times, release dates, and deadlines of existing instances~\cite{tanaka2008branch}. To increase instance diversity and allow fine-grained control over problem size and difficulty, we resample the original instances by randomly scaling processing times, release dates, and deadlines within predefined ranges, while preserving feasibility by ensuring that each job’s deadline remains no earlier than its release date plus processing time. In addition, instances can be extended or reduced by adding or removing jobs, with new jobs generated to match the statistical properties of the original data.

\subsubsection{Single-Machine Total Weighted Tardiness (SMTWT)}

The Single Machine Total Weighted Tardiness (SMTWT) problem involves scheduling a set of jobs on one machine. Each job needs to be completed without stopping, and the machine can only work on one job at a time. 

\paragraph{Instance Creation.}
The instances are randomly generated following~\citet{beasley1990or}. For each job, the processing time $p(j)$ is drawn uniformly from $[1,100]$ and the weight $w(j)$ from $[1,10]$. Due dates are generated to control instance difficulty using two parameters: the relative range of due dates (RDD) and the tardiness factor (TF), with values in $\{0.2,0.4,0.6,0.8,1.0\}$. Let $P = \sum_{j=1}^{n} p(j)$. For a given pair $(\text{RDD}, \text{TF})$, each due date $d(j)$ is drawn uniformly from the interval
$
\bigl[P(1-\text{TF}-\text{RDD}/2),\; P(1-\text{TF}+\text{RDD}/2)\bigr].
$

\subsection{Constraint-family Group: COUNT}
\subsubsection{Minimum Dominating Set (MDS)}
The Minimum Dominating Set Problem (MDS) finds the smallest subset of vertices such that every vertex in the graph is either selected or adjacent to a selected vertex.

\paragraph{Instance Creation.}
MDS instances are generated using the similar RWR-based subgraph extraction procedure as in STP.  
We sample subgraphs from the DIMACS10 Street Networks~\cite{Bader2014}. Specifically, we use the following graphs: \textit{Belgium}, \textit{Great Britain}, \textit{Italy}, \textit{Luxembourg}, and \textit{Netherlands}.  
For each MDS instance, we randomly select one of these networks and RWR procedure to obtain a structurally meaningful local subgraph. Starting from a randomly chosen seed node, the random walk is executed for 10,000 steps with a restart probability of \(\alpha = 0.15\). The nodes with the highest visitation frequency are retained to form the subgraph. All sampled graphs are treated as unweighted, undirected, and required to be connected.

\begin{table*}[t]
\centering
\small
\renewcommand{\arraystretch}{1.2}
\setlength{\tabcolsep}{4pt}
\resizebox{\linewidth}{!}{
\begin{tabular}{llcccccccccccc}
\toprule
\textbf{Problem} & \textbf{Tier} &
$\overline{M}$ & $M_{\min}$ & $M_{\max}$ &
$\overline{N}$ & $N_{\min}$ & $N_{\max}$ &
$\overline{D}$ & $D_{\min}$ & $D_{\max}$ &
$\overline{k}$ & $k_{\min}$ & $k_{\max}$ \\
\midrule

\multirow{3}{*}{Set Covering (SCP)}
 & S &  9.92 &  8 & 12 &  9.20 &  6 & 12 & 0.416 & 0.278 & 0.661 & -- & -- & -- \\
 & M & 15.36 & 13 & 18 & 14.40 & 11 & 18 & 0.282 & 0.190 & 0.420 & -- & -- & -- \\
 & L & 21.44 & 19 & 25 & 20.44 & 15 & 25 & 0.208 & 0.150 & 0.275 & -- & -- & -- \\
\midrule

\multirow{3}{*}{Set Packing (SP)}
 & S &  9.78 &  8 & 12 &  9.02 &  6 & 12 & 0.400 & 0.265 & 0.625 & -- & -- & -- \\
 & M & 15.26 & 13 & 18 & 14.26 & 11 & 18 & 0.281 & 0.219 & 0.382 & -- & -- & -- \\
 & L & 21.42 & 19 & 25 & 20.28 & 17 & 25 & 0.207 & 0.150 & 0.318 & -- & -- & -- \\

\midrule
\multirow{3}{*}{Set Partitioning Problem (SPP)}
 & S &  9.88 &  8 & 12 &  30.18 & 19 & 46 & 0.168 & 0.146 & 0.188 & -- & -- & -- \\
 & M & 15.16 & 13 & 18 &  64.44 & 43 & 91 & 0.168 & 0.152 & 0.182 & -- & -- & -- \\
 & L & 22.20 & 19 & 25 & 123.98 & 85 & 163 & 0.172& 0.159 & 0.184 & -- & -- & -- \\

\midrule

\multirow{3}{*}{Hitting Set Problem (HSP)}
 & S &  9.84 &  8 & 12 &  8.90 &  6 & 12 & 0.410 & 0.303 & 0.554 & -- & -- & -- \\
 & M & 15.12 & 13 & 18 & 14.20 & 10 & 18 & 0.285 & 0.209 & 0.426 & -- & -- & -- \\
 & L & 21.86 & 19 & 25 & 20.68 & 17 & 25 & 0.211 & 0.158 & 0.313 & -- & -- & -- \\
\midrule

\multirow{3}{*}{Max $k$-Coverage (MkC)}
 & S &  9.78 &  8 & 12 &  9.20 &  7 & 12 & 0.416 & 0.285 & 0.571 & 1.34 & 1 & 2 \\
 & M & 15.28 & 13 & 18 & 14.16 & 10 & 18 & 0.279 & 0.183 & 0.354 & 2.44 & 2 & 3 \\
 & L & 21.54 & 19 & 25 & 20.54 & 17 & 25 & 0.211 & 0.150 & 0.316 & 3.64 & 3 & 5 \\

\bottomrule
\end{tabular}
}
\caption{Summary statistics of graph-based Set Covering Problem (SCP), Max $k$-Coverage (MkC), Set Packing (SP), Hitting Set Problem (HSP), and Set Partitioning (SPP) instances constructed from ROAD networks across three configurations (S, M, L). $M$ denotes the number of elements, $N$ the number of sets, $D$ the density $nnz/(M\cdot N)$. For MkC, $k$ denotes the budget (maximum number of selected sets).}
\label{tab:set_summary}
\end{table*}


\subsubsection{Set Covering Problem (SCP)}
The Set Covering Problem (SCP) aims to select the minimum number of sets such that their union covers all elements in the universe.


\paragraph{Instance Creation.} 
We generate SCP instances by mapping small graph neighborhoods into set systems, following the approach of \cite{khalil2017learning}. Specifically, we begin with the DIMACS10 \emph{road\_central} network~\cite{Bader2014} and sample connected subgraphs using a random walk with restart (RWR) procedure, as employed in the construction of MDS instances.

Given a sampled subgraph $G = (V,E)$, we define the SCP ground set by treating each vertex $v \in V$ as an element. For each vertex $u \in V$, we form a candidate set by taking the $k$-hop neighborhood of $u$ in $G$ (here $k=2$), so that each set corresponds to the elements (vertices) within distance at most~$k$ of its center. To induce stochastic sparsity, each vertex in this neighborhood is independently retained with probability~$0.7$. If the thinning process removes all vertices, the set is replaced with $\{u\}$, ensuring that every element contributes at least one valid candidate set.
Finally, we discard sets that are too small, remove duplicate sets with identical coverage. Instances in which some elements remain uncovered after set construction are discarded to guarantee feasibility. We finally compute SCP optimal solution by Gurobi.

\subsubsection{Set Packing (SP)}
Set Packing (SP) seeks to choose the maximum number of mutually disjoint sets, i.e., no two selected sets share an element.

\paragraph{Instance Creation.}

We generate SP instances using exactly the same graph-to-set construction pipeline as described in the SCP instance creation, including subgraph sampling, $k$-hop neighborhood formation, stochastic thinning, and duplicate removal. The only difference lies in the optimization objective: we solve the Set Packing problem using Gurobi to maximize the number of selected sets subject to element-disjointness constraints.

\subsubsection{Set Partitioning Problem (SPP)}
The Set Partitioning Problem (SPP) requires selecting a collection of sets such that each element in the universe is covered by exactly one selected set.

\paragraph{Instance Creation.}
We generate SPP instances following a randomized construction procedure inspired by \cite{kadioglu2012non}. Each instance is defined over a universe of $m$ elements and a collection of candidate sets.

We first determine the number of non-singleton sets as
$n = \lfloor f \cdot m \rceil$,
where the scaling factor $f$ is sampled uniformly from $[2,4]$ for small (S), $[3,5]$ for medium (M), and $[4,6]$ for large (L) instances.  
For each set $S_j$ ($j=1,\dots,n$), we sample a per-set density
$d_j \sim \mathrm{Uniform}(0.1, 0.3)$
and fix its cardinality as $|S_j| = \lfloor d_j \cdot m \rceil$.

Elements in each set are selected using a clustered sampling strategy: a random element is chosen as a center, and additional elements are sampled \emph{without replacement} with probabilities that decay with their index distance to the center according to a Gaussian weighting, inducing local overlap structure among sets. Duplicate sets with identical element coverage are removed.

Set costs follow the \emph{tails} cost model introduced in \cite{kadioglu2012non}. Specifically, each set $S_j$ is assigned a cost
$c_j = u_j \cdot |S_j|$,
where $u_j \sim \mathrm{Uniform}(1,100)$.

Since the randomly generated sets may not form an exact partition, we additionally include a singleton set $\{i\}$ for each element $i$, each assigned a large fixed cost of $10^{4}$. This guarantees that every instance admits at least one feasible exact partition.

\subsubsection{Hitting Set Problem (HSP)}
The Hitting Set Problem aims to choose the smallest number of elements such that each set in the collection contains at least one selected element.

\paragraph{Instance Creation.}
We generate HSP instances using the same graph-based set creation pipeline as in the SCP instance creation. We solve the Hitting Set Problem with Gurobi.

\subsubsection{Max $k$-Coverage (MkC)}
Max $k$-Coverage (MkC) is a constrained version of Maximum Coverage: select at most $k$ sets to maximize the number of covered elements.

\paragraph{Instance Creation.}
MkC instances are built on the same graph-based set creation pipeline as in the SCP instance creation; however, the optimization objective differs. Instead of covering all elements with as few sets as possible, MkC restricts the solution to a fixed number of sets and maximizes the number of elements they collectively cover.

After sets are constructed from an RWR-sampled subgraph (the same as SCP), we set a budget $k$ that limits how many sets may be selected.
This budget is determined automatically as a fixed ratio (i.e., 20\%) of the number of available sets. We get the reference solution using Gurobi.

\subsubsection{Generalized Assignment Problem (GAP)}
The Generalized Assignment Problem (GAP) assigns \(m\) tasks to \(n\) agents such that each task is assigned to exactly one agent, agent capacities are respected, and the total assignment cost is minimized. Each assignment \((i,j)\) incurs a cost \(c_{ij}\) and consumes \(a_{ij}\) units of the limited resource of agent \(i\), which has capacity \(b_i\). The objective is to minimize \(\sum_{i=1}^{n}\sum_{j=1}^{m} c_{ij} x_{ij}\) subject to \(\sum_{i=1}^{n} x_{ij}=1\) for all tasks \(j\), \(\sum_{j=1}^{m} a_{ij} x_{ij}\le b_i\) for all agents \(i\), and \(x_{ij}\in\{0,1\}\).

\paragraph{Instance Creation.}
We generate GAP instances by sub-sampling from the OR-Library GAP dataset \cite{chu1997genetic}. For a target size of \(m\) tasks, the number of agents \(n\) is determined proportionally to maintain a challenging agent-to-task ratio. Specifically, \(n\) is set as \(\max(2, \mathrm{round}(m/\alpha))\), where the ratio \(\alpha\) is randomly sampled from a uniform distribution \(\mathcal{U}(2.5, 3.5)\). We then randomly select \(m\) tasks and \(n\) agents from a larger source instance and extract the corresponding \(n \times m\) sub-matrices for assignment costs and resource consumptions, along with the capacity sub-vector for the \(n\) selected agents. Ground truth optimal solutions are obtained by solving the MILP formulation for the assignment problem with capacity constraints, using the Gurobi optimizer.

\subsubsection{Uncapacitated Facility Location (UFLP)}

The Uncapacitated Facility Location Problem (UFLP) chooses which facilities to open and assigns each customer to an open facility to minimize opening plus service costs.

\paragraph{Instance Creation.}
UFLP instances are constructed by sub-sampling from the Bilde–Krarup benchmark \cite{bilde1977sharp} to preserve its original cost structures. For each generated instance, we first define the number of customers, $n_c$, and determine the number of candidate facilities, $n_f$, based on a facility-to-customer density ratio $\alpha$. Specifically, we sample $\alpha$ uniformly from the interval $[2.0, 2.67]$ ($2.0$ and $2.67$ are the ratio in the original dataset) and set $n_f = \text{round}(n_c / \alpha)$. We then randomly sample $n_c$ customers and $n_f$ facilities from the source data, extracting the associated opening and connection cost sub-matrices. Ground truth optimal solutions are obtained using the Gurobi mixed-integer programming solver.

\subsubsection{Capacitated Facility Location (CFLP)}

The Capacitated Facility Location Problem (CFLP) selects a subset of facilities to open and assigns customers to open facilities to minimize the sum of facility opening costs and assignment costs, while ensuring that the total demand assigned to each facility does not exceed its capacity.

\paragraph{Instance Creation.}
Similar to UFLP, CFLP instances are generated by sub-sampling from the existing benchmark \cite{holmberg1999exact}. The sampling of customers $n_c$ and facilities $n_f$ based on the density ratio $\alpha$ follows the same protocol described above. In addition to opening and connection costs, we extract the corresponding facility capacities and customer demands. Ground truth optimal solutions are also obtained using Gurobi.

\subsubsection{\texorpdfstring{$p$}{p}-Median Facility Location (PMED)}

The \texorpdfstring{$p$}{p}-median problem selects exactly \(p\) facility locations and assigns each customer to its chosen facility to minimize the total assignment cost.

\paragraph{Instance Creation.}
Similar to UFLP, $p$-median instances are constructed by sub-sampling from a specific dataset used in the classical OR literature \cite{beasley1985note}. For each generated instance with $n$ vertices, we determine the number of medians, $p$, by scaling the original parameter proportionally to the new problem size (maintaining the original ratio $p/N$). We randomly sample $n$ vertices and extract the corresponding pairwise shortest path distances, computed on the original graph, to form the cost matrix. Ground truth optimal solutions are solved by Gurobi.

\subsubsection{\texorpdfstring{$p$}{p}-Center Facility Location (PCENTER)}

The \texorpdfstring{$p$}{p}-center problem selects exactly \(p\) facility locations and assigns each customer to a chosen facility to minimize the maximum assignment distance.

\paragraph{Instance Creation.}
The generation of $p$-center instances is the same as the $p$-median facility location problem, but with a different Gurobi model for modeling the different objective.

\subsubsection{Maximum Independent Set (MIS)}
The Maximum Independent Set problem (MIS) seeks the largest subset of vertices in which no two selected vertices share an edge.

\paragraph{Instance Creation.}
We generate MIS instances by extracting subgraphs from the DIMACS10 Redistricting Graph collection \cite{Bader2014}, which provides high-quality real-world regional adjacency graphs used in political redistricting. In our experiments, we use the five state-level graphs (\textit{vt2010}, \textit{wa2010}, \textit{wi2010}, \textit{wv2010}, \textit{wy2010}) as the sources for random-walk subgraph sampling. For each MIS instance, we randomly select one of these networks and apply the same RWR procedure as in MDS to obtain a structurally meaningful local subgraph. To avoid trivial structures that are either nearly empty or overly dense, we set density constraints specific to each difficulty tier:
\[
\begin{array}{ll}
\text{Set-S:} & 0.15 \le \rho \le 0.55, \\
\text{Set-M:} & 0.12 \le \rho \le 0.50, \\
\text{Set-L:} & 0.08 \le \rho \le 0.45.
\end{array}
\]

Ground truth optimal solutions are obtained using Gurobi.

\subsubsection{Minimum Vertex Cover (MVC)}
The Minimum Vertex Cover Problem (MVC) aims to find the smallest set of vertices such that every edge in the graph has at least one endpoint selected.

\paragraph{Instance Creation.}
MVC instances are generated using the same subgraph extraction pipeline as MIS.  
We sample subgraphs from the DIMACS10 Redistricting Graph collection \cite{Bader2014}, using the same random walk with restart (RWR) procedure as in MDS.
We apply the same density constraints as MIS.

\subsubsection{Maximum Clique Problem (MCP)}
The Maximum Clique Problem (MCP) searches for the largest subset of vertices that are all pairwise adjacent on a graph. 

\paragraph{Instance Creation.}
We again employ the same RWR-based sampling strategy used for MDS. For MCP, we use citation graphs from the DIMACS10 Co-author and Citation Networks collection~\cite{Bader2014}. We restrict MCP instances to the following density ranges:

\[
\begin{array}{ll}
\text{Set-S:} & 0.30 \le \rho \le 0.70, \\
\text{Set-M:} & 0.25 \le \rho \le 0.65, \\
\text{Set-L:} & 0.20 \le \rho \le 0.60.
\end{array}
\]

\subsubsection{Knapsack Problem (KP)}

The Knapsack Problem (KP) seeks a subset of \(n\) items to maximize total value subject to a capacity constraint. Each item \(i\) has a value \(v_i\) and a weight \(w_i\), and the decision variable \(x_i \in \{0,1\}\) indicates whether the item is selected. The objective is to maximize \(\sum_{i=1}^{n} v_i x_i\) subject to \(\sum_{i=1}^{n} w_i x_i \le C\), where \(C\) is the knapsack capacity.

\paragraph{Instance Creation.}
For all KP instances, the creation follows the uncorrelated instance class defined in \citet{pisinger2005hard}.
Item weights and profits are generated i.i.d.\ from the discrete uniform distribution on \(\{1,\dots,100\}\). The knapsack capacity is set to $C = \alpha \sum_{i=1}^{n} w_i,$
with \(\alpha = 0.5\). The reference optimal solution is obtained through Gurobi.



\begin{table*}[thb]
\centering
\small
\renewcommand{\arraystretch}{1.2}
\setlength{\tabcolsep}{5pt}
\resizebox{\linewidth}{!}{

\begin{tabular}{llcccccccccc}
\toprule
\textbf{Problem} & \textbf{Tier} &
$\overline{|V|}$ & $|V|_{\min}$ & $|V|_{\max}$ &
$\overline{|E|}$ & $|E|_{\min}$ & $|E|_{\max}$ &
$\overline{D}$   & $D_{\min}$   & $D_{\max}$ \\
\midrule
\multirow{3}{*}{Maximum Independent Set (MIS)}
 & S &  9.74 &  8 & 12 & 16.38 & 10 & 24 & 0.391 & 0.273 & 0.536 \\
 & M & 15.24 & 13 & 18 & 28.70 & 19 & 40 & 0.267 & 0.203 & 0.333 \\
 & L & 21.70 & 19 & 25 & 43.24 & 30 & 54 & 0.195 & 0.156 & 0.247 \\
\midrule
\multirow{3}{*}{Minimum Vertex Cover (MVC)}
 & S & 10.04 &  8 & 12 & 16.70 & 10 & 27 & 0.372 & 0.258 & 0.500 \\
 & M & 15.24 & 13 & 18 & 27.94 & 18 & 37 & 0.260 & 0.206 & 0.333 \\
 & L & 21.64 & 19 & 25 & 42.62 & 32 & 56 & 0.192 & 0.147 & 0.240 \\
\midrule
\multirow{3}{*}{Max Clique (MCP)}
 & S & 10.00 &  8 & 12 & 19.72 &  9 & 42 & 0.427 & 0.303 & 0.691 \\
 & M & 15.44 & 13 & 18 & 42.96 & 22 & 98 & 0.384 & 0.250 & 0.647 \\
 & L & 22.18 & 19 & 25 & 73.04 & 39 &129 & 0.314 & 0.203 & 0.579 \\
\midrule
\multirow{3}{*}{Graph Coloring (GCP)}
 & S & 10.16 &  8 & 12 & 21.22 & 10 & 39 & 0.448 & 0.306 & 0.679 \\
 & M & 15.54 & 13 & 18 & 43.10 & 21 & 92 & 0.378 & 0.250 & 0.618 \\
 & L & 22.20 & 19 & 25 & 79.06 & 38 &174 & 0.333 & 0.205 & 0.580 \\
\midrule
\multirow{3}{*}{Minimum Dominating Set (MDS)}
 & S &  9.92 &  8 & 12 &  9.30 &  7 & 13 & 0.214 & 0.167 & 0.357 \\
 & M & 15.48 & 13 & 18 & 14.94 & 12 & 19 & 0.135 & 0.111 & 0.179 \\
 & L & 22.04 & 19 & 25 & 21.78 & 18 & 26 & 0.095 & 0.080 & 0.126 \\
\midrule
\multirow{3}{*}{Max Cut (MAXCUT)}
 & S & 10.34 &  8 & 12 & 17.26 &  8 & 24 & 0.358 & 0.242 & 0.500 \\
 & M & 15.86 & 13 & 18 & 29.18 & 18 & 41 & 0.250 & 0.158 & 0.333 \\
 & L & 22.72 & 19 & 25 & 44.40 & 24 & 62 & 0.180 & 0.133 & 0.240 \\
\bottomrule

\end{tabular}
}
\caption{Summary statistics of the generated graph for
the Maximum Independent Set (MIS), Minimum Vertex Cover (MVC), Maximum Clique Problem (MCP), Graph Coloring Problem (GCP), Minimum Dominating Set (MDS), and Maximum Cut (MAXCUT) across
the three difficult tiers (S, M, L).
For each tier, we report the average, minimum, and maximum values of the number of vertices ($|V|$), number of edges ($|E|$),
and graph density ($D$).}
\label{tab:graph_summary}
\end{table*}

\subsubsection{Maximum Diversity Problem (MDP)}

The MDP aims to select a subset of \(m\) elements from a ground set of \(n\) elements such that the selected elements are as mutually dissimilar as possible. Given a pairwise distance matrix \(D = (d_{ij})\), the objective is to maximize the total pairwise distance among the selected elements, i.e., \(\sum_{i=1}^{n}\sum_{j=i+1}^{n} d_{ij} x_i x_j\), subject to selecting exactly \(m\) elements: \(\sum_{i=1}^{n} x_i = m\) with \(x_i \in \{0,1\}\).

\paragraph{Instance Creation.}
Maximum Diversity Problem instances are generated by sub-sampling from GKD and SOM sets in MDPLIB~\cite{marti2021mdplib}. For each generated instance, we specify the number of vertices, \(n\), and randomly select a source problem from the dataset. We then sample \(n\) vertices from the full graph and extract the corresponding pairwise distance sub-matrix. The diversity parameter \(m\) (number of elements to select) is scaled proportionally to maintain the original selection ratio: \(m = \max(2, \mathrm{round}(m_{\text{original}} \times n / N))\), where \(N\) is the size of the source instance. Distance matrices are represented as integers. Ground truth optimal solutions are obtained using Gurobi mixed-integer programming with a linearized formulation of the quadratic diversity objective.

\subsubsection{Quadratic Knapsack Problem (QKP)}

QKP extends the 0--1 knapsack by allowing the profit of a selected set of items to include pairwise interaction gains (or penalties). Each item \(i\) has a weight \(a_i\), and the selection must satisfy a single capacity constraint \(C\). The objective is to maximize the total profit consisting of linear terms \(c_i\) and quadratic interaction terms \(c_{ij}\), i.e., \(\sum_{i=1}^{n} c_i x_i + \sum_{i=1}^{n}\sum_{j=i+1}^{n} c_{ij} x_i x_j\), subject to \(\sum_{i=1}^{n} a_i x_i \le C\) and \(x_i \in \{0,1\}\).

\paragraph{Instance Creation.}
QKP instances are constructed by sub-sampling from the QAPLIB benchmark datasets to preserve their original coefficient structures. For each generated instance, we randomly select a subset \(S\) of \(n\) variables from a source problem of size \(N\), and extract the corresponding linear coefficients \(\{c_i\}_{i\in S}\), quadratic coefficients \(\{c_{ij}\}_{i,j\in S,\, i<j}\) (upper triangular), and item weights \(\{a_i\}_{i\in S}\). To maintain the constraint tightness, we scale the capacity proportionally as:
\[
C_{\text{new}} = C_{\text{original}} \times \frac{\sum_{i \in S} a_i}{\sum_{i=1}^{N} a_i},
\]
so that the capacity-to-weight ratio remains consistent with the source instance. The resulting QKP maximizes $ \sum_{i=1}^{n} c_i x_i + \sum_{i=1}^{n} \sum_{j=i+1}^{n} c_{ij} x_i x_j $ subject to \(\sum_{i=1}^{n} a_i x_i \leq C_{\text{new}}\) and \(x_i \in \{0,1\}\). Ground truth optimal solutions are obtained using the Gurobi mixed-integer programming solver with Kaufman--Broeckx linearization.
  
\subsubsection{Maximum Cut Problem (MAXCUT)}
The Maximum Cut Problem (MAXCUT) partitions the vertices into two subsets to maximize the number of edges crossing between them.

\paragraph{Instance Creation.}
We sample subgraphs from the DIMACS10 Redistricting Graph collection \cite{Bader2014}, using RWR as in MDS, enforcing connectivity, and adopting the same density constraints as those in MIS.

\subsection{Constraint-family Group: COMP}
\subsubsection{3-Index Assignment Problem (AP3)}
The 3-Index Assignment Problem (AP3) is a higher-dimensional generalization of the classic linear assignment. It seeks to select $n$ triples $(i, j, k)$ from three disjoint sets of size $n$ such that every element in each set is selected exactly once, minimizing the total cost. Given a cost tensor $C = (c_{ijk})$ where $i, j, k \in \{1, \dots, n\}$, the objective is to find binary variables $x_{ijk}$ that minimize the total cost:
$$ \sum_{i=1}^{n}\sum_{j=1}^{n}\sum_{k=1}^{n} c_{ijk} x_{ijk} $$
subject to the constraints:
\[
\begin{aligned}
&\sum_{j=1}^{n}\sum_{k=1}^{n} x_{ijk} = 1 && \forall i, \\
&\sum_{i=1}^{n}\sum_{k=1}^{n} x_{ijk} = 1 && \forall j, \\
&\sum_{i=1}^{n}\sum_{j=1}^{n} x_{ijk} = 1 && \forall k
\end{aligned}
\]

\paragraph{Instance Creation.}
We generate AP3 instances by sub-sampling from Balas benchmark~\citep{balas1991algorithm}, which provides dense integer cost tensors. For a target dimension $n$, we randomly select $n$ indices from the source instance's dimension $N$ (where $N \ge n$) and extract the corresponding $n \times n \times n$ sub-tensor. This approach maintains the structural correlations present in the original data. Ground truth optimal solutions are computed using a direct MILP formulation solved via Gurobi.


\subsubsection{Quadratic Assignment Problem (QAP)}
The Quadratic Assignment Problem (QAP) seeks an assignment of \(n\) facilities to \(n\) locations that minimizes the total interaction cost induced by facility-to-facility flows and inter-location distances. Given a flow matrix \(F=(f_{ij})\) and a distance matrix \(D=(d_{kl})\), the objective is to find a permutation \(\pi\) that minimizes \(\sum_{i=1}^{n}\sum_{j=1}^{n} f_{ij}\, d_{\pi(i)\pi(j)}\).

\paragraph{Instance Creation.}
We generate QAP instances by sub-sampling from the QAPLIB dataset \cite{burkard1997qaplib} to preserve realistic flow and distance structures. For a target problem size \(n\), we randomly select \(n\) locations from a source instance and extract the corresponding \(n \times n\) flow and distance sub-matrices. Ground truth optimal solutions are obtained by solving an MILP formulation, specifically utilizing the Kaufman--Broeckx linearization, via the Gurobi optimizer.

\subsubsection{Graph Coloring Problem (GCP)}
The Graph Coloring Problem (GCP) assigns the minimum number of colors to vertices so that no adjacent vertices share the same color. 

\paragraph{Instance Creation.}
GCP uses the same RWR extraction method and size ranges as MDS.  
Instances are generated from the same DIMACS10 Co-author and Citation Networks as MCP~\cite{Bader2014}.
We adopt the same density constraints as MCP.

\subsubsection{Cutwidth Minimization Problem (CMP)}
The Cutwidth Minimization Problem (CMP) seeks a linear layout of a graph's vertices such that the maximum number of edges crossing any cut between consecutive vertices in the layout is minimized. Given a graph $G=(V, E)$ with $n$ vertices, the goal is to find a bijection $\pi: V \to \{1, \dots, n\}$ that minimizes the cutwidth:
$$ \max_{1 \le i < n} \left| \{(u, v) \in E : \pi(u) \le i < \pi(v)\} \right| $$

\paragraph{Instance Creation.}
We generate CMP instances by extracting induced subgraphs from CMPLIB graphs~\citep{marti2012benchmark}. For a target size $n$, we randomly select $n$ vertices and include all edges connecting these vertices (induced subgraph). To ensure non-trivial instances, we initialize the selection with the endpoints of a randomly chosen edge before sampling the remaining vertices. Ground truth optimal solutions are computed using an MILP formulation solved via Gurobi.

\subsubsection{Linear Ordering Problem (LOP)}
The Linear Ordering Problem (LOP) asks for a permutation (ordering) of a set of \(n\) nodes that maximizes the total weight of arcs that are consistent with the order. Given a weighted directed complete graph represented by a cost matrix \(C \in \mathbb{R}^{n \times n}\), the objective is to find a permutation \(\pi\) over \(\{1,\dots,n\}\) that maximizes $\sum_{i<j} C_{\pi_i,\pi_j}$,
i.e., the sum of weights from nodes placed earlier in the order to nodes placed later.

\paragraph{Instance Creation.}
For each LOP instance, a base problem is sampled uniformly at random from the LOPLIB dataset~\cite{marti2012benchmark}, where each entry specifies a single square cost matrix \(C\). Given a target size \(n\), the extractor sub-samples \(n\) nodes uniformly without replacement from the original matrix indices, and induces the corresponding principal submatrix \(C' = C[\mathcal{I},\mathcal{I}]\), where \(\mathcal{I}\) is the selected index set. No additional normalization or transformation is applied; the submatrix values are inherited directly from the source instance.

Each sampled instance is solved using a mixed-integer programming formulation with Gurobi. 

\clearpage
\section{Details of Natural-Language Contextualization}
\label{sec:appendix-context}
This section provides additional details of the natural-language contextualization pipeline used to construct textual instances from numerical CO instances.
The pipeline transforms a numerical instance into a scenario-grounded textual input.
At a high level, the pipeline proceeds by 1) generating and verifying diverse scenario descriptions for a target problem, 2) constructing reusable scenario-level instruction templates, 3) deriving scenario-specific field specifications and natural-language input templates, and 4) instantiating numerical instances under sampled surface input formats. Across all steps in Appendix~\ref{sec:appendix-scenario}--\ref{sec:appendix-instance},
we use GPT-5-mini for LLM generation and checking.

\subsection{Scenario Pool Construction (Generate--Filter--Verify)}
\label{sec:appendix-scenario}

\paragraph{What this step produces.}
For each CO problem $P$, we build a scenario pool $\mathcal{S}_P = \{ s_{P,1}, s_{P,2}, \dots, s_{P,N_P} \}$ where each $s_{P,i}$ is a short, everyday story that describes the same task as $P$.
Across the pool, scenarios should look different on the surface (domain, wording),
but they keep the same decision variables, constraints, and objective.

\noindent
\textbf{Generate candidates.}
We iteratively expand $\mathcal{S}_P$.
In each round, a generator LLM is prompted to produce $K{=}20$ candidate scenarios for $P$
using the prompt in Prompt~\ref{fig:prompt-scenario-generation}.
The candidates are encouraged to be brief, and to only change narrative details.

\noindent
\textbf{Filter near-duplicates.}
To ensure the diversity of the scenarios,
we remove candidates that are too similar to existing ones in $\mathcal{S}_P$.
We embed each text using SentenceTransformer \texttt{all-MiniLM-L6-v2} and compute cosine similarity.
A candidate $c$ is discarded if $\max_{s \in \mathcal{S}_P} \cos\big(e(c), e(s)\big) > 0.7.$

\noindent
\textbf{Verify task consistency.}
For the remaining candidates, we run an LLM as a semantic verifier.
Given a candidate scenario $s$ and the formal CO task description $\mathcal{D}_P$,
the verifier checks whether $s$ still describes the same CO task:
it must preserve the original problem structure, so we reject scenarios that add extra rules, change the objective or constraints.
The verifier prompt is shown in Prompt~\ref{fig:prompt-verifier}.

\noindent
\textbf{Stopping rule.}
Only candidates that pass verification are added to $\mathcal{S}_P$.
We repeat the generate--filter--verify loop until $|\mathcal{S}_P| \ge N_P$,
and we set $N_P{=}50$ in our construction.

\paragraph{Example.}
Taking CVRP as an example, below are two accepted scenarios that differ in narrative but encode the same problem structure:
\begin{itemize}
    \item \emph{I need the postmaster to assign parcel delivery runs from the distribution center so that every address is served exactly once, each run starts and ends at the center, no van is loaded beyond its capacity, and the overall route length is minimized.}
    \item \emph{The school district planner should design bus routes from the depot that pick up each stop exactly once, begin and finish at the depot without exceeding bus seating limits, and keep the total driving distance as low as possible.}
\end{itemize}

\noindent
This scenario pool $\mathcal{S}_P$ is used in the next step to build reusable instruction templates.

\subsection{Instruction Template Construction}
\label{sec:appendix-template}

\paragraph{What this step produces.}
For each verified scenario $s \in \mathcal{S}_P$, we build a reusable instruction template
$\mathcal{T}_s$.
The key idea is to separate the \emph{scenario wording} from the \emph{instance data},
so we do not need to regenerate a full instruction for every instance.

\noindent
\textbf{Two-stage prompting: intro then continuation.}
Template construction uses two prompts.

\smallskip
\noindent\textbf{Stage 1: scenario intro variants.}
Given the scenario text $s$, the problem $P$, and the formal description $\mathcal{D}_P$,
the LLM rewrites $s$ into multiple short introductions that keep the same task but vary in tone and framing $(s,\, P,\, \mathcal{D}_P) \;\longrightarrow\;
\big\{\,\iota^{(1)}_s,\dots,\iota^{(8)}_s\,\big\}.$
We then randomly select one introduction $\iota^{*}_s$ and denote it as the template prefix $\mathcal{I}^{\mathrm{intro}}_s \;=\; \iota^{*}_s.$ The prompt used in this stage is shown in Prompt~\ref{fig:prompt-casual-intro}.

\smallskip
\noindent\textbf{Stage 2: instruction continuation with output requirements.}
Next, conditioned on the chosen intro $\mathcal{I}^{\mathrm{intro}}_s$,
the LLM generates a continuation that 1) explains what to do with instance data and
2) states the required output format $\mathcal{O}_P$: $(\mathcal{I}^{\mathrm{intro}}_s,\, P,\, \mathcal{D}_P,\, \mathcal{O}_P)
\;\longrightarrow\;
\kappa_s.$ The prompt used in this stage is shown in Prompt~\ref{fig:prompt-instruction-continuation}.

\noindent
\textbf{Final template with a single instance placeholder.}
We form the final instruction template by inserting exactly one placeholder for instance input:
\[
\mathcal{T}_s
\;=\;
\mathcal{I}^{\mathrm{intro}}_s
\;\Vert\;
\langle \texttt{INSTANCE\_INPUT} \rangle
\;\Vert\;
\kappa_s.
\]
At dataset construction time, we replace \texttt{INSTANCE\_INPUT} with instance-specific content
(rendered in one of the surface formats described in the next subsection).

\subsection{Field Specifications and Surface Input Formats}
\label{sec:appendix-style}

\paragraph{What this step produces.}
This step produces scenario-specific rendering assets that control how instance data is shown: 1) a scenario-specific field specification $\mathcal{H}_s$ for structured formats (JSON/CSV/Markdown); 2) a natural-language rendering template $\mathcal{N}_s$ for the free-text (NL) format. Both assets change only how the input is presented to mimic realistic user inputs, not the underlying numbers,
constraints, or feasible solution space.

\noindent
\textbf{Scenario-specific field specification for structured formats.}
We begin with a generic structured field definition $\mathcal{H}$, shared across scenarios for the same problem family.
Concretely, $\mathcal{H}$ specifies 1) which input fields exist, 2) short explanations of what each field means,
and 3) how fields are grouped. For example, \texttt{global\_fields} that describe the whole instance
(e.g., total capacity) versus \texttt{item\_fields} that repeat for each record (e.g., each location).
Given a verified scenario $s \in \mathcal{S}_P$ and its chosen intro $\mathcal{I}^{\mathrm{intro}}_s$,
we ask the LLM to rewrite $\mathcal{H}$ into a scenario-aligned specification $(\mathcal{H},\, s,\, \mathcal{I}^{\mathrm{intro}}_s)
\;\longrightarrow\;
\mathcal{H}_s.$

$\mathcal{H}_s$ preserves the same structure as $\mathcal{H}$ (same groups and same fields),
but replaces technical names and descriptions with story-consistent ones.
For example, in a CVRP-style delivery narrative, a generic field like \texttt{depot} may be renamed to \texttt{hub\_node\_id}
and described as ``the hub where routes start and end'';
\texttt{capacity} may become \texttt{truck\_capacity\_units} with a description such as
``maximum load each truck can carry'';
and an item-level field like \texttt{demand} may become \texttt{store\_demand\_units} to match a ``store replenishment'' story.
Similarly, coordinate fields \texttt{x,y} can be renamed to \texttt{coord\_x, coord\_y} while keeping the same meaning
(used to compute travel distances).
These changes are purely semantic \emph{relabeling}: they make structured inputs easier to read under the scenario,
without changing any data content.

\noindent
The prompt used to generate field specifications is shown in Prompt~\ref{fig:prompt-field-interpretation}.

\noindent
\textbf{Rendering into JSON/CSV/Markdown.}
Using $\mathcal{H}_s$, we render a numerical instance $I_{P,j}$ into different structured surface formats $(\mathcal{H}_s,\, I_{P,j},\, f)
\;\longrightarrow\;
\text{input}^{(f)}_{P,j}, f \in \{\textsc{json},\,\textsc{csv},\,\textsc{md}\}$.
This step only changes labels and presentation; it does not change any values, constraints, or identifier relationships.

\noindent
\textbf{Natural-language rendering for the NL format.}
For the natural-language surface format, we also build an item-level template $\mathcal{N}_s$
that turns structured instance data into readable sentences.
Concretely, $\mathcal{N}_s$ specifies:
1) a short header for the overall instance information (e.g., Total locations: 12; hub node: 0; truck capacity: 30 units.), 2) a per-record sentence pattern (e.g., Location B at coordinates (4, 7) has demand 10 units.),
and 3) an optional footer for constraint specification (e.g., Every route must start and end at the hub and never exceed the truck capacity {capacity}.).
We then render $I_{P,j}$ into free text by applying $\mathcal{N}_s$ record-by-record,
so the input resembles what a user might type while remaining fully consistent with the structured data.

\subsection{Instance-Level Contextualization and Representation Diversification}
\label{sec:appendix-instance}

\paragraph{What this step produces.}
This step produces the final dataset instances: each numerical instance is paired with
a natural-language instruction template and a chosen input surface format.

\noindent
\textbf{Notation for numerical instances.}
Let $\mathcal{I}_P = \{ I_{P,1}, \dots, I_{P,M_P} \}$ be the set of numerical instances for $P$.
(We use $\mathcal{D}_P$ for the formal task description, and $\mathcal{O}_P$ for the output format.)

\noindent
\textbf{Sampling a scenario template.}
For each numerical instance $I_{P,j}$, we uniformly sample one scenario index
$u \in \{1,\dots,N_P\}$ and take the corresponding template $\mathcal{T}_{s_{P,u}}$.
This chooses the narrative framing and output requirements, while $I_{P,j}$ supplies the instance data.

\noindent
\textbf{Choosing an input surface format.}
To reflect variation in user input, we randomly sample a surface format $f \in \{\textsc{nl},\,\textsc{json},\,\textsc{csv},\,\textsc{md}\}$, and render $I_{P,j}$ using the natural-language rendering template
$\mathcal{N}_{s_{P,u}}$ if $f=\textsc{nl}$;
otherwise, we use the structured specification $\mathcal{H}_{s_{P,u}}$
to obtain $\text{input}^{(f)}_{P,j}$.

\noindent
\textbf{Identifier variations.}
We also vary naming schemes for identifiers (e.g., numeric indices vs.\ alphabetical labels)
to increase surface diversity.
We enforce strict consistency: identifiers used in the input always match those expected in the output.
These changes affect only presentation and do not modify the underlying optimization problem.

\noindent
\textbf{Final assembly.}
We then fill the template placeholder \texttt{INSTANCE\_INPUT} with the rendered input,
producing the final contextualized instance $d_{P,j} \;=\; \big(\mathcal{T}_{s_{P,u}},\; \text{input}^{(f)}_{P,j}\big)$ for LLM evaluation. An example of contextualized instance for CVRP (with CSV format) is presented by Table~\ref{fig:context-example-postal-csv}.

\onecolumn

\begin{center}
\begin{promptboxstar}{Scenario Generation Prompt}

\setstretch{1.08}
\setlength{\parskip}{2pt}
\setlength{\parindent}{0pt}

\setlist[itemize]{leftmargin=1.2em, itemsep=0.6pt, topsep=0.6pt, parsep=0pt}
\setlist[enumerate]{leftmargin=1.4em, itemsep=0.8pt, topsep=0.6pt, parsep=0pt}

You are an expert in creating diverse and realistic real-world scenarios that correspond to optimization problems. You will be given:
\begin{itemize}
  \item \textbf{\$problem\_type} (internal only; don’t mention it)
  \item \textbf{\$task\_description} (internal only)
\end{itemize}

\textbf{Your reasoning}

Before writing anything, silently determine:
\begin{enumerate}
  \item \textbf{What the decision-maker chooses}
  (e.g., selecting, ordering, grouping, assigning, matching, scheduling, placing)
  \item \textbf{The single optimization objective}
  (its direction, how it is aggregated, and its meaning)
  \item \textbf{All feasibility constraints}
  (no more, no fewer than described in the description)
\end{enumerate}

Once inferred, \textbf{these elements must remain completely fixed}:
\begin{itemize}
  \item Do \textbf{not} add, remove, change, relax, or tighten any constraint.
  \item Do \textbf{not} alter the type, direction, or structure of the objective.
  \item Do \textbf{not} introduce any additional rules (time windows, capacities, precedence, fairness, deadlines, multi-resources, spatial layout, etc.) unless explicitly present in the canonical description.
  \item The feasible set and the difficulty of the problem must stay exactly the same.
  \item Only the \emph{surface narrative} may change---not the underlying mathematical meaning.
\end{itemize}

\textbf{Scenario transformation rules}

\emph{Semantic preservation}

You may change the real-world context and nouns, but:
\begin{itemize}
  \item Each item must correspond to a concrete entity in the story.
  \item The meaning of the decision remains identical.
  \item The objective must preserve its mathematical form.
  \item All constraints must preserve their exact structure.
\end{itemize}

\emph{Objective wording}

You may phrase the objective in natural language (e.g., ``keep the overall effort as small as possible''),
but the mathematical nature of the objective must stay the same.

\textbf{What to generate}

Produce \textbf{exactly \texttt{\$k}} natural-language instructions.
Each instruction must:
\begin{itemize}
  \item Be one or two sentences.
  \item Be a realistic, everyday request.
  \item Encode the \textbf{same} decision structure, objective, and constraints as the canonical problem.
  \item Omit all additional constraints; include all canonical ones.
  \item Contain \textbf{no digits or ordinal words} in the prose.
  \item Be independently interpretable and solvable.
\end{itemize}

\textbf{Strict output format}

\begin{COTBlock}
1. ...
2. ...
\end{COTBlock}
No additional text.

\end{promptboxstar}
\captionof{prompt}{Prompt used to generate diverse natural-language decision-making scenarios.}
\label{fig:prompt-scenario-generation}
\end{center}

\begin{center}
\begin{promptboxstar}{Problem-Consistency Verification Prompt}

\setstretch{1.08}
\setlength{\parskip}{2pt}
\setlength{\parindent}{0pt}

\setlist[itemize]{leftmargin=1.2em, itemsep=0.6pt, topsep=0.6pt, parsep=0pt}
\setlist[enumerate]{leftmargin=1.4em, itemsep=0.8pt, topsep=0.6pt, parsep=0pt}

You are a task type verifier for optimization benchmark generation.

\textbf{Target Problem}

The expected problem type is: \textbf{expected\_problem}

Problem description: ``problem\_description''

\textbf{Verification Task}

Determine whether the candidate instruction fundamentally describes a standard expected\_problem problem.

\textbf{Rules}

\emph{YES if:}
\begin{itemize}
  \item Core decision structure is equivalent to the described formulation.
  \item Exactly one constraint and one objective.
  \item Real-world setting may vary but math structure is unchanged.
  \item Objective/constraint synonyms are acceptable (e.g., budget $\approx$ cost $\approx$ capacity).
\end{itemize}

\emph{NO if:}
\begin{itemize}
  \item Extra structural complexity (multiple constraints, dependencies, time windows, multi-objective).
  \item Deviation from the standard structure.
\end{itemize}

Do not treat objective wording as an extra constraint.

\textbf{Candidate Instruction}

\begin{COTBlock}
text
\end{COTBlock}

\textbf{Required Output}

\begin{COTBlock}
Answer: YES or NO
Reason: brief justification
\end{COTBlock}

\end{promptboxstar}
\vspace{-15pt}
\captionof{prompt}{Verifier prompt used to check whether a generated scenario preserves the target optimization problem.}
\label{fig:prompt-verifier}
\end{center}

\begin{center}
\begin{promptboxstar}{Casual Scenario Introduction Rewriting Prompt}

\setstretch{1.08}
\setlength{\parskip}{2pt}
\setlength{\parindent}{0pt}

\setlist[itemize]{leftmargin=1.2em, itemsep=0.6pt, topsep=0.6pt, parsep=0pt}
\setlist[enumerate]{leftmargin=1.4em, itemsep=0.8pt, topsep=0.6pt, parsep=0pt}

You are helping to rewrite a task description so it sounds casual, human, and conversational.

You will be given:
\begin{itemize}
  \item \textbf{problem type} (internal only; not to be mentioned)
  \item \textbf{task description} (internal only)
  \item \textbf{base text}: a stiff scenario that should be rewritten in a more natural style
\end{itemize}

Your task is to produce \textbf{eight different rewritten introductions}, each satisfying the following requirements.

\textbf{Casual scenario intro}

For each version:
\begin{itemize}
  \item Rewrite the meaning of the base text in natural, everyday language.
  \item Implicitly reflect the core task described in the task description, but do so in an informal, story-like way.
  \item The rewritten version must clearly convey:
  \begin{itemize}
    \item what situation is taking place,
    \item what decision needs to be made,
    \item what makes one decision better than another,
    \item how that notion of ``better'' is determined,
    \item and what practical requirements must be respected (e.g., nothing can be left out or duplicated).
  \end{itemize}
  \item All of this should be communicated naturally through the scenario itself, without explicitly mentioning objectives, constraints, optimization, formulas, or technical terms.
  \item Do not mention the problem type, internal terminology, or any formal modeling concepts.
  \item Each version must start differently (e.g., I…, We…, There is…, Someone…, Recently…, Many people…).
  \item Do not address the reader directly or imply that the reader is a character in the scenario.
  \item Keep the tone human, conversational, and slightly varied across versions.
\end{itemize}

\textbf{Reference to upcoming instance details}

Each version must casually mention that the concrete details of the situation will be shown below.

\textbf{Diversity requirements}

\begin{itemize}
  \item Number the versions using lettered markers (A), (B), (C), etc.
  \item Each version must sound noticeably different in opening style, tone, and framing.
  \item Avoid reusing similar sentence structures or phrasing patterns.
  \item All versions must remain faithful to the meaning of the base text.
\end{itemize}

\textbf{Output rules}

\begin{itemize}
  \item Output exactly eight versions.
  \item Do not use JSON or structured formats.
  \item Do not include placeholders.
  \item Do not use quotation marks or code formatting.
\end{itemize}

\end{promptboxstar}

\captionof{prompt}{Prompt used to rewrite formal task descriptions into multiple casual, human-sounding scenario introductions while preserving the underlying task semantics.}
\label{fig:prompt-casual-intro}
\end{center}

\begin{center}
\begin{promptboxstar}{Instruction Continuation with Structured Output Prompt}

\setstretch{1.08}
\setlength{\parskip}{2pt}
\setlength{\parindent}{0pt}

\setlist[itemize]{leftmargin=1.2em, itemsep=0.6pt, topsep=0.6pt, parsep=0pt}
\setlist[enumerate]{leftmargin=1.4em, itemsep=0.8pt, topsep=0.6pt, parsep=0pt}

You are continuing a casual, human-sounding task description.

You will be given:
\begin{itemize}
  \item \textbf{problem type} (internal only; not to be mentioned)
  \item \textbf{task description} (internal only)
  \item \textbf{output format}: the exact JSON structure the final answer must follow
\end{itemize}

Below is a draft that already covers the first part of the description, including a casual scenario introduction and a reference to the instance details.

\textbf{Existing introduction}

\begin{COTBlock}
[Partial introduction text omitted here]
\end{COTBlock}

Do not change or repeat anything from the existing introduction.
Your task is to add the remaining part that follows naturally from it.

\textbf{Bringing up the JSON format casually}

When writing the continuation:
\begin{itemize}
  \item Work the mention of the JSON format into the flow in a relaxed, conversational way, as if casually noting how the response should be written.
  \item After that casual mention, include a JSON block that follows exactly the same structure as the given output format.
\end{itemize}

For the JSON block:
\begin{itemize}
  \item The top-level JSON keys must remain exactly identical to those in the output format.
  \item For nested placeholder keys and values:
  \begin{itemize}
    \item If a placeholder looks like an identifier-style label (e.g., depot identifiers, node identifiers, item identifiers), rename only the value prefix to better match the story context.
    \item Keep all JSON keys unchanged.
    \item Keep the overall JSON structure unchanged.
    \item Keep the placeholder format unchanged (it must remain a placeholder, not a concrete value).
  \end{itemize}
  \item If the placeholder values are not identifier-style (e.g., vectors, coordinates, counts, costs, times, or booleans), do not rename or modify them.
\end{itemize}

After presenting the JSON block:
\begin{itemize}
  \item Give a short, easygoing explanation of what each part of the JSON represents in the context of the story.
  \item Keep the explanation informal and non-technical, more like describing a form than a data schema.
  \item Make clear that the JSON is only a sketch of the expected structure, not the actual answer.
\end{itemize}

Finally, gently remind that all identifiers must be used exactly as they appear in the instance input, with no renaming and no new labels.
Valid identifiers include:
\begin{itemize}
  \item plain numbers such as ``1'' or ``23'',
  \item single capital letters such as ``A'' or ``B'',
  \item a capital letter followed by digits such as ``A1'' or ``X7''.
\end{itemize}

\textbf{Critical constraints}

\begin{itemize}
  \item Do not repeat or alter the existing introduction.
  \item Do not mention the problem type or any technical optimization terminology.
  \item Keep the style consistent with the introduction: casual, conversational, and human.
  \item Do not number any part of the continuation.
  \item Do not write lines starting with digits and a period.
\end{itemize}

\end{promptboxstar}

\captionof{prompt}{Prompt used to continue a casual task description with a structured JSON output while preserving tone, narrative flow, and schema constraints.}
\label{fig:prompt-instruction-continuation}
\end{center}

\begin{center}
\begin{promptboxstar}{Scenario-Specific Field Interpretation Prompt}

\setstretch{1.08}
\setlength{\parskip}{2pt}
\setlength{\parindent}{0pt}

\setlist[itemize]{leftmargin=1.2em, itemsep=0.6pt, topsep=0.6pt, parsep=0pt}
\setlist[enumerate]{leftmargin=1.4em, itemsep=0.8pt, topsep=0.6pt, parsep=0pt}

You are an expert at interpreting generic structured-field definitions within a specific natural-language scenario.

You will receive:
\begin{itemize}
  \item Task name
  \item Task description
  \item Scenario text describing the concrete instance
  \item A precomputed instruction introduction that sets the tone, role, and framing
  \item A generic hint JSON defining the structured fields
\end{itemize}

\textbf{Your task}

Rewrite the hint JSON so that every field has a scenario-specific meaning.

You must interpret the fields \emph{as they would be understood under the instruction introduction style} provided above.
In particular:
\begin{itemize}
  \item The implied role, perspective, and constraints from the introduction must be respected.
  \item Field meanings should align with how a solver would understand the task after reading that introduction.
\end{itemize}

The \textbf{top-level structure must remain identical} to the input hint:
\begin{itemize}
  \item Keep all existing top-level keys exactly as they appear.
  \item Preserve all field groups (e.g., global fields, item fields, facility item fields).
  \item If the input is flat, keep it flat.
  \item Do not add, remove, or rename any top-level keys.
\end{itemize}

\textbf{How to rewrite each field}

Each field object in the input has the form:
\begin{COTBlock}
{
  "name": "field_name",
  "description": "generic meaning"
}
\end{COTBlock}

For every field object in every group, output an object with exactly the following structure:
\begin{COTBlock}
{
  "name": "field_name",
  "new_name": "scenario_specific_snake_case",
  "scenario_description": "scenario-specific interpretation"
}
\end{COTBlock}

The following rules must be followed:
\begin{itemize}
  \item The \texttt{name} field must be copied exactly from the input.
  \item The \texttt{new\_name} must be a concise snake\_case label that is meaningful in the scenario and unique within its group.
  \item The \texttt{scenario\_description} must clearly explain how the field should be understood \emph{within the scenario text}.
  \item Do not include the original generic description or any extra keys.
\end{itemize}

\textbf{Allowed placeholders}

Copy the \texttt{allowed\_placeholders} field from the input \emph{exactly}, without any modification.

\textbf{Output (strict)}

Output exactly one JSON object:
\begin{itemize}
  \item The top-level keys and their ordering must match the input.
  \item All field groups must be rewritten as specified above.
  \item Do not include any text outside the JSON object.
\end{itemize}

\end{promptboxstar}

\captionof{prompt}{Prompt used to reinterpret generic structured-field hints into scenario-specific semantics while preserving the original schema.}
\label{fig:prompt-field-interpretation}
\end{center}

\begin{center}
\begin{promptboxstar}{Item-Level Natural Language Template Generation Prompt}

\setstretch{1.08}
\setlength{\parskip}{2pt}
\setlength{\parindent}{0pt}

\setlist[itemize]{leftmargin=1.2em, itemsep=0.6pt, topsep=0.6pt, parsep=0pt}
\setlist[enumerate]{leftmargin=1.4em, itemsep=0.8pt, topsep=0.6pt, parsep=0pt}

You will create reusable natural-language templates for verbalizing item-level input in an optimization problem.

All generated text must follow the tone, perspective, and style implied by:
\begin{itemize}
  \item the task type,
  \item the instruction template,
  \item the field-shape hint describing the structured input.
\end{itemize}

\textbf{Output mode}

If the hint contains exactly one item-field group named \texttt{item\_fields} and no other item-field groups, use the \emph{single-group mode}:
\begin{COTBlock}
{"line_template": "...",
  "header": "...",
  "footer": "..."}
\end{COTBlock}

If the hint contains multiple item-field groups, use the \emph{multi-group mode}:
\begin{COTBlock}
{"<group1>_line_template": "...",
  "<group2>_line_template": "...",
  "...": "...",
  "header": "...",
  "footer": "..."}
\end{COTBlock}

Each group name must match the corresponding item-field group name in the hint exactly.

\textbf{Template rules (strict)}

\emph{Allowed placeholders}

Only use placeholders that appear in:
\begin{itemize}
  \item the list of allowed placeholders in the hint,
  \item global field names (for the header and footer),
  \item item-field names of the relevant group (for its line template).
\end{itemize}

Use placeholders only in the form \texttt{\{field\_name\}}.
Do not use scenario-specific names, aliases, or rewritten field names as placeholders.
Do not invent any additional fields or placeholder names.

\emph{Line templates}

For each item-field group:
\begin{itemize}
  \item Include \emph{all} placeholders from that group at least once.
  \item Do not include placeholders from other item groups or from global fields.
  \item Write a short, natural sentence describing exactly one item.
  \item Match the tone, perspective, and style of the instruction template.
  \item Do not introduce new meanings or constraints.
\end{itemize}

\emph{Header}

If global fields exist:
\begin{itemize}
  \item Include each global placeholder at least once.
  \item Exclude all item-level placeholders.
  \item Write the header as a direct continuation of the instruction template, as if it were the next line in the same explanation.
  \item Preserve the exact narrative perspective and tone of the instruction template.
  \item Keep the header concise and scenario-consistent.
\end{itemize}

If no global fields exist, the header must be an empty string.

\emph{Footer}

The footer is optional and may be an empty string.
If present:
\begin{itemize}
  \item Do not include any item-level placeholders.
  \item Use global placeholders only if they also appear in the header.
  \item Maintain the same tone and narrative perspective as the instruction template.
  \item Provide a brief, natural closing remark.
\end{itemize}

\end{promptboxstar}

\captionof{prompt}{Prompt used to generate reusable item-level natural language templates that verbalize structured inputs while strictly preserving schema and narrative consistency.}
\label{fig:prompt-item-template}
\end{center}

\begin{center}
\begin{promptboxstar}{Illustrative Contextualized Instance Example (CVRP)}

\setstretch{1.05}
\setlength{\parskip}{2pt}
\setlength{\parindent}{0pt}

\begin{COTBlock}
We’re coordinating the day’s routes so a handful of carriers can hand over every piece of mail one time and then return to the post office. The choice is which streets and houses each carrier covers, keeping each route’s total mail within the capacity of that carrier’s bag. The goal is to keep the overall driving low: take each carrier’s route length, add them up, and the plan with the lowest total miles is best. The concrete stop-by-stop details appear below.

# total_stops_including_post_office=9
# post_office_stop_id=A
# carrier_mailbag_capacity=100
stop_id,map_x_coordinate,map_y_coordinate,mail_items_to_deliver
A,83,76,0
B,50,5,20
C,100,14,26
D,21,0,17
...
... (remaining stops omitted)

Also, when you send the planned routes back, please use this simple JSON shape so it's easy to read and parse:

{
  "solution": [[post_office_id, house_id, ... , post_office_id], [post_office_id, house_id, ... , post_office_id], ...]
}

This shows the general layout: "solution" is a list of routes, each inner list is the sequence of stops a single carrier will follow (starting and ending at the post office). It's just a sketch of the expected shape --- not the actual answer.

Please make sure to use the exact identifiers from the instance input --- do not rename them or invent new labels. 
- for example: "Valid identifiers look like plain numbers such as “1” or “23”, single capital letters like “A” or “B”, or a capital letter followed by digits like “A1” or “X7”."
\end{COTBlock}

\end{promptboxstar}

\captionof{table}{Contextualized instance example under a postal-delivery narrative.}
\label{fig:context-example-postal-csv}
\end{center}

\twocolumn

\clearpage

\section{Experiment Configurations}
\label{sec:config}
\paragraph{Parameter Settings.} 
To ensure a fair evaluation across both reasoning and non-reasoning models, as well as reasoning models operating in non-reasoning modes, we describe our parameter settings below, which aim to allow each model to perform at its full capability.

For standard chat LLMs, we set the decoding temperature to 0. For reasoning-oriented LLMs, we do not specify a temperature. The maximum output token number is a crucial parameter for evaluation, as solving CO problems typically requires a large number of tokens for reasoning. Specifically, for DeepSeek and all proprietary LLMs, we evaluate them by calling their official APIs while setting the maximum output token to the maximum allowed value by their API service, i.e., 8,192 for DeepSeek-V3.2 (standard) and 64K for Claude-Sonnet-4.5 (with a thinking budget of 60k). For DeepSeek-V3.2 (reasoning), o4-mini (reasoning), GPT-5.1 (reasoning), and Gemini-3-Flash (reasoning), we do not specify the maximum output token, allowing the API to utilize its full supported output range. For Llama-4-Maverick-Instruct and Qwen3-235B-Instruct, we utilize the Google Vertex AI platform for LLM calling and set the maximum output token count to 8,192 and 16,384, respectively, i.e., the maximum allowed value. 
For other models, we use the API service of the OpenRouter Platform for LLM calling, and we set the maximum output token to the highest value allowed by the platform for each corresponding model: 16384 for Qwen3-14B, 65536 for Ministral3-14B, 65536 for Nemotron3-Nano-30B-A3B, 65536 for MiMo-V2-Flash, and 30K for Grok-4.1-Fast (reasoning).

For models that support explicit reasoning mode, we configure the reasoning-related parameters. Specifically, for Gemini-3-Flash, we set the thinking level to high. For OpenAI models, we enable reasoning and set the reasoning effort to high for o4-mini, and to medium for GPT-5.1.


\paragraph{Prompting Strategies. }For all LLMs with explicit reasoning mode, we directly invoke the model with NLCO instances, as they will autonomously generate the thinking processes. For the standard chat LLMs, we append the prompt "Please think step by step, output your reasoning trace, and then output the JSON in the required format" to the textual instances, guiding the LLMs to generate the chain of thought for solving CO problems. 

\section{Additional Results}
\label{sec:app_res}
\paragraph{Performance profiles.}
To complement aggregate statistics (e.g., ALOG) on a benchmark with various tasks, we also use \textbf{performance profiles}~\citep{dolan2002benchmarking}, a standard tool in optimization for comparing methods across heterogeneous tasks without allowing a small number of difficult cases to dominate the overall picture. 
Let $f_{i,m}$ denote the objective returned by LLM model $m$ on instance $i$ (lower is better; infeasible outputs are set to $f_{i,m}=+\infty$), and let $f_i^{\min}=\min_m f_{i,m}$ be the best value achieved on instance $i$ across all LLMs. Assuming objectives are nonnegative, we define the per-instance performance ratio
\(r_{i,m}=\frac{f_{i,m}+\varepsilon}{f_i^{\min}+\varepsilon}\) with a small \(\varepsilon>0\) to avoid division by zero, and plot for each model the cumulative fraction of instances where it is within a factor \(\gamma\) of the best (with $N$ denotes the number of instances):
\[
\rho_m(\gamma)=\frac{1}{N}\left|\left\{ i : r_{i,m}\le \gamma \right\}\right|,\qquad \gamma\ge 1.
\]

A curve that is higher, especially near $\gamma=1$, indicates a model that more consistently matches the strongest observed performance across instances.

\begin{figure*}[thp]
        \centering
        \includegraphics[width=\linewidth]{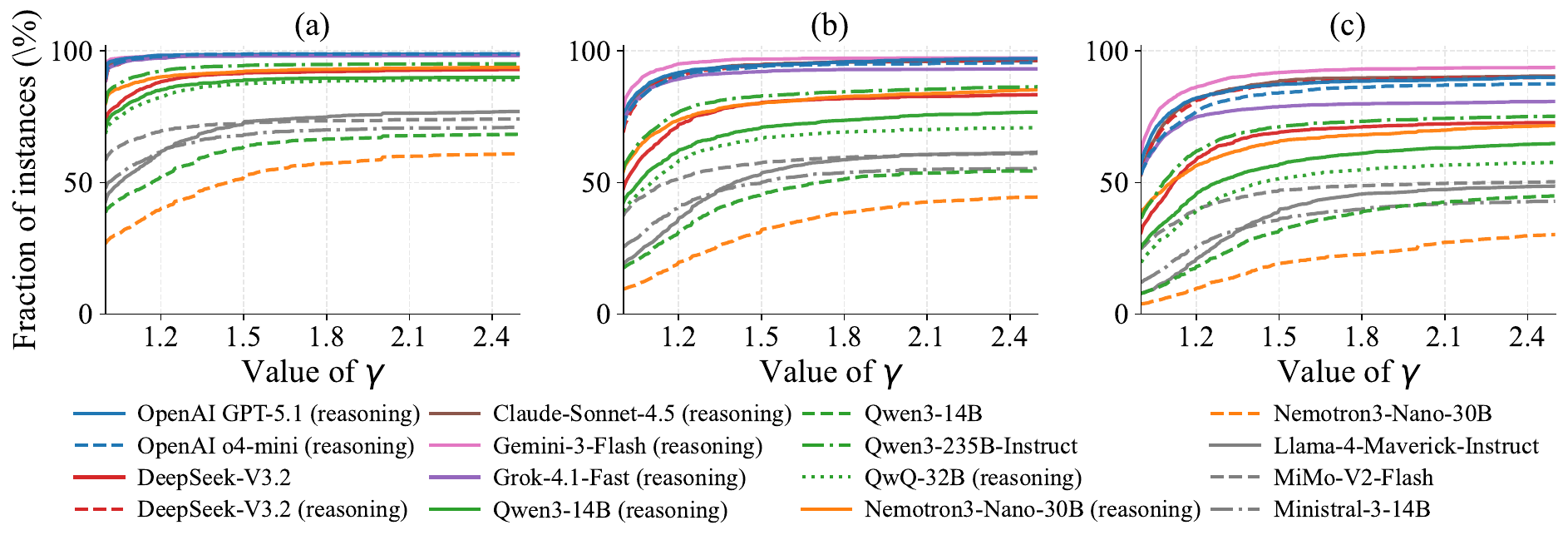}
        \captionof{figure}{LLM performance profiles on different NLCO difficulty tiers. (a) Set-S; (b) Set-M; (c) Set-L.}
        \label{fig:performance_profile}
\end{figure*}

\begin{figure*}
        \includegraphics[width=\linewidth]{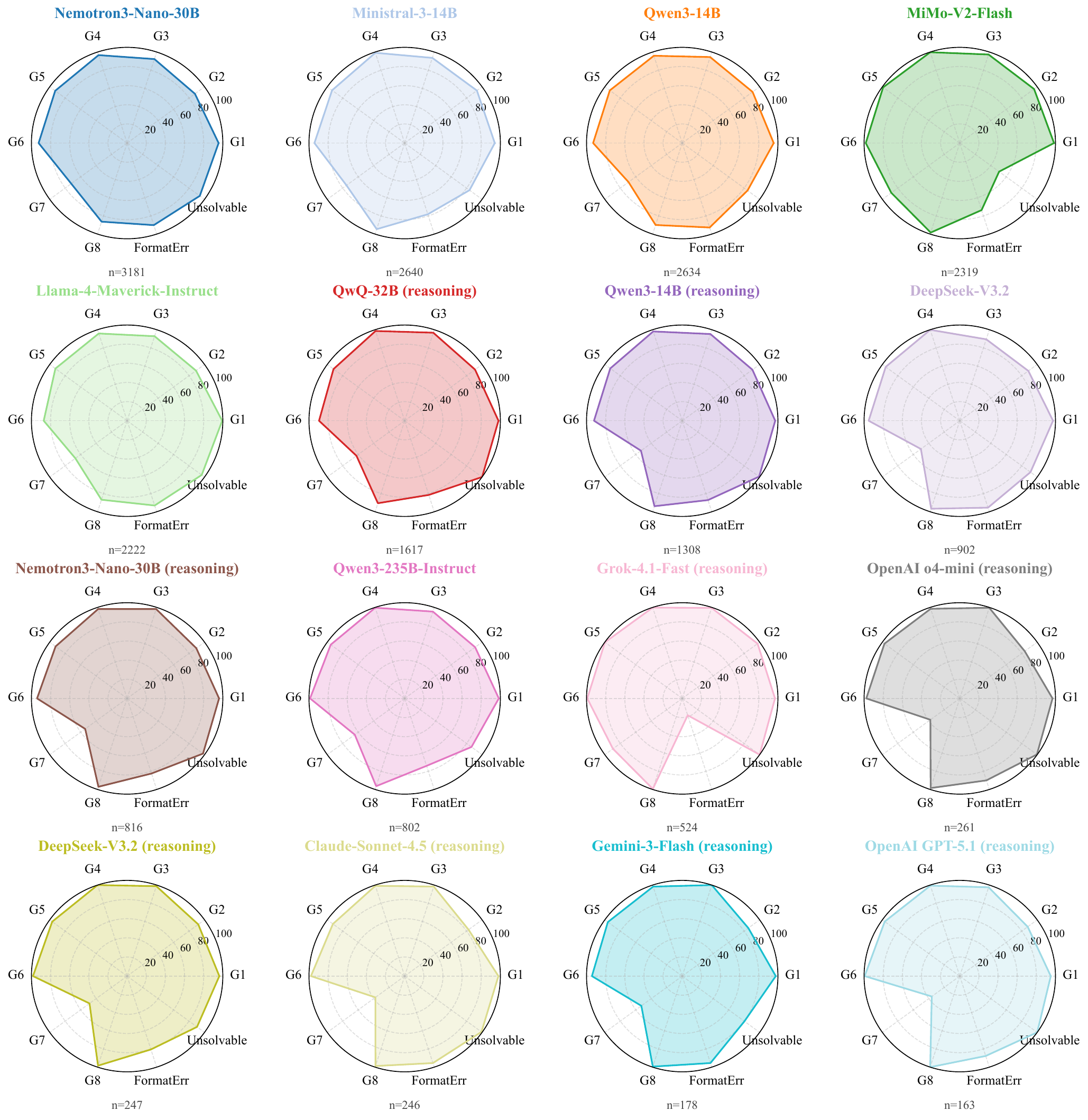}
        \captionof{figure}{Per-model radar charts showing infeasibility mode distribution across LLMs; larger filled area indicates lower frequency of that mode, with $n$ denoting sample count per model.}
        \label{fig:rader}
\end{figure*}

\begin{table*}
\centering
\small
\setlength{\tabcolsep}{0pt}
\renewcommand{\arraystretch}{1.15}

\begin{tabularx}{\textwidth}{@{}>{\raggedright\arraybackslash}X@{\hskip 6pt}*{9}{C{0.062\textwidth}}@{}}
\toprule
\multirow{2}{*}{\textbf{Problem}} &
\multicolumn{3}{c}{\textbf{AFR}} &
\multicolumn{3}{c}{\textbf{Acc.}} &
\multicolumn{3}{c}{\textbf{tok.}} \\
\cmidrule(lr){2-4} \cmidrule(lr){5-7} \cmidrule(l){8-10}
& \textbf{Set-S} & \textbf{Set-M} & \textbf{Set-L}
& \textbf{Set-S} & \textbf{Set-M} & \textbf{Set-L}
& \textbf{Set-S} & \textbf{Set-M} & \textbf{Set-L} \\
\midrule
\midrule

Traveling Salesman Problem
& \cellcolor{BlueMain!70!BlueLight}\textcolor{white}{0.97}
& \cellcolor{BlueMain!60!BlueLight}\textcolor{white}{0.93}
& \cellcolor{BlueMain!60!BlueLight}\textcolor{white}{0.91}
& \cellcolor{BlueMain!5!BlueLight}\textcolor{black}{0.21}
& \cellcolor{BlueMain!5!BlueLight}\textcolor{black}{0.16}
& \cellcolor{BlueMain!5!BlueLight}\textcolor{black}{0.05}
& \cellcolor{gray!19}{12.8K}
& \cellcolor{gray!22}{15.3K}
& \cellcolor{gray!20}{14.0K} \\

Prize-Collecting TSP
& \cellcolor{BlueMain!60!BlueLight}\textcolor{white}{0.93}
& \cellcolor{BlueMain!70!BlueLight}\textcolor{white}{0.96}
& \cellcolor{BlueMain!60!BlueLight}\textcolor{white}{0.93}
& \cellcolor{BlueMain!10!BlueLight}\textcolor{black}{0.47}
& \cellcolor{BlueMain!5!BlueLight}\textcolor{black}{0.15}
& \cellcolor{BlueLight}\textcolor{black}{0.03}
& \cellcolor{gray!20}{13.8K}
& \cellcolor{gray!23}{15.9K}
& \cellcolor{gray!24}{16.6K} \\

Orienteering Problem
& \cellcolor{BlueMain!50!BlueLight}\textcolor{white}{0.89}
& \cellcolor{BlueMain!50!BlueLight}\textcolor{white}{0.85}
& \cellcolor{BlueMain!50!BlueLight}\textcolor{white}{0.83}
& \cellcolor{BlueMain!40!BlueLight}\textcolor{white}{0.76}
& \cellcolor{BlueMain!20!BlueLight}\textcolor{black}{0.52}
& \cellcolor{BlueMain!10!BlueLight}\textcolor{black}{0.40}
& \cellcolor{gray!17}{10.6K}
& \cellcolor{gray!19}{13.5K}
& \cellcolor{gray!22}{15.1K} \\

Capacitated VRP
& \cellcolor{BlueMain!70!BlueLight}\textcolor{white}{0.96}
& \cellcolor{BlueMain!60!BlueLight}\textcolor{white}{0.94}
& \cellcolor{BlueMain!50!BlueLight}\textcolor{white}{0.85}
& \cellcolor{BlueMain!10!BlueLight}\textcolor{black}{0.42}
& \cellcolor{BlueMain!5!BlueLight}\textcolor{black}{0.09}
& \cellcolor{BlueLight}\textcolor{black}{0.02}
& \cellcolor{gray!18}{11.2K}
& \cellcolor{gray!19}{12.1K}
& \cellcolor{gray!19}{11.9K} \\

TSP with Time Windows
& \cellcolor{BlueMain!40!BlueLight}\textcolor{white}{0.72}
& \cellcolor{BlueMain!10!BlueLight}\textcolor{black}{0.46}
& \cellcolor{BlueMain!5!BlueLight}\textcolor{black}{0.19}
& \cellcolor{BlueMain!10!BlueLight}\textcolor{black}{0.42}
& \cellcolor{BlueMain!5!BlueLight}\textcolor{black}{0.10}
& \cellcolor{BlueLight}\textcolor{black}{0.01}
& \cellcolor{gray!23}{15.6K}
& \cellcolor{gray!27}{19.1K}
& \cellcolor{gray!28}{19.3K} \\

Pickup-and-Delivery Problem
& \cellcolor{BlueMain!30!BlueLight}\textcolor{white}{0.69}
& \cellcolor{BlueMain!10!BlueLight}\textcolor{black}{0.49}
& \cellcolor{BlueMain!10!BlueLight}\textcolor{black}{0.34}
& \cellcolor{BlueMain!10!BlueLight}\textcolor{black}{0.40}
& \cellcolor{BlueMain!5!BlueLight}\textcolor{black}{0.07}
& \cellcolor{BlueLight}\textcolor{black}{0.00}
& \cellcolor{gray!22}{14.7K}
& \cellcolor{gray!24}{16.8K}
& \cellcolor{gray!23}{15.7K} \\

Minimum Latency Problem
& \cellcolor{BlueMain!60!BlueLight}\textcolor{white}{0.91}
& \cellcolor{BlueMain!60!BlueLight}\textcolor{white}{0.91}
& \cellcolor{BlueMain!50!BlueLight}\textcolor{white}{0.87}
& \cellcolor{BlueMain!5!BlueLight}\textcolor{black}{0.13}
& \cellcolor{BlueLight}\textcolor{black}{0.03}
& \cellcolor{BlueMain!5!BlueLight}\textcolor{black}{0.05}
& \cellcolor{gray!18}{11.5K}
& \cellcolor{gray!19}{13.5K}
& \cellcolor{gray!19}{13.3K} \\

Quadratic Shortest Path Problem
& \cellcolor{BlueMain!70!BlueLight}\textcolor{white}{0.98}
& \cellcolor{BlueMain!70!BlueLight}\textcolor{white}{0.96}
& \cellcolor{BlueMain!60!BlueLight}\textcolor{white}{0.92}
& \cellcolor{BlueMain!60!BlueLight}\textcolor{white}{0.94}
& \cellcolor{BlueMain!50!BlueLight}\textcolor{white}{0.86}
& \cellcolor{BlueMain!30!BlueLight}\textcolor{white}{0.65}
& \cellcolor{gray!9}{5.13K}
& \cellcolor{gray!11}{7.28K}
& \cellcolor{gray!19}{12.9K} \\

Steiner Tree Problem
& \cellcolor{BlueMain!50!BlueLight}\textcolor{white}{0.82}
& \cellcolor{BlueMain!30!BlueLight}\textcolor{white}{0.68}
& \cellcolor{BlueMain!30!BlueLight}\textcolor{white}{0.61}
& \cellcolor{BlueMain!30!BlueLight}\textcolor{white}{0.67}
& \cellcolor{BlueMain!10!BlueLight}\textcolor{black}{0.48}
& \cellcolor{BlueMain!10!BlueLight}\textcolor{black}{0.41}
& \cellcolor{gray!14}{9.02K}
& \cellcolor{gray!22}{15.0K}
& \cellcolor{gray!25}{17.1K} \\

Steiner Forest Problem
& \cellcolor{BlueMain!50!BlueLight}\textcolor{white}{0.84}
& \cellcolor{BlueMain!30!BlueLight}\textcolor{white}{0.68}
& \cellcolor{BlueMain!20!BlueLight}\textcolor{black}{0.57}
& \cellcolor{BlueMain!30!BlueLight}\textcolor{white}{0.69}
& \cellcolor{BlueMain!10!BlueLight}\textcolor{black}{0.47}
& \cellcolor{BlueMain!10!BlueLight}\textcolor{black}{0.33}
& \cellcolor{gray!17}{10.6K}
& \cellcolor{gray!24}{16.8K}
& \cellcolor{gray!29}{20.3K} \\

$k$-Minimum Spanning Tree
& \cellcolor{BlueMain!50!BlueLight}\textcolor{white}{0.88}
& \cellcolor{BlueMain!40!BlueLight}\textcolor{white}{0.74}
& \cellcolor{BlueMain!30!BlueLight}\textcolor{white}{0.62}
& \cellcolor{BlueMain!40!BlueLight}\textcolor{white}{0.76}
& \cellcolor{BlueMain!30!BlueLight}\textcolor{white}{0.60}
& \cellcolor{BlueMain!10!BlueLight}\textcolor{black}{0.41}
& \cellcolor{gray!17}{10.1K}
& \cellcolor{gray!24}{16.8K}
& \cellcolor{gray!28}{19.5K} \\

\midrule

Bin Packing Problem
& \cellcolor{BlueMain!70!BlueLight}\textcolor{white}{0.98}
& \cellcolor{BlueMain!60!BlueLight}\textcolor{white}{0.92}
& \cellcolor{BlueMain!50!BlueLight}\textcolor{white}{0.83}
& \cellcolor{BlueMain!60!BlueLight}\textcolor{white}{0.94}
& \cellcolor{BlueMain!50!BlueLight}\textcolor{white}{0.81}
& \cellcolor{BlueMain!30!BlueLight}\textcolor{white}{0.64}
& \cellcolor{gray!6}{3.04K}
& \cellcolor{gray!11}{6.71K}
& \cellcolor{gray!17}{10.2K} \\

Cutting Stock Problem
& \cellcolor{BlueMain!40!BlueLight}\textcolor{white}{0.72}
& \cellcolor{BlueMain!20!BlueLight}\textcolor{black}{0.58}
& \cellcolor{BlueMain!10!BlueLight}\textcolor{black}{0.48}
& \cellcolor{BlueMain!30!BlueLight}\textcolor{white}{0.65}
& \cellcolor{BlueMain!10!BlueLight}\textcolor{black}{0.48}
& \cellcolor{BlueMain!10!BlueLight}\textcolor{black}{0.34}
& \cellcolor{gray!14}{9.26K}
& \cellcolor{gray!23}{15.5K}
& \cellcolor{gray!27}{18.6K} \\

2D Strip Packing
& \cellcolor{BlueMain!20!BlueLight}\textcolor{black}{0.53}
& \cellcolor{BlueMain!10!BlueLight}\textcolor{black}{0.33}
& \cellcolor{BlueMain!5!BlueLight}\textcolor{black}{0.15}
& \cellcolor{BlueMain!10!BlueLight}\textcolor{black}{0.38}
& \cellcolor{BlueMain!5!BlueLight}\textcolor{black}{0.11}
& \cellcolor{BlueMain!5!BlueLight}\textcolor{black}{0.04}
& \cellcolor{gray!23}{15.8K}
& \cellcolor{gray!30}{20.6K}
& \cellcolor{gray!32}{22.3K} \\

Job-Shop Scheduling Problem
& \cellcolor{BlueMain!40!BlueLight}\textcolor{white}{0.71}
& \cellcolor{BlueMain!10!BlueLight}\textcolor{black}{0.48}
& \cellcolor{BlueMain!10!BlueLight}\textcolor{black}{0.44}
& \cellcolor{BlueMain!20!BlueLight}\textcolor{black}{0.59}
& \cellcolor{BlueMain!5!BlueLight}\textcolor{black}{0.12}
& \cellcolor{BlueLight}\textcolor{black}{0.00}
& \cellcolor{gray!13}{8.54K}
& \cellcolor{gray!19}{15.2K}
& \cellcolor{gray!19}{12.9K} \\

Flow-Shop Scheduling Problem
& \cellcolor{BlueMain!40!BlueLight}\textcolor{white}{0.80}
& \cellcolor{BlueMain!20!BlueLight}\textcolor{black}{0.58}
& \cellcolor{BlueMain!10!BlueLight}\textcolor{black}{0.49}
& \cellcolor{BlueMain!40!BlueLight}\textcolor{white}{0.74}
& \cellcolor{BlueMain!5!BlueLight}\textcolor{black}{0.20}
& \cellcolor{BlueLight}\textcolor{black}{0.01}
& \cellcolor{gray!11}{6.37K}
& \cellcolor{gray!23}{16.1K}
& \cellcolor{gray!23}{15.6K} \\

Open-Shop Scheduling Problem
& \cellcolor{BlueMain!30!BlueLight}\textcolor{white}{0.69}
& \cellcolor{BlueMain!20!BlueLight}\textcolor{black}{0.53}
& \cellcolor{BlueMain!10!BlueLight}\textcolor{black}{0.45}
& \cellcolor{BlueMain!5!BlueLight}\textcolor{black}{0.28}
& \cellcolor{BlueMain!5!BlueLight}\textcolor{black}{0.06}
& \cellcolor{BlueLight}\textcolor{black}{0.01}
& \cellcolor{gray!13}{8.35K}
& \cellcolor{gray!21}{14.4K}
& \cellcolor{gray!20}{13.9K} \\

RCPSP (makespan)
& \cellcolor{BlueMain!30!BlueLight}\textcolor{white}{0.60}
& \cellcolor{BlueMain!10!BlueLight}\textcolor{black}{0.38}
& \cellcolor{BlueMain!10!BlueLight}\textcolor{black}{0.31}
& \cellcolor{BlueMain!20!BlueLight}\textcolor{black}{0.56}
& \cellcolor{BlueMain!5!BlueLight}\textcolor{black}{0.27}
& \cellcolor{BlueMain!5!BlueLight}\textcolor{black}{0.13}
& \cellcolor{gray!11}{6.98K}
& \cellcolor{gray!21}{14.4K}
& \cellcolor{gray!25}{17.5K} \\

Parallel Machines $P\parallel T_{\max}$
& \cellcolor{BlueMain!70!BlueLight}\textcolor{white}{0.96}
& \cellcolor{BlueMain!50!BlueLight}\textcolor{white}{0.89}
& \cellcolor{BlueMain!50!BlueLight}\textcolor{white}{0.81}
& \cellcolor{BlueMain!50!BlueLight}\textcolor{white}{0.81}
& \cellcolor{BlueMain!10!BlueLight}\textcolor{black}{0.46}
& \cellcolor{BlueMain!5!BlueLight}\textcolor{black}{0.19}
& \cellcolor{gray!11}{6.70K}
& \cellcolor{gray!18}{11.8K}
& \cellcolor{gray!21}{14.4K} \\

Single-Machine Total Weighted Tardiness
& \cellcolor{BlueMain!70!BlueLight}\textcolor{white}{0.98}
& \cellcolor{BlueMain!60!BlueLight}\textcolor{white}{0.94}
& \cellcolor{BlueMain!50!BlueLight}\textcolor{white}{0.88}
& \cellcolor{BlueMain!30!BlueLight}\textcolor{white}{0.63}
& \cellcolor{BlueMain!10!BlueLight}\textcolor{black}{0.32}
& \cellcolor{BlueMain!5!BlueLight}\textcolor{black}{0.20}
& \cellcolor{gray!19}{12.1K}
& \cellcolor{gray!22}{14.6K}
& \cellcolor{gray!22}{15.1K} \\

\midrule

Minimum Dominating Set
& \cellcolor{BlueMain!50!BlueLight}\textcolor{white}{0.84}
& \cellcolor{BlueMain!40!BlueLight}\textcolor{white}{0.72}
& \cellcolor{BlueMain!30!BlueLight}\textcolor{white}{0.63}
& \cellcolor{BlueMain!40!BlueLight}\textcolor{white}{0.78}
& \cellcolor{BlueMain!30!BlueLight}\textcolor{white}{0.64}
& \cellcolor{BlueMain!20!BlueLight}\textcolor{black}{0.54}
& \cellcolor{gray!9}{4.91K}
& \cellcolor{gray!14}{8.86K}
& \cellcolor{gray!19}{13.2K} \\

Set Cover Problem
& \cellcolor{BlueMain!70!BlueLight}\textcolor{white}{0.96}
& \cellcolor{BlueMain!50!BlueLight}\textcolor{white}{0.90}
& \cellcolor{BlueMain!50!BlueLight}\textcolor{white}{0.83}
& \cellcolor{BlueMain!60!BlueLight}\textcolor{white}{0.92}
& \cellcolor{BlueMain!50!BlueLight}\textcolor{white}{0.84}
& \cellcolor{BlueMain!40!BlueLight}\textcolor{white}{0.73}
& \cellcolor{gray!6}{3.25K}
& \cellcolor{gray!11}{7.06K}
& \cellcolor{gray!16}{10.8K} \\

Set Packing
& \cellcolor{BlueMain!50!BlueLight}\textcolor{white}{0.88}
& \cellcolor{BlueMain!50!BlueLight}\textcolor{white}{0.83}
& \cellcolor{BlueMain!50!BlueLight}\textcolor{white}{0.83}
& \cellcolor{BlueMain!50!BlueLight}\textcolor{white}{0.86}
& \cellcolor{BlueMain!40!BlueLight}\textcolor{white}{0.79}
& \cellcolor{BlueMain!40!BlueLight}\textcolor{white}{0.75}
& \cellcolor{gray!17}{10.2K}
& \cellcolor{gray!19}{15.2K}
& \cellcolor{gray!26}{17.9K} \\

Set Partitioning Problem
& \cellcolor{BlueMain!50!BlueLight}\textcolor{white}{0.80}
& \cellcolor{BlueMain!30!BlueLight}\textcolor{white}{0.65}
& \cellcolor{BlueMain!10!BlueLight}\textcolor{black}{0.44}
& \cellcolor{BlueMain!30!BlueLight}\textcolor{white}{0.68}
& \cellcolor{BlueMain!10!BlueLight}\textcolor{black}{0.41}
& \cellcolor{BlueMain!5!BlueLight}\textcolor{black}{0.12}
& \cellcolor{gray!18}{11.2K}
& \cellcolor{gray!28}{19.7K}
& \cellcolor{gray!35}{25.4K} \\

Hitting Set Problem
& \cellcolor{BlueMain!70!BlueLight}\textcolor{white}{0.96}
& \cellcolor{BlueMain!60!BlueLight}\textcolor{white}{0.92}
& \cellcolor{BlueMain!40!BlueLight}\textcolor{white}{0.77}
& \cellcolor{BlueMain!50!BlueLight}\textcolor{white}{0.87}
& \cellcolor{BlueMain!40!BlueLight}\textcolor{white}{0.79}
& \cellcolor{BlueMain!20!BlueLight}\textcolor{black}{0.60}
& \cellcolor{gray!7}{3.37K}
& \cellcolor{gray!11}{6.41K}
& \cellcolor{gray!17}{11.0K} \\

Max $k$-Coverage
& \cellcolor{BlueMain!70!BlueLight}\textcolor{white}{0.99}
& \cellcolor{BlueMain!70!BlueLight}\textcolor{white}{0.98}
& \cellcolor{BlueMain!70!BlueLight}\textcolor{white}{0.96}
& \cellcolor{BlueMain!70!BlueLight}\textcolor{white}{0.96}
& \cellcolor{BlueMain!50!BlueLight}\textcolor{white}{0.85}
& \cellcolor{BlueMain!30!BlueLight}\textcolor{white}{0.66}
& \cellcolor{gray!5}{2.17K}
& \cellcolor{gray!12}{7.87K}
& \cellcolor{gray!19}{13.2K} \\

Generalized Assignment Problem
& \cellcolor{BlueMain!70!BlueLight}\textcolor{white}{0.96}
& \cellcolor{BlueMain!60!BlueLight}\textcolor{white}{0.92}
& \cellcolor{BlueMain!40!BlueLight}\textcolor{white}{0.77}
& \cellcolor{BlueMain!50!BlueLight}\textcolor{white}{0.84}
& \cellcolor{BlueMain!30!BlueLight}\textcolor{white}{0.66}
& \cellcolor{BlueMain!10!BlueLight}\textcolor{black}{0.37}
& \cellcolor{gray!8}{4.35K}
& \cellcolor{gray!11}{6.61K}
& \cellcolor{gray!16}{10.5K} \\

Uncapacitated Facility Location
& \cellcolor{BlueMain!70!BlueLight}\textcolor{white}{0.99}
& \cellcolor{BlueMain!70!BlueLight}\textcolor{white}{0.96}
& \cellcolor{BlueMain!60!BlueLight}\textcolor{white}{0.91}
& \cellcolor{BlueMain!60!BlueLight}\textcolor{white}{0.93}
& \cellcolor{BlueMain!40!BlueLight}\textcolor{white}{0.78}
& \cellcolor{BlueMain!20!BlueLight}\textcolor{black}{0.56}
& \cellcolor{gray!6}{2.88K}
& \cellcolor{gray!14}{9.45K}
& \cellcolor{gray!23}{15.6K} \\

Capacitated Facility Location
& \cellcolor{BlueMain!70!BlueLight}\textcolor{white}{0.99}
& \cellcolor{BlueMain!70!BlueLight}\textcolor{white}{0.97}
& \cellcolor{BlueMain!50!BlueLight}\textcolor{white}{0.90}
& \cellcolor{BlueMain!60!BlueLight}\textcolor{white}{0.90}
& \cellcolor{BlueMain!40!BlueLight}\textcolor{white}{0.70}
& \cellcolor{BlueMain!10!BlueLight}\textcolor{black}{0.47}
& \cellcolor{gray!6}{3.02K}
& \cellcolor{gray!14}{8.91K}
& \cellcolor{gray!20}{13.9K} \\

$p$-Median
& \cellcolor{BlueMain!70!BlueLight}\textcolor{white}{0.97}
& \cellcolor{BlueMain!70!BlueLight}\textcolor{white}{0.95}
& \cellcolor{BlueMain!50!BlueLight}\textcolor{white}{0.87}
& \cellcolor{BlueMain!60!BlueLight}\textcolor{white}{0.90}
& \cellcolor{BlueMain!40!BlueLight}\textcolor{white}{0.70}
& \cellcolor{BlueMain!10!BlueLight}\textcolor{black}{0.44}
& \cellcolor{gray!6}{2.79K}
& \cellcolor{gray!15}{9.72K}
& \cellcolor{gray!22}{15.3K} \\

$p$-Center
& \cellcolor{BlueMain!70!BlueLight}\textcolor{white}{0.96}
& \cellcolor{BlueMain!60!BlueLight}\textcolor{white}{0.95}
& \cellcolor{BlueMain!50!BlueLight}\textcolor{white}{0.88}
& \cellcolor{BlueMain!60!BlueLight}\textcolor{white}{0.92}
& \cellcolor{BlueMain!40!BlueLight}\textcolor{white}{0.73}
& \cellcolor{BlueMain!20!BlueLight}\textcolor{black}{0.54}
& \cellcolor{gray!6}{2.85K}
& \cellcolor{gray!12}{7.48K}
& \cellcolor{gray!19}{12.6K} \\

Maximum Independent Set
& \cellcolor{BlueMain!50!BlueLight}\textcolor{white}{0.88}
& \cellcolor{BlueMain!50!BlueLight}\textcolor{white}{0.86}
& \cellcolor{BlueMain!40!BlueLight}\textcolor{white}{0.79}
& \cellcolor{BlueMain!50!BlueLight}\textcolor{white}{0.83}
& \cellcolor{BlueMain!40!BlueLight}\textcolor{white}{0.73}
& \cellcolor{BlueMain!30!BlueLight}\textcolor{white}{0.63}
& \cellcolor{gray!13}{8.59K}
& \cellcolor{gray!19}{12.2K}
& \cellcolor{gray!22}{15.0K} \\

Minimum Vertex Cover
& \cellcolor{BlueMain!50!BlueLight}\textcolor{white}{0.86}
& \cellcolor{BlueMain!40!BlueLight}\textcolor{white}{0.77}
& \cellcolor{BlueMain!30!BlueLight}\textcolor{white}{0.66}
& \cellcolor{BlueMain!50!BlueLight}\textcolor{white}{0.80}
& \cellcolor{BlueMain!40!BlueLight}\textcolor{white}{0.72}
& \cellcolor{BlueMain!20!BlueLight}\textcolor{black}{0.54}
& \cellcolor{gray!14}{9.15K}
& \cellcolor{gray!19}{13.2K}
& \cellcolor{gray!26}{17.7K} \\

Maximum Clique Problem
& \cellcolor{BlueMain!60!BlueLight}\textcolor{white}{0.93}
& \cellcolor{BlueMain!50!BlueLight}\textcolor{white}{0.89}
& \cellcolor{BlueMain!50!BlueLight}\textcolor{white}{0.85}
& \cellcolor{BlueMain!50!BlueLight}\textcolor{white}{0.90}
& \cellcolor{BlueMain!50!BlueLight}\textcolor{white}{0.83}
& \cellcolor{BlueMain!40!BlueLight}\textcolor{white}{0.78}
& \cellcolor{gray!8}{4.40K}
& \cellcolor{gray!10}{6.28K}
& \cellcolor{gray!12}{7.65K} \\

Knapsack Problem
& \cellcolor{BlueMain!70!BlueLight}\textcolor{white}{0.96}
& \cellcolor{BlueMain!60!BlueLight}\textcolor{white}{0.92}
& \cellcolor{BlueMain!50!BlueLight}\textcolor{white}{0.83}
& \cellcolor{BlueMain!50!BlueLight}\textcolor{white}{0.85}
& \cellcolor{BlueMain!30!BlueLight}\textcolor{white}{0.68}
& \cellcolor{BlueMain!20!BlueLight}\textcolor{black}{0.52}
& \cellcolor{gray!11}{6.57K}
& \cellcolor{gray!18}{11.4K}
& \cellcolor{gray!21}{14.5K} \\

Maximum Diversity Problem
& \cellcolor{BlueMain!70!BlueLight}\textcolor{white}{0.97}
& \cellcolor{BlueMain!70!BlueLight}\textcolor{white}{0.95}
& \cellcolor{BlueMain!70!BlueLight}\textcolor{white}{0.95}
& \cellcolor{BlueMain!50!BlueLight}\textcolor{white}{0.88}
& \cellcolor{BlueMain!40!BlueLight}\textcolor{white}{0.72}
& \cellcolor{BlueMain!20!BlueLight}\textcolor{black}{0.56}
& \cellcolor{gray!10}{6.26K}
& \cellcolor{gray!14}{9.81K}
& \cellcolor{gray!18}{11.6K} \\

Quadratic Knapsack Problem
& \cellcolor{BlueMain!60!BlueLight}\textcolor{white}{0.94}
& \cellcolor{BlueMain!50!BlueLight}\textcolor{white}{0.90}
& \cellcolor{BlueMain!50!BlueLight}\textcolor{white}{0.85}
& \cellcolor{BlueMain!40!BlueLight}\textcolor{white}{0.75}
& \cellcolor{BlueMain!10!BlueLight}\textcolor{black}{0.48}
& \cellcolor{BlueMain!5!BlueLight}\textcolor{black}{0.29}
& \cellcolor{gray!13}{8.65K}
& \cellcolor{gray!22}{14.7K}
& \cellcolor{gray!23}{16.1K} \\

Maximum Cut Problem
& \cellcolor{BlueMain!70!BlueLight}\textcolor{white}{0.97}
& \cellcolor{BlueMain!60!BlueLight}\textcolor{white}{0.95}
& \cellcolor{BlueMain!60!BlueLight}\textcolor{white}{0.94}
& \cellcolor{BlueMain!30!BlueLight}\textcolor{white}{0.60}
& \cellcolor{BlueMain!10!BlueLight}\textcolor{black}{0.37}
& \cellcolor{BlueMain!5!BlueLight}\textcolor{black}{0.16}
& \cellcolor{gray!18}{11.7K}
& \cellcolor{gray!24}{16.4K}
& \cellcolor{gray!24}{16.0K} \\

\midrule

Three-Index Assignment
& \cellcolor{BlueMain!60!BlueLight}\textcolor{white}{0.93}
& \cellcolor{BlueMain!50!BlueLight}\textcolor{white}{0.85}
& \cellcolor{BlueMain!40!BlueLight}\textcolor{white}{0.75}
& \cellcolor{BlueMain!30!BlueLight}\textcolor{white}{0.61}
& \cellcolor{BlueMain!5!BlueLight}\textcolor{black}{0.18}
& \cellcolor{BlueLight}\textcolor{black}{0.01}
& \cellcolor{gray!19}{12.5K}
& \cellcolor{gray!27}{18.7K}
& \cellcolor{gray!27}{18.7K} \\

Quadratic Assignment Problem
& \cellcolor{BlueMain!70!BlueLight}\textcolor{white}{0.98}
& \cellcolor{BlueMain!60!BlueLight}\textcolor{white}{0.95}
& \cellcolor{BlueMain!60!BlueLight}\textcolor{white}{0.91}
& \cellcolor{BlueMain!50!BlueLight}\textcolor{white}{0.85}
& \cellcolor{BlueMain!10!BlueLight}\textcolor{black}{0.44}
& \cellcolor{BlueMain!5!BlueLight}\textcolor{black}{0.22}
& \cellcolor{gray!11}{6.97K}
& \cellcolor{gray!22}{14.6K}
& \cellcolor{gray!23}{15.6K} \\

Graph Coloring Problem
& \cellcolor{BlueMain!50!BlueLight}\textcolor{white}{0.81}
& \cellcolor{BlueMain!40!BlueLight}\textcolor{white}{0.71}
& \cellcolor{BlueMain!30!BlueLight}\textcolor{white}{0.61}
& \cellcolor{BlueMain!40!BlueLight}\textcolor{white}{0.76}
& \cellcolor{BlueMain!30!BlueLight}\textcolor{white}{0.67}
& \cellcolor{BlueMain!20!BlueLight}\textcolor{black}{0.55}
& \cellcolor{gray!9}{5.30K}
& \cellcolor{gray!13}{8.69K}
& \cellcolor{gray!19}{12.2K} \\

Cutwidth Minimization Problem
& \cellcolor{BlueMain!70!BlueLight}\textcolor{white}{0.98}
& \cellcolor{BlueMain!70!BlueLight}\textcolor{white}{0.96}
& \cellcolor{BlueMain!60!BlueLight}\textcolor{white}{0.92}
& \cellcolor{BlueMain!50!BlueLight}\textcolor{white}{0.84}
& \cellcolor{BlueMain!30!BlueLight}\textcolor{white}{0.69}
& \cellcolor{BlueMain!10!BlueLight}\textcolor{black}{0.47}
& \cellcolor{gray!11}{6.47K}
& \cellcolor{gray!14}{9.45K}
& \cellcolor{gray!19}{13.2K} \\

Linear Ordering Problem
& \cellcolor{BlueMain!70!BlueLight}\textcolor{white}{0.96}
& \cellcolor{BlueMain!60!BlueLight}\textcolor{white}{0.94}
& \cellcolor{BlueMain!50!BlueLight}\textcolor{white}{0.89}
& \cellcolor{BlueMain!20!BlueLight}\textcolor{black}{0.51}
& \cellcolor{BlueMain!5!BlueLight}\textcolor{black}{0.16}
& \cellcolor{BlueLight}\textcolor{black}{0.02}
& \cellcolor{gray!20}{13.4K}
& \cellcolor{gray!27}{18.9K}
& \cellcolor{gray!29}{20.5K} \\

\bottomrule
\end{tabularx}
\caption{Average feasibility, accuracy, and token usage across LLMs on 43 NLCO tasks.}
\label{tab:all_performance}
\end{table*}

With the performance profiles shown in Figure~5, which report the fraction of instances on which each LLM achieves an objective within a factor $\gamma$
 of the best observed value (and $\gamma = 1$
corresponds to best/tied-best), we find that instance size is the main separator:
from Set-S (a) to Set-L (c), curves systematically shift downward and the gaps between models widen. This indicate that many approaches that look “competitive” on small instances stop being reliably near-best once the problem size grows. Two patterns consistently stand out: 

\begin{itemize}
    \item Heavy-tailed behavior of some LLMs: top-tier models (e.g., OpenAI GPT-5.1, and Gemini-3-Flash) have curves that rise steeply and plateau high even on Set-L, indicating high consistency. In contrast, several baselines (e.g., Llama-4-Maverick-Instruct, MiMo-V2-Flash, Ministral-3-14B) rise more gradually and plateau much lower, especially in (c), which is characteristic of heavy-tailed behavior: they sometimes find decent solutions but frequently fall far from the best.

    \item Reasoning helps more as scale increases: the benefit of reasoning-enabled variants is modest on Set-S (many models already reach high fractions under relaxed $\gamma$), but becomes pronounced on Set-M/L. For example, Nemotron3-Nano-30B (reasoning) is consistently above its non-reasoning counterpart across panels, and the gap is largest on Set-L, suggesting that additional structured computation is particularly valuable for large instances. A similar pattern appears within the Qwen family (e.g., Qwen3-14B (reasoning) vs. Qwen3-14B), where the reasoning variant retains a noticeably higher fraction of near-best solutions as $\gamma$ tightens.
\end{itemize}

\paragraph{Per-model Infeasibility mode Distribution. } According to Figure~\ref{fig:rader}, most LLMs’ infeasibility is concentrated in a small number of modes, rather than evenly spread. In particular, $\text{Global}_7$ is the most common sink (the most pronounced dip in several panels), indicating a persistent structural bottleneck. FormatError also varies substantially by model: some models have a noticeable dip (e.g., Grok) on the FormatErr axis (format compliance is a major source of infeasibility), while others show a much larger radius there, suggesting far fewer parsing/format failures. Finally, unsolvable appears model-dependent: for some LLMs it is relatively rare (large radius), whereas for others (e.g., Mimo-V2-Flash) it forms a clearer dip, consistent with token-budget exhaustion being a non-negligible cause of infeasibility in those systems.

\paragraph{Per-task Performance.} We report the average performance of all LLMs on each NLCO task in Table~\ref{tab:all_performance}, revealing substantial heterogeneity across problems. It can be found that:

1)  \textbf{ Feasibility can remain high while optimality collapses on routing tasks.} For classical routing problems, models often find feasible solutions but struggle to reach optimality, especially as size increases.
For instance, TSP maintains high AFR (0.97/0.93/0.91 from Set-S/M/L) but exhibits very low Acc., suggesting that producing a valid tour is far easier than globally optimizing it.
A similar pattern also holds for CVRP and MLP.

2) \textbf{Some tasks are ``LLM-friendly'' even at larger sizes.}
Several problems retain both high feasibility and relatively strong optimality, suggesting that their structure admits more direct constructive reasoning or local verification.
Examples include Quadratic Shortest Path (AFR 0.98/0.96/0.92; Acc.\ 0.94/0.86/0.65), Bin Packing (AFR 0.98/0.92/0.83; Acc.\ 0.94/0.81/0.64), and graph selection tasks such as Maximum Clique (AFR 0.93/0.89/0.85; Acc.\ 0.90/0.83/0.78), which remain relatively stable compared to routing/scheduling.

3) \textbf{Token usage broadly tracks output complexity and grows with size.}
Across most tasks, tok.\ increases from Set-S to Set-L (e.g., Set Partitioning: 11.2K $\rightarrow$ 25.4K; Linear Ordering: 13.4K $\rightarrow$ 20.5K), reflecting longer solution representations and/or more extensive intermediate reasoning.
Notably, high token usage does not guarantee high accuracy: several tasks remain expensive yet low-Acc.\ (e.g., TSPTW, 2SP), indicating that larger combinatorial spaces and tighter constraints can dominate additional inference-time computation.

\begin{table}[thb]
\centering
\begin{tabular}{lccc}
\toprule
\textbf{Set} & \textbf{1-global} & \textbf{2-global} & $\textbf{Diff.}$ \\
\midrule
Set-S & 88.8\% & 88.1\% & $-0.7\%$ \\
Set-M & 80.9\% & 79.8\% & $-1.1\%$ \\
Set-L & 73.6\% & 70.7\% & $-2.9\%$ \\
\bottomrule
\end{tabular}
\caption{AFR on NLCO tasks with different number of global constraint patterns. 1-global: tasks with 1 global patterns; 2-global: tasks with 2 global patterns. $\textbf{Diff.}$ indicates the difference between two values (2-1)}
\label{tab:feasible_by_size_global}
\end{table}

\paragraph{The Impact of Global Pattern Numbers.} From the NLCO taxonomy presented in Table~\ref{tab:cop-taxonomy}, we can find that some CO problems are featured by one global pattern, while some are with two, which means that an LLM needs to make sure that the produced solutions for these problems are supposed to simultaneously satisfy two distinct types of global constraints. We aggregate the performance of all LLMs and report the average feasibility performance in both scenarios in Table~\ref{tab:feasible_by_size_global}. It can be observed that, on average, the LLMs achieve lower feasibility on CO tasks with 2 global patterns. Meanwhile, with the increase of problem scale (from Set-S to Set-L), the gap in AFR becomes more significant. This observation is intuitive: LLM reasoning is more likely to fail in a more constrained and combinatorial space.

\begin{figure*}[t]
    \centering
    \begin{subfigure}[t]{0.48\linewidth}
        \centering
        \includegraphics[width=\linewidth]{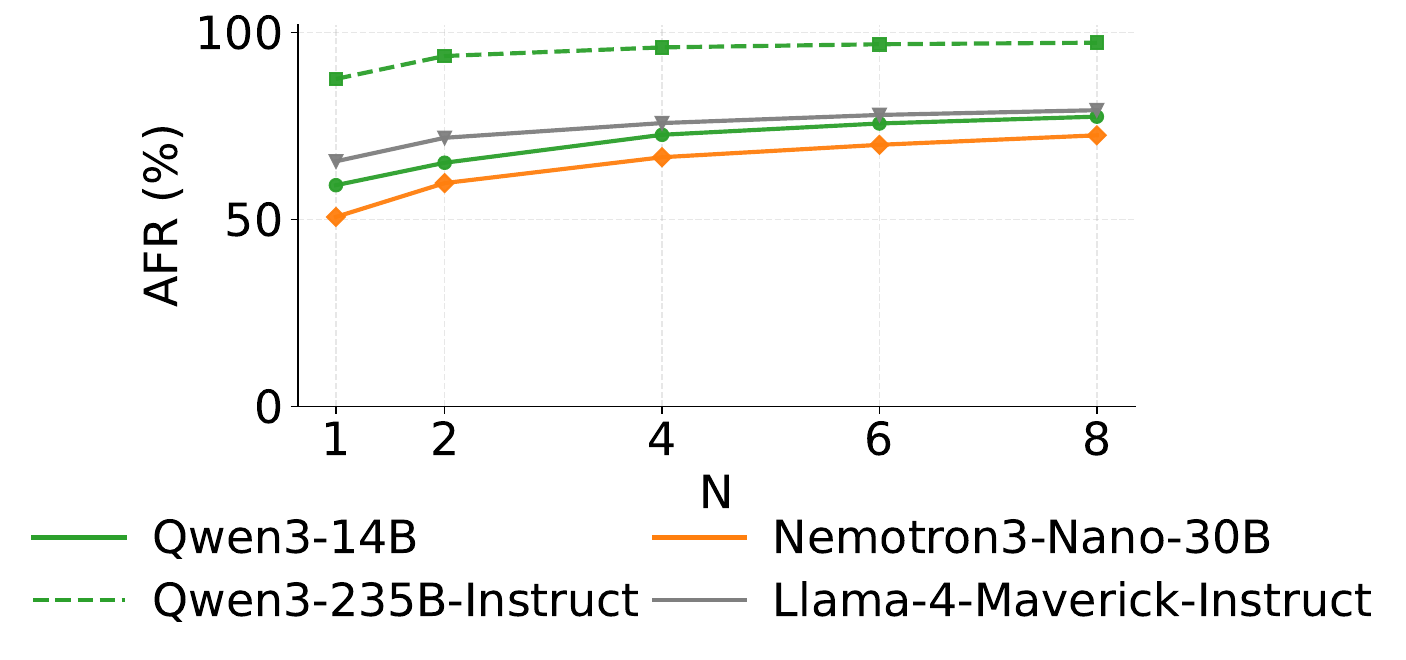}
        \caption{}
        \label{fig:bestofn_afr}
    \end{subfigure}
    \hfill
    \begin{subfigure}[t]{0.48\linewidth}
        \centering
        \includegraphics[width=\linewidth]{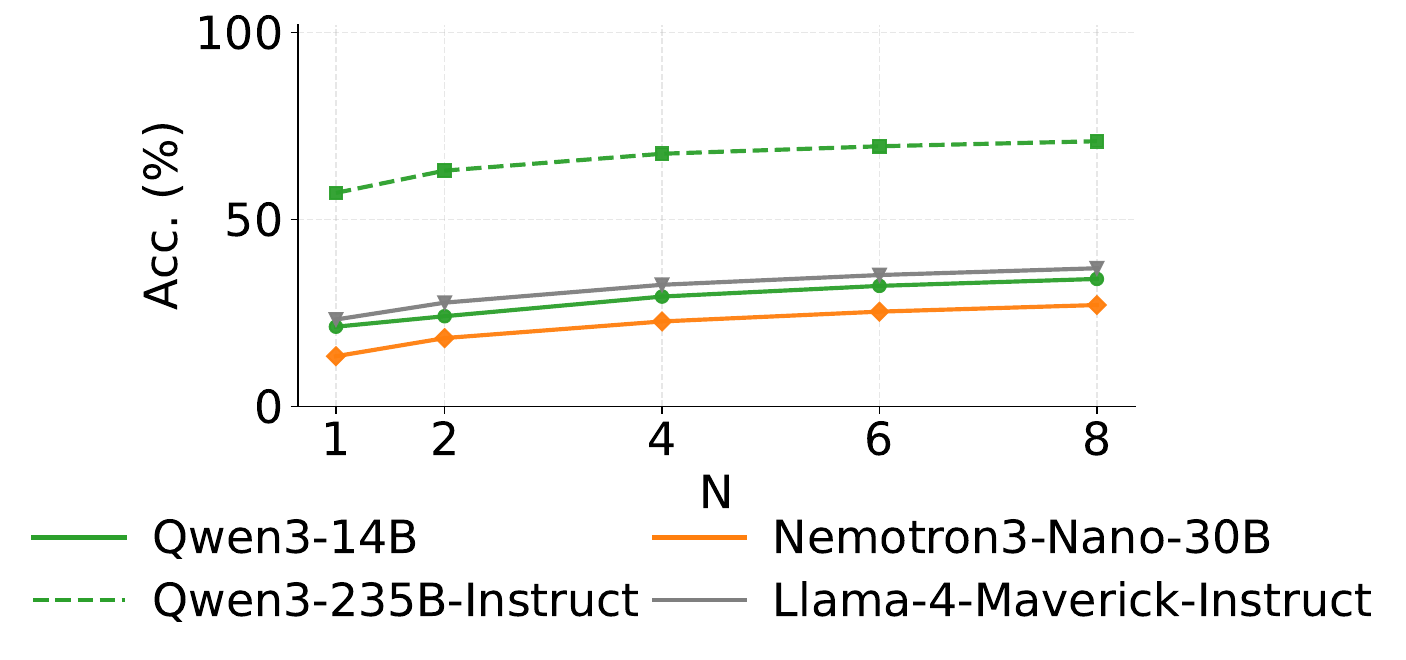}
        \caption{}
        \label{fig:bestofn_acc}
    \end{subfigure}

    \begin{subfigure}[t]{0.48\linewidth}
        \centering
        \includegraphics[width=\linewidth]{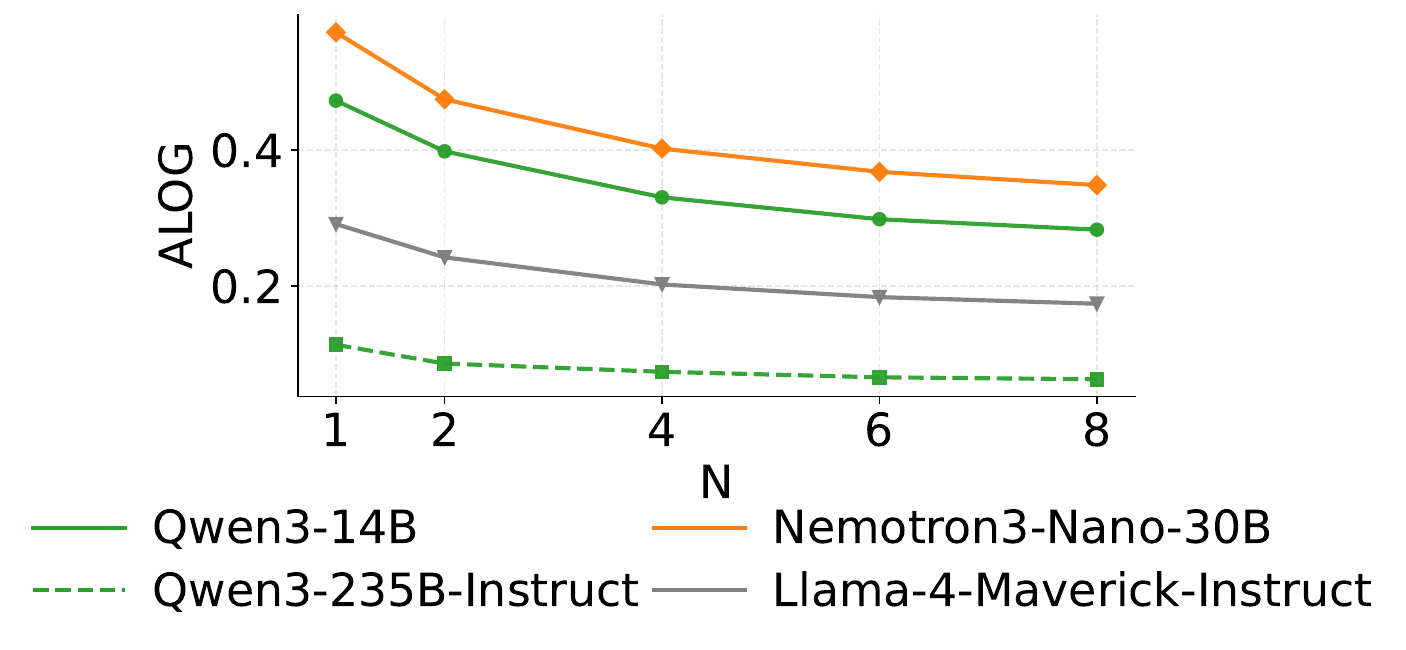}
        \caption{}
        \label{fig:bestofn_alog}
    \end{subfigure}
    \hfill
    \begin{subfigure}[t]{0.48\linewidth}
        \centering
        \includegraphics[width=\linewidth]{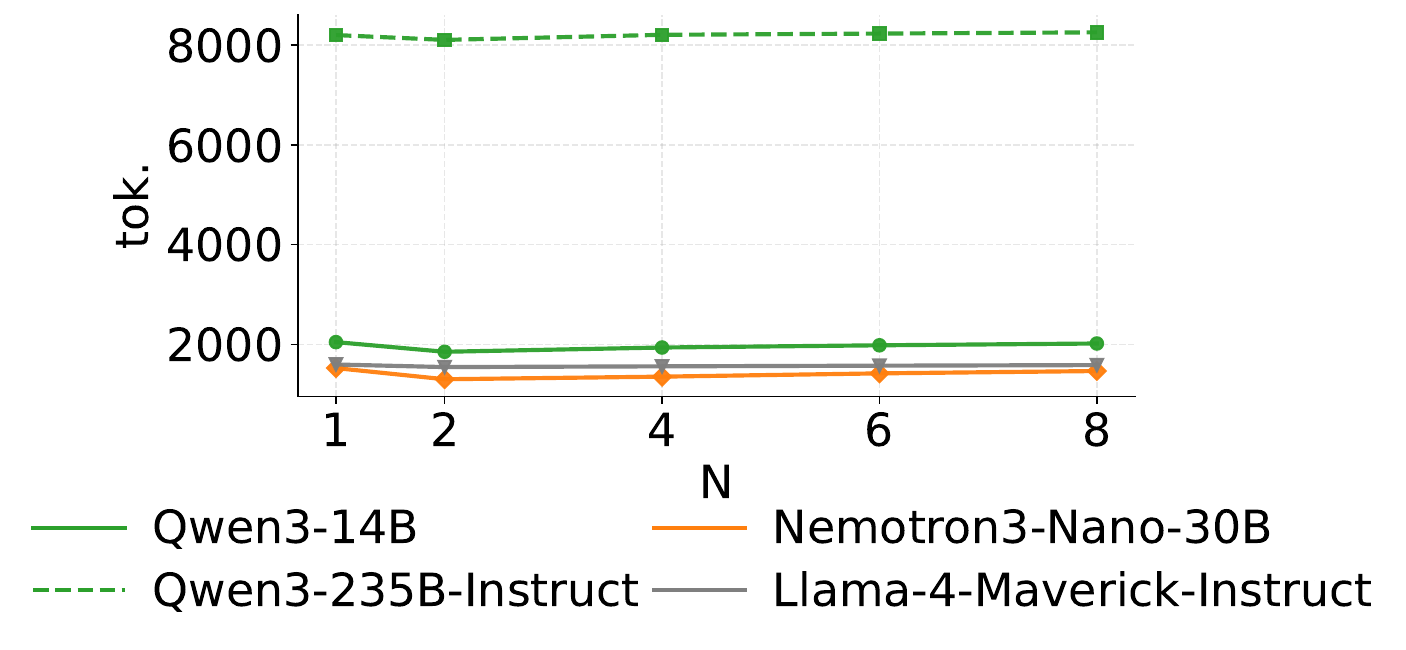}
        \caption{}
        \label{fig:bestofn_tok}
    \end{subfigure}

    \caption{Best-of-$N$ performance scaling across four models.
    We report (a) AFR (\%), (b) Acc. (\%), (c) ALOG, and (d) tok. (of the selected candidate solution) as functions of $N$.}
    \label{fig:best_of_n_scaling}
\end{figure*}


\paragraph{Example Chain-of-thought (CoT) reasoning traces. } Taking the instance shown in Table~\ref{tab:cvrp-instance-prompt} as an example, we present the CoT processes of DeepSeek-V3.2 in Table~\ref{tab:cvrp-s}, Table~\ref{tab:cvrp-m}, and Table~\ref{tab:cvrp-l} for different tiers. Across the 3 difficulty tiers, the reasoning evolves from distance-explicit, savings-inspired manual optimization (Set-S) to capacity-feasible clustering with light cross-route refinement (Set-M), and finally to scalable geographic decomposition with pragmatic route splitting (Set-L). In Set-S, the model computes many explicit Euclidean distances, identifies geographic clusters, and uses Clarke–Wright–style “savings” intuition to justify linking nearby customers far from the depot; it then iteratively improves by reordering stops within a route and explicitly comparing the two-route solution against a three-route alternative, making the reasoning strongly numerical and improvement-driven. In Set-M, the model prioritizes feasibility first by constructing two capacity-balanced groups, then performs a small cross-route move (swapping a customer between vans) guided by approximate route-length calculations; the reasoning is less exhaustive than Set-S (fewer pairwise checks) and reads like a construct-then-refine heuristic rather than a detailed savings narrative. In Set-L, the model shifts to a more scalable strategy: it starts with coarse geographic clustering and distance-to-depot cues, assigns customers to routes under a larger capacity, and then introduces an additional route by splitting a spatially dispersed cluster to avoid extreme detours; the emphasis is on managing spatial spread and tour complexity rather than detailed distance arithmetic, reflecting a transition from local numerical optimization to higher-level decomposition and route-structure control.

\paragraph{Scaling Inference-time Computation. } 

A natural question is whether increasing inference-time computation can improve reasoning outcome (i.e., solution feasibility and optimality), without changing the model or training. We therefore evaluate a simple compute-scaling strategy: Best-of-$N$ sampling, where we generate $N$ independent candidate solutions per instance and select the best one (feasibility first, then ALOG metric among feasible solutions). This isolates the effect of allocating more test-time compute to search over candidates.

Figure~\ref{fig:best_of_n_scaling} shows consistent benefits as $N$ increases. First, feasibility improves monotonically: even moderate sampling (e.g., $N{=}2$) yields clear gains, indicating that additional rollouts can reduce the chance of producing an invalid solution.
Second, solution quality also improves, suggesting that compute scaling helps not only ``get a feasible answer'' but also ``get a better one.''
Meanwhile, the gains exhibit diminishing returns: most improvements occur at small $N$, while the curves flatten as $N$ grows, implying a practical knee point for trading off compute and performance.
Finally, Figure~\ref{fig:best_of_n_scaling} (d) shows that the token usage of the selected solution is largely stable across $N$, so the improvements are primarily driven by exploring more CO solutions rather than producing longer single outputs; however, the total inference cost still scales roughly linearly with $N$ due to multiple generations.

Overall, best-of-$N$ provides a simple and effective way to scale inference-time computation, improving both feasibility and optimality, but it does not eliminate the performance gap between LLMs: stronger models remain ahead across all $N$.

\onecolumn

\begin{center}

\begin{instance}{Problem Overview}
\noindent
\textbf{Task:} Capacitated Vehicle Routing Problem (CVRP) \\
\textbf{Instance tiers:} Set-S / Set-M / Set-L \\
\textbf{Instance format:} Someone from the store is lining up the drivers and thinking about how to split the work: every driver should start at the shop, cover a set of houses with their requested groceries exactly once (nothing can be skipped or visited twice), and return to the shop without exceeding the van's load limit. To pick the best way, add up the distance each van travels (using Euclidean distance between coordinates) and choose the arrangement with the smallest total distance. The concrete list of locations, coordinates, grocery demands, and the van capacity follows below. There are $N$ locations in total; the shop is node \texttt{A}, and each van can carry up to $C$ units.

\begin{verbatim}
<INSTANCE_DATA>
\end{verbatim}

Assign routes so that: (i) each route starts and ends at the shop \texttt{A}; (ii) every house \texttt{B--J} is visited exactly once across all routes; (iii) the total demand on each route never exceeds $C$; and (iv) the sum of travel distances across all routes is minimized.

Also, to keep things tidy, when you send back the routing plan just stick to a simple JSON shape like this so I can read it straight away:

\begin{verbatim}
{ "solution": [[shop_id, house_id, ... , shop_id],
              [shop_id, house_id, ... , shop_id],
              ...] }
\end{verbatim}

Pretty much: \texttt{"solution"} is the outer list of routes; each inner list is one route that \emph{starts and ends} with the \texttt{shop\_id} (here, \texttt{A}) and lists the visited \texttt{house\_id} values in travel order. It's just an example of the shape I expect---don't treat these placeholders as real answers. Please use the exact identifiers from the instance input---no renaming and no made-up labels. For example: ``Valid identifiers look like plain numbers such as \texttt{1} or \texttt{23}, single capital letters like \texttt{A} or \texttt{B}, or a capital letter followed by digits like \texttt{A1} or \texttt{X7}.'' Please think step by step, output your reasoning trace, and then output the JSON in the required format.

\end{instance}
\vspace{20pt}
\captionof{table}{CVRP instance example used for LLM evaluation.}
\label{tab:cvrp-instance-prompt}

\cotrowsep

\begin{cotrow}{CoT (Set-S / CVRP example)}
Alright, let's go step by step.

---

(*@\textbf{1. Understanding the problem}@*)

We have:
- Shop: node A (0, 71), demand 0.
- Houses: B, C, D, E, F, G, H, I, J, each with given coordinates and demand.
- Van capacity: 100 units.
- Goal: Partition houses into routes (each starting and ending at A) so that total demand per route <= 100, and total Euclidean travel distance is minimized.
- Output: JSON list of routes, each route like ["A", ... ,"A"].

---

(*@\textbf{2. Data table}@*)

From the table:

| Node | x | y | demand |
|------|---|---|--------|
| A | 0 | 71 | 0 |
| B | 62 | 6 | 12 |
| C | 71 | 8 | 24 |
| D | 62 | 90 | 3 |
| E | 76 | 9 | 14 |
| F | 87 | 59 | 19 |
| G | 70 | 0 | 4 |
| H | 100 | 31 | 24 |
| I | 75 | 100 | 19 |
| J | 68 | 98 | 2 |

Total demand (excluding A):
12 + 24 + 3 + 14 + 19 + 4 + 24 + 19 + 2 = 121 units.

Since capacity = 100, we need at least 2 vans.
Possible to do in 2 vans because 121 <= 200.

---

(*@\textbf{3. Euclidean distances}@*)

Let's compute distances between all points (rounded to 2 decimals for reasoning, but final route choice will be based on minimizing total).

From A (0,71):
- A->B: sqrt((62-0)^2 + (6-71)^2) = sqrt(3844 + 4225) = sqrt(8069) ~= 89.83
- A->C: sqrt(71^2 + (8-71)^2) = sqrt(5041 + 3969) = sqrt(9010) ~= 94.92
- A->D: sqrt(62^2 + (90-71)^2) = sqrt(3844 + 361) = sqrt(4205) ~= 64.85
- A->E: sqrt(76^2 + (9-71)^2) = sqrt(5776 + 3844) = sqrt(9620) ~= 98.08
- A->F: sqrt(87^2 + (59-71)^2) = sqrt(7569 + 144) = sqrt(7713) ~= 87.82
- A->G: sqrt(70^2 + (0-71)^2) = sqrt(4900 + 5041) = sqrt(9941) ~= 99.70
- A->H: sqrt(100^2 + (31-71)^2) = sqrt(10000 + 1600) = sqrt(11600) ~= 107.70
- A->I: sqrt(75^2 + (100-71)^2) = sqrt(5625 + 841) = sqrt(6466) ~= 80.41
- A->J: sqrt(68^2 + (98-71)^2) = sqrt(4624 + 729) = sqrt(5353) ~= 73.16

---

Between houses (just note clusters):
- Lower-left cluster: B(62,6), C(71,8), E(76,9), G(70,0) -- all near bottom-left of grid, far from A horizontally but far in y from A.
- Upper cluster: D(62,90), I(75,100), J(68,98) -- near top. Actually A is at y=71, so D is at y=90 (19 vertical), I at y=100 (29 vertical), J at y=98 (27 vertical). D is closest to A among these.
- Middle-right: F(87,59), H(100,31).

---

(*@\textbf{4. Demand grouping for capacity}@*)

We want to minimize total travel, so likely group geographically.

Cluster 1 (Lower-left): B(12), C(24), E(14), G(4) -> total = 54 units. Could add more if nearby.
Cluster 2 (Upper): D(3), I(19), J(2) -> total = 24 units. Could add F or H if they fit geographically.
Cluster 3 (Middle-right): F(19), H(24) -> total = 43 units. But we must assign all to vans, possibly 2 vans only.

---

Check if 2 vans possible:
Van1: Cluster1 (54) + maybe D (3) = 57, still room.
Van2: Cluster3 (43) + Cluster2 without D (I,J = 21) = 64, total 64.

That leaves D unassigned in that scheme, but D is close to A and I,J, so maybe:
Better: Van1: B,C,E,G (54) + F (19) = 73 (F is not too far from lower-left? Actually F is 87,59 -- far from B,C,E,G in x but y=59 is between top and bottom). Might not be good geographically.

Let's think of savings heuristic (Clarke-Wright).

---

(*@\textbf{5. Clarke-Wright savings intuition}@*)

Savings s(i,j) = d(A,i) + d(A,j) - d(i,j).
We want to merge routes if savings high and capacity allows.

Let's compute a few key savings (approx):

B and C:
d(A,B)~=89.83, d(A,C)~=94.92,
d(B,C)=sqrt((71-62)^2+(8-6)^2)=sqrt(81+4)=sqrt(85)~=9.22
Savings ~= 89.83+94.92-9.22 ~= 175.53 -> huge because B and C are very close.

B and G:
d(B,G)=sqrt((70-62)^2+(0-6)^2)=sqrt(64+36)=sqrt(100)=10
Savings ~= 89.83+99.70-10 ~= 179.53 -> even bigger.

C and E:
d(C,E)=sqrt((76-71)^2+(9-8)^2)=sqrt(25+1)=sqrt(26)~=5.10
Savings ~= 94.92+98.08-5.10 ~= 187.90 -> huge.

So B,C,E,G all close to each other, far from A -> big savings if linked.

Upper cluster:
D and J:
d(D,J)=sqrt((68-62)^2+(98-90)^2)=sqrt(36+64)=sqrt(100)=10
d(A,D)~=64.85, d(A,J)~=73.16
Savings ~= 64.85+73.16-10 ~= 128.01

D and I:
d(D,I)=sqrt((75-62)^2+(100-90)^2)=sqrt(169+100)=sqrt(269)~=16.40
Savings ~= 64.85+80.41-16.40 ~= 128.86

I and J:
d(I,J)=sqrt((68-75)^2+(98-100)^2)=sqrt(49+4)=sqrt(53)~=7.28
Savings ~= 80.41+73.16-7.28 ~= 146.29

So upper cluster D,I,J also good to link.

F and H:
d(F,H)=sqrt((100-87)^2+(31-59)^2)=sqrt(169+784)=sqrt(953)~=30.87
d(A,F)~=87.82, d(A,H)~=107.70
Savings ~= 87.82+107.70-30.87 ~= 164.65 -> big, so F and H should be together.

---

(*@\textbf{6. Try feasible 2-route partition}@*)

We need each route <= 100 demand.

Option 1:
Route1: B(12), C(24), E(14), G(4) = 54, add F(19) = 73, add H(24) = 97 (within 100).
That leaves D,I,J for other route: D(3)+I(19)+J(2)=24, well under 100.

Check distances:
Route1: A -> B -> C -> E -> G -> F -> H -> A? Or better order to minimize distance.
Route2: A -> D -> I -> J -> A.

Option 2:
Route1: B,C,E,G,F = 73
Route2: D,I,J,H = 48 (3+19+2+24=48)
But H is far from D,I,J, so maybe not good.
Better: Route1: B,C,E,G,H = 78 (H linked to G? far)
Route2: D,I,J,F = 43 (F far from D,I,J) -> seems worse.

So Option 1 seems better: Route1: Lower-left + F + H; Route2: Upper cluster (D,I,J).

---

(*@\textbf{7. Optimize route order within each}@*)

Route1: A -> ... -> A with B,C,E,G,F,H.

Start from A: A to nearest among them?
A to B=89.83, A to G=99.70, A to F=87.82, A to H=107.70, A to C=94.92, A to E=98.08.
Actually F is closest to A in this group (87.82), but F is in middle-right, not close to B,C,E,G.

Better to start with cluster B,C,E,G then go to F then H then back to A.

Try sequence: A -> B -> C -> E -> G -> F -> H -> A.

Compute total:
A->B: 89.83
B->C: 9.22
C->E: 5.10
E->G: sqrt((70-76)^2+(0-9)^2)=sqrt(36+81)=sqrt(117)~=10.82
G->F: 61.40
F->H: 30.87
H->A: 107.70
Sum ~= 314.94

Route2: A -> D -> I -> J -> A.
A->D: 64.85
D->I: 16.40
I->J: 7.28
J->A: 73.16
Sum ~= 161.69

Total both routes ~= 314.94+161.69=476.63

---

(*@\textbf{8. Try to improve Route1 order}@*)

Maybe A -> G -> B -> C -> E -> F -> H -> A.

A->G: 99.70
G->B: 10.00
B->C: 9.22
C->E: 5.10
E->F: sqrt((87-76)^2+(59-9)^2)=sqrt(121+2500)=sqrt(2621)~=51.20
F->H: 30.87
H->A: 107.70
Sum ~= 313.79 (slightly better than 314.94)

Total with Route2 (161.69) ~= 475.48

Try A -> G -> B -> C -> E -> H -> F -> A:

A->G: 99.70
G->B: 10.00
B->C: 9.22
C->E: 5.10
E->H: sqrt((100-76)^2+(31-9)^2)=sqrt(576+484)=sqrt(1060)~=32.56
H->F: 30.87
F->A: 87.82
Sum ~= 275.27 (much better!)

Check capacity still OK:
B,C,E,G,H,F = 12+24+14+4+24+19=97.

Route2 same: D,I,J = 24 units.

Total ~= 275.27 + 161.69 = 436.96 (big improvement).

---

(*@\textbf{9. Check if 3 routes could be shorter}@*)

If we split into 3 routes, each shorter distances but more A-trips.

Example:
R1: B,C,E,G = 54
A->B->C->E->G->A:
A->B: 89.83
B->C: 9.22
C->E: 5.10
E->G: 10.82
G->A: 99.70
Total ~= 214.67

R2: F,H = 43
A->F->H->A:
A->F: 87.82
F->H: 30.87
H->A: 107.70
Total ~= 226.39

R3: D,I,J = 24
A->D->I->J->A:
Total ~= 161.69

Total ~= 214.67+226.39+161.69=602.75 -> worse than 436.96.
So 2 routes better.

---

(*@\textbf{10. Final best found}@*)

Route1: A -> G -> B -> C -> E -> H -> F -> A (demand 97)
Route2: A -> D -> I -> J -> A (demand 24)
Total distance ~= 436.96.

Final JSON:
{
  "solution": [
    ["A","G","B","C","E","H","F","A"],
    ["A","D","I","J","A"]
  ]
}
\end{cotrow}
\captionof{table}{DeepSeek CoT trace on a Set-S CVRP instance.}
\label{tab:cvrp-s}

\cotrowsep
\begin{cotrow}{CoT (Set-M / CVRP example)}
(*@\textbf{1. Understanding the problem}@*)

We have:

- Shop: node A (demand 0)
- Houses: B, C, D, E, F, G, H, I, J, K (10 houses total)
- Each has a demand in units (given in table).
- Van capacity: 100 units.
- Euclidean distances between coordinates (since coordinates given).
- Need to partition houses into routes starting and ending at A, total demand per route <= 100, minimize total travel distance (sum of lengths of all routes).
- Output: list of routes, each route as [A, ... houses ..., A].

---

(*@\textbf{2. Data extraction}@*)

Coordinates (x, y):

A: (87, 24)
B: (8, 13)
C: (84, 43)
D: (0, 52)
E: (55, 100)
F: (88, 43)
G: (61, 39)
H: (0, 98)
I: (97, 31)
J: (100, 6)
K: (84, 0)

Demands:
B: 23
C: 12
D: 19
E: 19
F: 17
G: 8
H: 12
I: 18
J: 18
K: 21

Total demand = 23+12+19+19+17+8+12+18+18+21 = 167 units.

Since van capacity = 100, minimum number of vans = ceil(167/100) = 2 vans.
But because demands are not huge, maybe 2 vans possible if split well.

---

(*@\textbf{3. Check if 2 vans possible}@*)

We need to partition 10 houses into 2 groups, each <= 100 units.

Total 167 means if one van carries X, the other carries 167-X <= 100 -> X >= 67.
So each van must carry between 67 and 100 units.

Largest demands: B(23), D(19), E(19), F(17), I(18), J(18), K(21), C(12), G(8), H(12).

Try to make two groups ~83.5 each.

Group 1: B(23) + D(19) + E(19) + H(12) = 73, add C(12) -> 85, add G(8) -> 93.
That's: B, D, E, H, C, G = 93 units.

Remaining: F(17), I(18), J(18), K(21) = 74 units.

So yes, 2 vans possible:
Van 1: 93 units (B, D, E, H, C, G)
Van 2: 74 units (F, I, J, K)

---

(*@\textbf{4. Routing for minimal distance}@*)

We need to sequence each route starting and ending at A to minimize total distance.

---

### Van 1: B, D, E, H, C, G (93 units)

Coordinates:
A(87,24), B(8,13), D(0,52), E(55,100), H(0,98), C(84,43), G(61,39).

Try a reasonable path:
A -> B -> D -> H -> E -> G -> C -> A.

Compute distances roughly (Euclidean):

d(A,B) = sqrt((87-8)^2 + (24-13)^2) = sqrt(79^2 + 11^2) = sqrt(6241+121)= sqrt(6362) ~= 79.76
d(B,D) = sqrt((8-0)^2 + (13-52)^2) = sqrt(64 + 1521) = sqrt(1585) ~= 39.81
d(D,H) = sqrt((0-0)^2 + (52-98)^2) = 46
d(H,E) = sqrt((55-0)^2 + (100-98)^2) = sqrt(3025+4)= sqrt(3029) ~= 55.04
d(E,G) = sqrt((61-55)^2 + (39-100)^2) = sqrt(36+3721)= sqrt(3757) ~= 61.30
d(G,C) = sqrt((84-61)^2 + (43-39)^2) = sqrt(529+16)= sqrt(545) ~= 23.35
d(C,A) = sqrt((87-84)^2 + (24-43)^2) = sqrt(9+361)= sqrt(370) ~= 19.24

Total ~= 79.76+39.81+46+55.04+61.30+23.35+19.24 = 324.5

---

### Van 2: F, I, J, K (74 units)

Coordinates:
F(88,43), I(97,31), J(100,6), K(84,0), A(87,24).

Good order: A -> K -> J -> I -> F -> A.

d(A,K) = sqrt((87-84)^2 + (24-0)^2) = sqrt(9+576)= sqrt(585) ~= 24.19
d(K,J) = sqrt((100-84)^2 + (6-0)^2) = sqrt(256+36)= sqrt(292) ~= 17.09
d(J,I) = sqrt((97-100)^2 + (31-6)^2) = sqrt(9+625)= sqrt(634) ~= 25.18
d(I,F) = sqrt((88-97)^2 + (43-31)^2) = sqrt(81+144)= sqrt(225)=15
d(F,A) = sqrt((87-88)^2 + (24-43)^2) = sqrt(1+361)= sqrt(362) ~= 19.03

Total ~= 24.19+17.09+25.18+15+19.03 = 100.49

---

(*@\textbf{5. Check if swapping houses between vans reduces total distance}@*)

Van 1 length ~= 324.5, van 2 ~= 100.5, total ~= 425.

Move C from van 1 to van 2:
Van 1: B, D, E, H, G (23+19+19+12+8 = 81)
Van 2: F, I, J, K, C (17+18+18+21+12 = 86)

Van 2 route: A -> C -> F -> I -> J -> K -> A:
d(A,C)=19.24
d(C,F)= sqrt((88-84)^2+(43-43)^2)=4
d(F,I)=15
d(I,J)=25.18
d(J,K)=17.09
d(K,A)=24.19
Sum ~= 19.24+4+15+25.18+17.09+24.19 = 104.7

Van 1 without C: A -> B -> D -> H -> E -> G -> A:
d(G,A)= sqrt((87-61)^2+(24-39)^2)= sqrt(676+225)= sqrt(901) ~= 30.02
Sum ~= 79.76+39.81+46+55.04+61.30+30.02 = 311.93

Total ~= 311.93+104.7 = 416.63 (better than 425).

---

(*@\textbf{6. Final routes after improvement}@*)

Van 1: B, D, E, H, G (81 units)
Route: A -> B -> D -> H -> E -> G -> A
Length ~= 311.93

Van 2: F, I, J, K, C (86 units)
Route: A -> C -> F -> I -> J -> K -> A
Length ~= 104.7

Total ~= 416.63

---

(*@\textbf{7. Final answer in required JSON}@*)

{
  "solution": [
    ["A","B","D","H","E","G","A"],
    ["A","C","F","I","J","K","A"]
  ]
}
\end{cotrow}
\captionof{table}{DeepSeek CoT trace on a Set-M CVRP instance.}
\label{tab:cvrp-m}

\cotrowsep
\begin{cotrow}{CoT (Set-L / CVRP example)}
Alright, let's go step by step.

(*@\textbf{Step 1 -- Understanding the problem}@*)

We have:

- Shop node ID: `A` (coordinates 43, 54), demand 0.
- Total locations: 19 (shop + 18 houses).
- Van capacity: 150 units.
- Each house has a demand (in units).
- We need to partition the 18 houses into routes, each starting and ending at `A`, total demand per route <= 150.
- Objective: minimize total Euclidean distance traveled by all vans.
- Output: list of routes in JSON format.
---

(*@\textbf{Step 2 -- Data summary}@*)

From the table:

| ID | x  | y  | demand |
|----|----|----|--------|
| A  | 43 | 54 | 0      |
| B  | 76 | 68 | 30     |
| C  | 26 | 32 | 9      |
| D  | 60 | 38 | 21     |
| E  | 28 | 65 | 15     |
| F  | 21 | 90 | 19     |
| G  | 81 | 43 | 11     |
| H  | 79 | 24 | 5      |
| I  | 53 | 16 | 10     |
| J  | 14 | 11 | 9      |
| K  | 5  | 73 | 10     |
| L  | 3  | 51 | 28     |
| M  | 38 | 98 | 7      |
| N  | 43 | 67 | 15     |
| O  | 55 |100 | 11     |
| P  | 97 | 43 | 26     |
| Q  |100 |100 | 6      |
| R  | 0  | 0  | 7      |
| S  | 0  | 92 | 11     |

Total demand (excluding A):
30+9+21+15+19+11+5+10+9+10+28+7+15+11+26+6+7+11 = 250 units.

Minimum number of vans (if perfectly packed): ceil(250/150) = 2 vans, but likely more because of spatial spread.

---

(*@\textbf{Step 3 -- Initial clustering by geography}@*)

Let's group houses by rough region relative to shop `A` (43,54):

1. North-East (higher x, higher y): B(76,68), O(55,100), Q(100,100), M(38,98), N(43,67), maybe F(21,90) is more north-west.
2. East (higher x, near y=54): G(81,43), P(97,43), H(79,24), D(60,38), I(53,16).
3. West (low x): C(26,32), J(14,11), R(0,0), L(3,51), K(5,73), S(0,92), F(21,90), E(28,65), M(38,98) -- but M is north of A.

Actually, let's just compute distances from A to see natural clusters.

---

(*@\textbf{Step 4 -- Distances from A (approx Euclidean)}@*)

Let's compute roughly:

- A to B: dx=33, dy=14 -> dist ~= 35.8
- A to C: dx=-17, dy=-22 -> dist ~= 27.8
- A to D: dx=17, dy=-16 -> dist ~= 23.3
- A to E: dx=-15, dy=11 -> dist ~= 18.6
- A to F: dx=-22, dy=36 -> dist ~= 42.0
- A to G: dx=38, dy=-11 -> dist ~= 39.6
- A to H: dx=36, dy=-30 -> dist ~= 46.9
- A to I: dx=10, dy=-38 -> dist ~= 39.4
- A to J: dx=-29, dy=-43 -> dist ~= 51.5
- A to K: dx=-38, dy=19 -> dist ~= 42.5
- A to L: dx=-40, dy=-3 -> dist ~= 40.1
- A to M: dx=-5, dy=44 -> dist ~= 44.3
- A to N: dx=0, dy=13 -> dist ~= 13.0
- A to O: dx=12, dy=46 -> dist ~= 47.6
- A to P: dx=54, dy=-11 -> dist ~= 55.1
- A to Q: dx=57, dy=46 -> dist ~= 73.4
- A to R: dx=-43, dy=-54 -> dist ~= 68.9
- A to S: dx=-43, dy=38 -> dist ~= 57.4

(*@\textbf{Step 5 – Grouping for TSP-like routes}@*)

We can try to form clusters by proximity and capacity.

Cluster 1 (North-East): B(30), N(15), O(11), M(7), Q(6) → total = 69. Add F(19) → 88. Add E(15) → 103. Add S(11) → 114. Add K(10) → 124. Still under 150.  
But S, K are far west, so maybe not good for distance.

Better:  
Route 1 (North of A): N(15), E(15), M(7), O(11), B(30) → total 78. Could add F(19) → 97. Could add S(11) → 108. But S is far from B/O.  

Actually, let’s try geographic clusters:

West group: R(7), J(9), C(9), L(28), K(10), S(11), F(19), E(15), M(7) → sum = 115. That’s one route under 150. But spatially huge spread --- maybe split.

Better:  
West-South: R, J, C, L, K → sum 7+9+9+28+10 = 63. Add S(11) → 74. Add F(19) → 93. Add E(15) → 108. Add M(7) → 115. Still okay. But visiting all in one route is long in distance.

---

Given time, I’ll try a sensible manual split minimizing distance:

Route 1 (Close North/East): A → N(15) → E(15) → M(7) → O(11) → B(30) → A.  
Demand = 78. Could add more but B is far from O? Actually O to B: $\Delta x=21,\ \Delta y=-32 \Rightarrow \text{dist}\sim 38$. Not too bad.

Route 2 (East): A → D(21) → G(11) → P(26) → H(5) → I(10) → A.  
Demand = 73. Could add Q(6) → 79, but Q is far from I/H. Q is near O/B but already in Route 1? No, Q not in Route 1 yet. Q is far from A though.

Route 3 (West): A → C(9) → L(28) → K(10) → S(11) → F(19) → A.  
Demand = 77. Add E(15)? E is in Route 1 already. Add M? M in Route 1. Add J(9) → 86. Add R(7) → 93.  

Actually, let’s systematically assign:

---

(*@\textbf{Step 6 – Assign with capacity constraint}@*)

We need all 18 houses covered exactly once.

Try clustering:

Cluster A (North): N, E, M, O, B, F, S, K  
Sum = 15+15+7+11+30+19+11+10 = 118.  
This is compact? Not really --- S,K are far west from B,O. Might be inefficient.

Better:  
Cluster 1 (North-East): N(15), E(15), M(7), O(11), B(30), Q(6) → total 84.  
Cluster 2 (East): D(21), G(11), P(26), H(5), I(10) → total 73.  
Cluster 3 (West): C(9), L(28), K(10), S(11), F(19), J(9), R(7) → total 93.

Check all houses:  
We have N,E,M,O,B,Q,D,G,P,H,I,C,L,K,S,F,J,R → 18 houses. Yes.

---

(*@\textbf{Step 7 – Route ordering within clusters}@*)

We want to sequence to minimize distance.

Route 1 (North-East): Start A → N(43,67) → E(28,65) → M(38,98) → O(55,100) → Q(100,100) → B(76,68) → A.  
Check: N→E is close, E→M is north, M→O is east, O→Q is far east, Q→B is southwest, B→A is southwest. Might be improved: maybe A → N → E → B → O → M → Q → A? Let’s keep simple for now.

Route 2 (East): A → D(60,38) → G(81,43) → P(97,43) → H(79,24) → I(53,16) → A.  
Seems fine.

Route 3 (West): A → C(26,32) → L(3,51) → K(5,73) → S(0,92) → F(21,90) → J(14,11) → R(0,0) → A.  
This is very spread out --- maybe split into 2 routes for shorter travel.

---

(*@\textbf{Step 8 – Split West cluster into 2 routes for lower distance}@*)

West cluster demand 93, but visiting R(0,0) after F(21,90) is a huge detour.

Better:  
Route 3a (West-North): A → C(26,32) → L(3,51) → K(5,73) → S(0,92) → F(21,90) → A.  
Demand = 9+28+10+11+19 = 77.  
Route 3b (West-South): A → J(14,11) → R(0,0) → A.  
Demand = 9+7=16. Very short route but maybe merge with another? Could merge J,R into Route 2? But Route 2 is east, J,R are far west --- increases distance a lot.

Maybe merge J,R into Route 3a? Then Route 3a: C,L,K,S,F,J,R → total 93. That’s fine for capacity but bad for TSP length.

Given the trade-off, maybe keep 4 routes total:

Route 1: N, E, M, O, B, Q (84)  
Route 2: D, G, P, H, I (73)  
Route 3: C, L, K, S, F (77)  
Route 4: J, R (16)  

That covers all: N,E,M,O,B,Q,D,G,P,H,I,C,L,K,S,F,J,R → 18.

---

(*@\textbf{Step 9 – Optimize sequence within each route}@*)

Route 1: A → N(43,67) → E(28,65) → M(38,98) → O(55,100) → Q(100,100) → B(76,68) → A.  
Maybe better: A → N → E → B → O → M → Q → A? Let’s pick a reasonable order:  
A → N → E → B → O → M → Q → A.  
Check: N→E close, E→B far? E(28,65) to B(76,68): $\Delta x=48,\ \Delta y=3,\ \text{dist}\sim 48.1$.  
E→M is closer? E to M: $\Delta x=10,\ \Delta y=33,\ \text{dist}\sim 34.5$.  
So maybe A → N → E → M → O → Q → B → A.

Let’s finalize:  
A → N → E → M → O → Q → B → A.

Route 2: A → D → G → P → H → I → A.

Route 3: A → C → L → K → S → F → A.

Route 4: A → J → R → A.

---

(*@\textbf{Step 10 – Final check}@*)

All houses assigned exactly once, each route $\le 150$ capacity.

---

Final routes:

1. ["A","N","E","M","O","Q","B","A"]  
2. ["A","D","G","P","H","I","A"]  
3. ["A","C","L","K","S","F","A"]  
4. ["A","J","R","A"]

---

{
  "solution": [
    ["A", "N", "E", "M", "O", "Q", "B", "A"],
    ["A", "D", "G", "P", "H", "I", "A"],
    ["A", "C", "L", "K", "S", "F", "A"],
    ["A", "J", "R", "A"]
  ]
}

\end{cotrow}
\captionof{table}{DeepSeek CoT trace on a Set-L CVRP instance.}
\label{tab:cvrp-l}

\par\smallskip

\end{center}